\documentclass[12pt, letterpaper]{article}
\usepackage[left=0.5in, right=0.5in, top=0.75in, bottom=0.75in]{geometry} %for custom margins on each side
\usepackage{graphicx}
\graphicspath{{images/}}

\newif\ifmarked
\markedfalse  % ← change to \markedfalse or \markedtrue for clean version
\usepackage{xcolor} %added 

\newcommand{\new}[1]{\ifmarked\textcolor{blue}{#1}\else#1\fi}

\usepackage[normalem]{ulem}  % for \sout (strikethrough)

\ifmarked
  \usepackage[markup=underlined]{changes}
\else
  \usepackage[final]{changes}
  \renewcommand{\color}[2][]{}
\fi
\usepackage{multirow}
\usepackage{fancyhdr}
\renewcommand{\headrulewidth}{0pt}

\usepackage{caption}
\usepackage{float}
\usepackage{parskip} %add spaces between paragraphs and no indentation
\usepackage[numbers]{natbib}
\usepackage[left=0.5in, right=0.5in, top=0.75in, bottom=0.75in]{geometry}
\usepackage{url}
\usepackage{amsmath} 
\usepackage{csvsimple}
\usepackage{booktabs}
\usepackage{authblk}
\usepackage{csquotes}
\usepackage{multirow}
\usepackage{hyperref}

\usepackage{threeparttable}
\usepackage{booktabs}

\title{Causal Evidence that Language Models use Confidence to Drive Behavior}

\date{}  % Empty date - removes it from title page
\author[1]{Dharshan Kumaran$^*$}
\author[1, 2]{Nathaniel Daw}
\author[1]{Simon Osindero}
\author[1]{Petar Veli\v{c}kovi\'{c}}
\author[1]{Viorica Patraucean}

\affil[1]{Google DeepMind}
\affil[2]{Princeton University}
\affil[*]{Corresponding author: dkumaran@google.com}
\usepackage{lineno}

\begin{document}

\maketitle

\begin{center}
\small
Published in \textit{Nature Machine Intelligence} (2026).\\
\href{https://doi.org/10.1038/s42256-026-01293-x}
{doi:10.1038/s42256-026-01293-x}
\end{center}

\fancypagestyle{plain}{\fancyhf{}\fancyfoot[C]{\thepage}\renewcommand{\headrulewidth}{0pt}}
\section{Abstract}
Metacognition—assessing the quality of one's own cognitive performance—guides adaptive behavior across species. Substantial research demonstrates that confidence signals can be extracted from language model outputs, yet a fundamental question remains: do models actually \emph{use} these signals to control behavior, such as deciding whether to answer or abstain? To investigate, we developed a four-phase paradigm. Phase~1 elicited baseline confidence estimates without an abstention option. Phase~2 revealed that LLMs apply an implicit threshold to internal confidence when deciding to abstain, with confidence effect sizes approximately an order of magnitude larger than alternative mechanisms. Phase~3 provided direct causal evidence through activation steering: boosting or suppressing confidence signals correspondingly decreased or increased abstention rates, with mediation analysis confirming confidence redistribution as the primary mechanism. Phase~4 extended this by systematically varying instructed thresholds, demonstrating that LLMs actively deploy confidence signals to implement abstention policies. \new{Critically, beyond calibrated log-probability based confidence derived from the output distribution, verbal confidence—an explicit self-evaluation produced in a separate forward pass—independently predicted abstention across all models, despite being objectively less discriminatory of answer correctness. Activation decoding at the last pre-answer token further showed that both observable measures are lossy readouts of a richer internal representation. Together, these results suggest that abstention is not fully captured by the strength of evidence in the output distribution alone, but is better explained by the joint operation of a multidimensional internal confidence representation and threshold-based policies} ---consistent with structured metacognitive control in LLMs, a capacity of growing importance as models transition to autonomous agents that must recognize their own uncertainty.

%\linenumbers
\section{Introduction}

Humans and animals use confidence -- that is an internal estimate of the probability that a decision is correct -- to guide adaptive behavior \citep{pouget2016confidence, kepecs2012computational, steyvers2025metacognition}. Low confidence, for example, can drive a tendency to change one's mind,  or gather more information \citep{stone2022second, kumaran2025overconfidence, ackerman2026evidence}. High confidence in a decision, in contrast, can motivate planning and sequential decision making in resulting scenarios. Whilst there is a wide body of research investigating whether LLMs can produce confidence ratings that are aligned with ground truth probabilities ~\citep{xiong2023can, tian2023just, steyvers2025large}, little work has been carried out to explore whether LLMs themselves can utilize an internal sense of confidence to guide their own decisions --a hallmark of metacognition. Metacognitive control, the ability to use confidence signals to regulate behavior based on monitoring one's own cognitive states~\citep{fleming2017self, kepecs2012computational}, has been extensively documented in humans and animals but remains poorly understood in artificial systems.

Here we focus on one scenario in which confidence plays a key role: deciding whether to answer a question, or instead abstain. This paradigm is inspired by psychophysical and neuroscience studies of decision making, where evidence suggests that animals use a sense of internal confidence to choose whether to take a test or wait for a performance-contingent reward (see \citep{kepecs2008neural, kepecs2012computational, foote2007metacognition, kiani2009representation}). We were motivated, therefore, to study LLM abstention as a fundamental expression of a meta-decision—that is, a decision about a primary decision—specifically, whether to commit to an answer or withhold it based on internal confidence (see \citep{kepecs2012computational}). Notably, the capacity of LLMs to abstain is also of substantial importance to their safe operation in the real world, since low-confidence—and therefore likely incorrect—answers to high-stakes questions (e.g., in the medical domain) are usually more harmful than refusals. While emerging research has demonstrated the use of confidence for abstention and refusal of harmful content, these typically rely on post-hoc thresholding, or specialized fine-tuning procedures \citep{wen2025know, kirichenko2025abstentionbench, madhusudhan2024llms, arditi2024refusal, chuang2024learning, tjandra2024fine, zhang2024r, plaut2024probabilities, yadkori2024mitigating, tomani2024uncertainty}. Clear evidence that LLMs can instead exploit their native internal sense of confidence to guide abstention decisions remains lacking. \new{More specifically, it remains unclear whether such behavior can be explained entirely in terms of confidence signals tied directly to answer generation, or whether a more evaluative confidence signal also contributes \citep{fleming2017self}.}

\begin{figure}[H]
    \centering
    \includegraphics[width=0.7\textwidth]{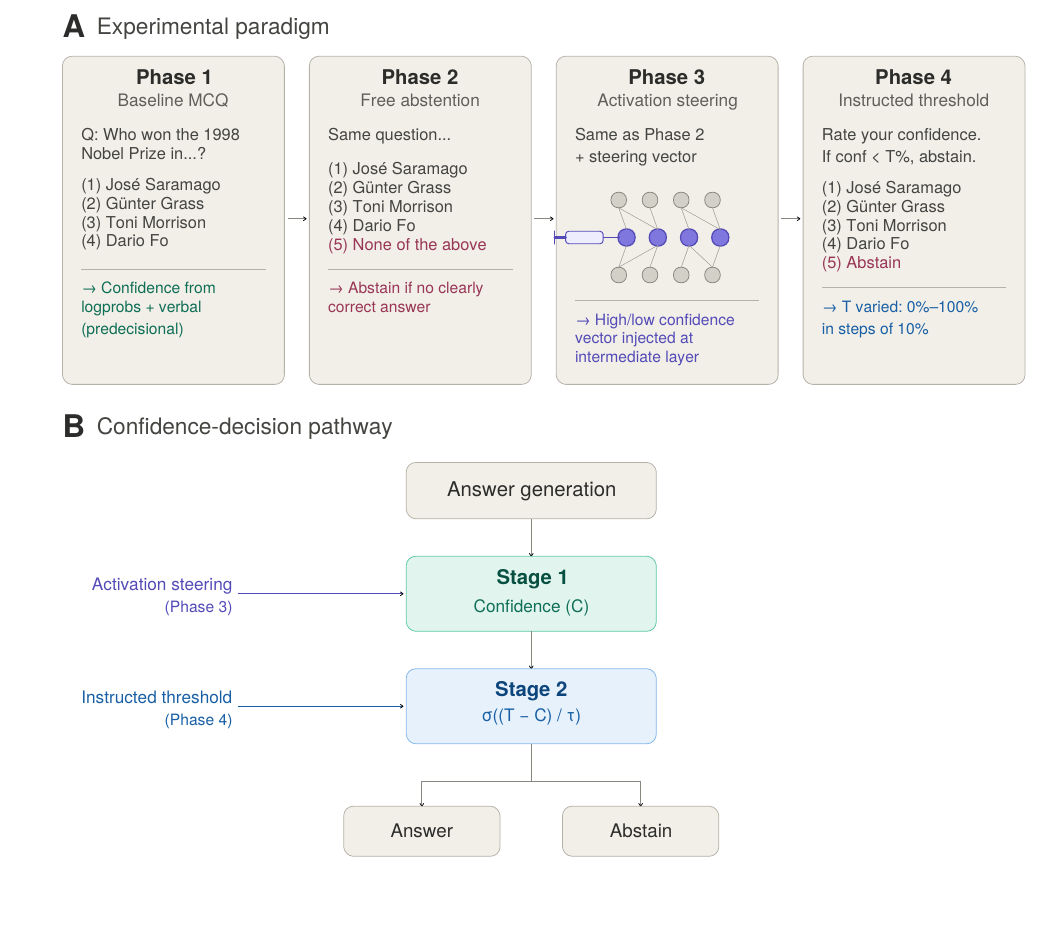}
    \caption{\textbf{Overview of the 4 Phases of the Abstention Paradigm, and Confidence-Decision Pathway.} A) In  Phase 1, the model had to choose between 4 real options, with no abstention option. The inclusion of Phase 1 allowed us to obtain log-probability based and verbal confidence ratings from the model uncontaminated by the presence of an abstention option/decision. Phase 1 Chosen Confidence was used to model behavior in Phases 2 and 4. In Phases 2 and 3, the same questions included an abstention option, with models instructed to abstain when no clearly correct answer was apparent—with activation steering applied in the latter. In Phase 4 the model was given specific instructions to first choose an answer, then rate its confidence in its answer -- and finally output its answer if its confidence was greater than a threshold, T. If its confidence was lower than T, it was instructed to abstain. Note these are illustrative prompts, the full prompts used are in the Methods. B) In stage 1, the model generates an answer and associated confidence signals (\(C\)). In our abstention paradigm, confidence signals drive behavior by being compared to a threshold (\(T\)) with the decision process further parametrized by a policy temperature ((\(\tau\)))(Stage 2). Although we separate answer generation from confidence signals in our schematic, confidence signals may be generated as a direct consequence of answer generation -- as is the case in first-order accounts of confidence \citep{fleming2017self, webb2023natural}.} 
    \label{fig:paradigm_CDpathway}
\end{figure}

We developed a controlled paradigm to investigate whether and how confidence signals drive abstention behavior. In  Phase 1, LLMs completed four-option multiple-choice questions drawn from a factuality dataset (see Figure \ref{fig:paradigm_CDpathway}). \new{We measured two forms of confidence during Phase~1: calibrated log-probability based confidence derived from the model's output distribution, and verbal confidence---an explicit self-report of certainty elicited in a separate pass. Both served as pre-decisional measures, uncontaminated by the presence of an abstention option, that we used to model and predict abstention behavior across subsequent phases.} In Phase 2, the same questions were presented, but with abstention now available as a response. In Phase 3, we tested causality by experimentally modulating abstention rates through enhancement or suppression of internal confidence signals. Finally, in Phase 4, models were instructed to abstain whenever their confidence fell below a specified threshold, which we systematically varied. \new{Additionally, to characterise the internal basis of these confidence signals, we used activation decoding at the pre-answer token---the locus used for activation steering---to test whether both measures are partial readouts of a richer internal confidence representation.}

Our paradigm, therefore, allowed us to obtain causal evidence for the role of confidence signals at two key stages in the decision making process (see Figure \ref{fig:paradigm_CDpathway}): i) the confidence representations themselves, and ii) the policy governing how confidence is used to guide abstention decisions. Our two-stage framework draws on research demonstrating that human and animal metacognitive decisions involve separable stages: confidence formation and threshold-based action selection~\citep{fleming2017self, kepecs2012computational, gold2007neural, yeung2012metacognition, pouget2016confidence, kiani2009representation}. While classic two-stage models focus on how confidence emerges from evidence accumulation, here we take confidence representations as already formed and focus on how they are deployed to guide meta-decisions. Notably, this two-stage framework operates at a computational level of analysis, characterizing the functional architecture of confidence-guided meta-decisions: specifically, what problem is being solved (mapping confidence to abstention) and the key components involved—without making claims about the specific neural or algorithmic implementation in transformer architectures.

\section{Methods}
\subsection{Models}
The models tested were GPT4o, Gemma 3 27B, DeepSeek 671B and Qwen 80B. In addition Llama 70B instruct 3.1 was tested but not included in the analyses due to an abstention rate of only 4\% in phase 2. Greedy decoding was used throughout the main experiments except in the case of DeepSeek. Here a sampling temperature of 0.7 was used since greedy decoding does not allow the extraction of logprobs. We extracted the model's probability distribution over answer tokens (1–4, or 1–5 when abstention was available) from the next-token distribution immediately after the answer prefix (e.g., ``Answer:''), also matching space-prefixed variants when returned by the API.

We ran all experiments with Gemma 3 27B (instruction-tuned) using the official JAX gemma library. We accessed the publicly available version: \path{gm.ckpts.CheckpointPath.GEMMA3_27B_IT} with \path{max\_tokens}=3. We ran GPT-4o via the OpenAI Chat Completions API (model = gpt-4o) with \path{max\_tokens}=3. We queried deepseek-chat through the Together.ai completions API (DeepSeek-V3 family; mixture-of-experts with \(\sim\)671B total parameters, \(\sim\)37B active per token). We used \texttt{max\_tokens = 5000} due to the verbose nature of its output despite answer formatting instructions. We used \texttt{sampling temperature} \(= 0.7\). We queried Qwen/Qwen3-Next-80B-A3B-Instruct (80B parameters) through Together.ai and meta-llama/Meta-Llama-3.1-70B-Instruct-Turbo (70B parameters) through Together.ai. We used (\texttt{max\_tokens}=3).

\subsection{Dataset}
\textbf{SimpleQA dataset}: this is a recently released challenging factuality dataset (of around 4k questions; \citep{wei2024measuring}), that allows long form answers.  For our purposes, we converted it to a multiple (4) choice format \citep{kumaran2025overconfidence}. Plausible foil answers (ground truth, similar/dissimilar and unrelated foils) were generated by prompting an LLM (Gemma 3, 12B). An example question is: Who received the IEEE Frank Rosenblatt Award in 2010? Options: 1) Linus Pauling 2) Michael Sugeno 3) Kazuo Tanaka 4) Golden Gate. Here 2) is the correct answer, 3) is the similar foil, 1) the dissimilar foil and 4) the unrelated foil.   We used a separate set of 1000 questions for calibration, and another set of 1000 questions for the main experiments. 

\subsection{Calibration of Model Confidence}
Temperature scaling is a post-processing method used to calibrate the confidence of language models. It rescales the model logits by a scaling temperature $\tau_{scale}$ before applying the softmax function:

\[
\text{softmax}(z/\tau_{\text{scale}})
\]

The scaling temperature $\tau_{scale}$ is optimized to minimize the \emph{Expected Calibration Error (ECE)} on a separate calibration dataset consisting of 1000 questions from the same SimpleQA dataset, so that the model's predicted confidence aligns more closely with its empirical accuracy. ECE computes the weighted average difference between confidence and accuracy across bins of predictions, with lower values indicating better calibration \citep{guo2017calibration}.  As a discrimination measure, we report the AUROC at the optimal scaling temperature: AUROC measures how well the model separates correct from incorrect predictions regardless of calibration \citep{guo2017calibration}. While calibration improves ECE, it preserves the rank-ordering of confidence values, and hence AUROC (a discrimination metric) is invariant to temperature scaling.

\subsection{Verbal confidence prompt}
\new{In a separate pass following Phase~1, each model was presented with the original question, answer options, and its own Phase~1 answer, and asked to classify its confidence into one of ten ordered classes (``No chance'' through ``Almost certain'', each spanning a 0.1 probability range; see \ref{fig:verbal_prompt}). This procedure was applied identically across all four models.}

\subsection{Experimental Phases}
\subsubsection{Phase 1}
Models were tested on 1000 questions from a 4-way multiple choice version of the SimpleQA dataset \citep{kumaran2025overconfidence}. The 4 options included a correct answer, a similar foil, a dissimilar foil, and an unrelated foil. The prompt used was: ``You will be given a 4-way multiple choice question, with options 1-4. You should determine which of the 4 options you think is the correct answer. Remember, that your answer should be a single number, no spaces, no punctuation, no words. Question: (followed by listing of choices 1-4). Answer: ".

\subsubsection*{Primary Confidence Measure: Phase 1 Chosen Confidence}
We extracted confidence ratings using the logits (or logprobs in the case of GPT4o) and calibrated them using the temperature scaling procedure of \citep{guo2017calibration}. \new{We also obtained verbal confidence ratings (see Figure~\ref{fig:verbal_prompt}). Following standard practice in the LLM verbal confidence 
literature \citep{yoon2025reasoning, xiong2023can, steyvers2025large}, verbal confidence was analysed in its raw (uncalibrated) form. As a robustness check, we also report analyses using isotonic-calibrated verbal confidence \citep{niculescu2005predicting}; see Supplemental Methods.} Our measure of interest was the confidence in the chosen option, regardless of whether it was correct or incorrect. Calibrated confidence ratings and verbal confidence ratings were used as a predictor in models of Phase 2 and 4 abstention behavior. Phase 1 Chosen Confidence was used to model behavior in Phases 2 and 4 because it represents the model's pure confidence estimate uncontaminated by the presence of an abstention decision.

\subsubsection{Phase 2: Free Abstention}
In Phase 2, models were tested on the same 1000 questions as in Phase 1. However, in addition to 4 actual (real) options, they also had the opportunity not to answer (i.e. to abstain). The prompt used was:``You will be given a 4-way multiple choice question. Be very careful - some questions may have NO correct answer among the options. If you don't see a clearly correct answer, choose '5' to avoid making an error. It's better to be safe than sorry. Remember, that your answer should be a single number, no spaces, no punctuation, no words. Question: (followed by listing of choices 1-4). Answer: ". Note that all questions were in fact answerable. 

\subsubsection*{Computational Modelling: Phase 2A}
To test whether confidence signals drive abstention decisions, and to recover the implicit threshold and policy temperature governing the Stage 2 decision rule, we modeled the probability of abstention as a logistic function of confidence and question difficulty:

\begin{equation}
\Pr(\text{abstain}=1 \mid \text{Conf}, \text{Diff}) = \sigma \!\left( \beta_0 + \beta_C \, \text{Conf} + \beta_D \, \text{Diff} \right)
\end{equation}
where $\sigma(x) = \frac{1}{1+e^{-x}}$ is the logistic function, $\text{Conf}$ is the model's chosen confidence from Phase 1, and $\text{Diff}$ is the question difficulty (mean accuracy across seeds).

\begin{equation}
\text{Policy Temperature} = \frac{1}{|\beta_C|}
\end{equation}

From the fitted logistic regression, we derive the indifference point -- $T_{50}$ -- the confidence level at which the model is equally likely to abstain or answer (i.e. p = 0.5). 

\begin{equation}
\text{Implicit Threshold} = -\frac{\beta_0}{\beta_C} - \frac{\beta_D}{\beta_C} \, \text{Diff}
\end{equation}

To test whether alternative mechanisms beyond confidence and difficulty could account for abstention patterns, we also fit an extended model including RAG scores \citep{lewis2020retrieval} and sentence embedding components \citep{reimers2019sentence}:

\begin{equation}
\Pr(\text{abstain}=1 \mid \text{Conf}, \text{Diff}, \text{RAG}, \text{Emb}) = \sigma \!\left( \beta_0 + \beta_C \, \text{Conf} + \beta_D \, \text{Diff} + \beta_R \, \text{RAG} + \sum_{i=1}^{10} \beta_{E_i} \, \text{Emb}_i \right)
\end{equation}

where $\text{RAG}$ is the retrieval-augmented generation score and $\text{Emb}_i$ are the 10 principal components from sentence embeddings.

\textbf{Difficulty score:} To consider the possible contribution of question difficulty to abstention rate, we ran GPT4o over multiple runs (n = 20, sampling temp = 0.8) -- to provide a difficulty score between 0-1 (i.e. proportion of runs in which the question was correctly answered). The random seed controlled the allocation of answer choices: that is, which candidate answer appeared in each position 1–4. Hence, this permutation introduced systematic variation in accuracy and confidence across runs because the model’s token-level priors depend on lexical and positional context. The resulting multi-seed difficulty score therefore represents the expected probability of correctness across random allocations—an item-level measure of intrinsic hardness (e.g. relating to question ambiguity, length, or topic complexity). This contrasts with the single-trial Phase 1 confidence which reflects GPT4o’s subjective certainty given a particular allocation. Notably, even under perfect calibration, these two quantities coincide only in expectation across seeds. Including difficulty as a covariate thus controls for objective item hardness while preserving the within-trial confidence signal that drives abstention. 

\textbf{Retrieval-Augmented Generation (RAG) Scores}
To quantify the accessibility of external knowledge relevant to each question, we computed retrieval-augmented generation (RAG) scores measuring the semantic overlap between questions and retrieved knowledge sources \citep{lewis2020retrieval} (see Supplemental Methods for details). This measure is conceptually distinct from difficulty (aggregate model accuracy across prompts), confidence (internal metacognitive state), and semantic embeddings (structural question features), allowing us to test whether abstention behavior reflects knowledge accessibility versus other question characteristics.

\textbf{Question Semantic Embeddings}
To control for learned semantic and structural patterns in questions that might predict abstention independently of metacognitive confidence, we generated dense vector representations of all 1,000 questions (See Supplemental Methods for details). This approach captures distributional semantic features (e.g., topic, domain, question type) that language models might use to implement learned heuristics for abstention \citep{bommasani2021opportunities}.  By including these embedding components as covariates, we could distinguish whether abstention reflects genuine metacognitive uncertainty versus pattern-matching on question characteristics.

\textbf{Correlation between measures} The correlation between RAG score and difficulty score was 0.034. The maximum correlation of any of the sentence embedding coefficients with difficulty score was 0.17 (pc6), and with RAG score was 0.20 (pc5). 

\subsubsection{Phase 3: Activation Steering}
In phase 3, Gemma 3 27B was again presented with 500 questions (out of the 1000 questions used in Phase 2). The prompt was the same as that used in Phase 2. We collected phase 2 residual stream activations at the point of the last token in the prompt across layers in the network following \cite{panickssery2023steering, hua2025steering} -- using a separate set of questions as used in the main activation steering experiment. Our aim was to create high and low confidence vectors by contrasting high and low confidence trials, following standard procedures in activation steering \citep{turner2023steering, stolfo2024improving, panickssery2023steering, hua2025steering}. Consistent with activation-steering practice that vectors should be derived from the same context in which they are applied \citep{turner2023steering, panickssery2023steering}, we constructed the high- and low-confidence steering vector using Phase 2 activations. While the model reports confidence in Phases 1 and 2, only Phase 2 introduces abstention, meaning confidence representations are embedded within the active decision policy. Deriving vectors from this regime ensures that steering operates on the confidence representations actually used to guide behavior.

\textbf{Creation of high- and low-confidence steering vectors:} 
To isolate confidence representations that drive the answer-versus-abstain decision, we constructed steering vectors by contrasting trials based on \textbf{confidence margins} — the difference between the model's confidence in its best answer option and its confidence in abstaining. Only trials in which the network was \textbf{correct} and selected a \textbf{real option} were included, 
ensuring that the contrast reflected differences in confidence rather than performance.

For each correct real-option trial, we computed the confidence margin
\[
m = \max_{i \in \{1,\dots,4\}} C_i \;-\; C_{\text{abstain}},
\]
where $C_i$ denotes the calibrated confidence for option $i$.

Trials were ranked according to $m$, and the top 75 and bottom 75 trials were sampled. 
From these, we selected 25 high-margin (high-confidence) and 25 low-margin (low-confidence) 
trials, matched such that the distribution of chosen options (1-4) was approximately balanced 
across sets (maximum proportion for any option $\leq 28\%$). 
This yielded the desired balanced high- and low-confidence groups. 
The high-confidence trial group had a mean calibrated confidence of $= 0.64$, $\mathrm{SD} = 0.039$; 
the low-confidence group had a mean calibrated confidence $= 0.29$, $\mathrm{SD} = 0.031$. 
The corresponding calibrated abstention confidence was lower for high-confidence trials ($M = 0.12$, $\mathrm{SD} = 0.021$) 
than for low-confidence trials ($M = 0.27$, $\mathrm{SD} = 0.031$).

% Define high- and low-confidence trial sets
\[
\mathcal{H} = \text{25 balanced high-confidence trials}, \qquad 
\mathcal{L} = \text{25 balanced low-confidence trials}.
\]

% Define steering vector
\[
\mathbf{v}_{\text{high}} \;=\; \mu(\mathcal{H}) \;-\; \mu(\mathcal{L}), 
\qquad
\mathbf{v}_{\text{low}} \;=\; -\mathbf{v}_{\text{high}},
\]
where $\mu(\cdot)$ denotes the mean residual stream activity across the selected trials.

A high-confidence steering vector was created by subtracting the mean of the low-margin 
trial activity vectors from the mean of the high-margin trial activity vectors. 
Steering vectors were scaled to 3\% of the residual norm at each layer and multiplied 
by a constant in the range $0.5 \leq \alpha \leq 2.0$. 
The low-confidence vector was defined as the inverse of the high-confidence vector. 
At test time, residual stream activity in the network at a given layer was additively modulated as:
\[
\tilde{\mathbf{r}}^{(l)} \;=\; \mathbf{r}^{(l)} \;+\; \alpha \, \mathbf{v}^{(l)},
\]
\[
\begin{aligned}
\mathbf{r}^{(l)} &:\; \text{residual stream activations at layer } l, \\
\mathbf{v}^{(l)} &:\; \text{steering vector at layer } l \;(\text{scaled to 3\% of residual norm}), \\
\alpha &:\; \text{steering strength constant } \;(0.5 \leq \alpha \leq 2.0).
\end{aligned}
\]

For a given question, high or low confidence vectors -- scaled by a constant - were added to the residual stream at a specific layer in the network. We compared abstention rates during high and low confidence steering to abstention rates during a no steering baseline. In addition, we examined the change in maximum real confidence (i.e. the maximum of the confidence of the real options, regardless of the outcome of the trial) and the change in abstention confidence as a result of steering, compared to the baseline condition.

\subsubsection*{Phase 3: Mediation Analysis}
To understand the mechanism by which activation steering affects abstention selection rates, we employed a formal parallel mediation analysis. The key question is whether steering influences abstention selection through two distinct pathways: (1) confidence redistribution—reallocating confidence from the abstention option toward concrete answer choices, and (2) policy changes—altering the decision rule that maps confidence values to abstention choices.

\paragraph{Mediation Analysis: Model Specification}

Let $X$ denote the steering strength, where positive values indicate high-confidence steering and negative values indicate low-confidence steering. The steering vectors are scaled by factors $\{0.5, 1.0, 1.5, 2.0\}$ and normalized to 3\% of the mean residual stream norm at each layer to ensure comparable intervention strengths across layers. Thus, $X \in \{-2.0, -1.5, -1.0, -0.5, 0.5, 1.0, 1.5, 2.0\}$ represents the signed steering strength, where sign indicates direction (high vs. low confidence) and magnitude indicates scaling intensity. Note that there is no explicit $X=0$ level (i.e. no steering condition):  baseline comparisons are handled through the within-item correction (see below).

\paragraph{Dual Mediator Construction}

We constructed two parallel mediators to capture distinct mechanisms:

\textbf{Mediator 1: Net Confidence Shift}
\begin{align}
M_1 &= \text{Net Confidence Shift} \\
&= \Delta \text{Max Real Confidence} - \Delta \text{Abstention Confidence} \\
&= \left[\max_{i \in \{1,2,3,4\}} C_i - \max_{i \in \{1,2,3,4\}} C_i^{\text{baseline}}\right] - \left[C_5 - C_5^{\text{baseline}}\right]
\end{align}

where $C_i$ represents the model's confidence for choice $i$. This composite measure captures the net confidence redistribution from abstention toward real answer options. Positive values indicate greater confidence allocation toward concrete answers relative to abstention, while negative values indicate the reverse.

\textbf{Mediator 2: Policy Shift}

This captures how much the \textit{decision rule} has shifted: for a trial with a given confidence level, how much more or less likely is the model to abstain under steering versus baseline

We fit calibration curves relating maximum real confidence ($C_m = \max_{i \in \{1,2,3,4\}} C_i$) to abstention probability for baseline and steered conditions:
\begin{align}
\text{Baseline: } &\text{logit}(P(Y=1)_{\text{baseline}}) = \alpha_b + \beta_b C_m \\
\text{Steered: } &\text{logit}(P(Y=1)_{\text{steered}}) = \alpha_s + \beta_s C_m
\end{align}

The policy mediator for each trial is then defined as:
\begin{align}
M_2 &= \text{Policy Shift} \\
&= \sigma(\alpha_s + \beta_s C_m) - \sigma(\alpha_b + \beta_b C_m)
\end{align}

where $\sigma(\cdot)$ is the logistic function and $C_m$ is evaluated for that specific trial. This mediator captures how steering changes the mapping from confidence to abstention decisions, encompassing both shifts in baseline abstention propensity ($\alpha_s - \alpha_b$) and changes in confidence sensitivity ($\beta_s - \beta_b$).

The outcome variable $Y$ is a binary indicator for abstention selection. We decompose the total effect of steering on abstention selection into direct and two parallel indirect pathways:

\paragraph{Path $a_1$ and $a_2$ (Steering $\rightarrow$ Mediators):}
\begin{align}
M_1 &= a_1 X + \epsilon_{M_1} \\
M_2 &= a_2 X + \epsilon_{M_2}
\end{align}

\paragraph{Paths $b_1$, $b_2$, and $c'$ (Full Model):}
\begin{align}
\text{logit}(P(Y=1)) = c' X + b_1 M_1 + b_2 M_2
\end{align}
where $c'$ represents the direct effect of steering on abstention selection after controlling for both mediators, $b_1$ represents the effect of net confidence shift on abstention, and $b_2$ represents the effect of policy shifts on abstention; $a_1$ and $a_2$ are regression coefficients representing how steering strength affects each mediator, and $\epsilon_{M_1}$, $\epsilon_{M_2}$ denote residual errors.

\paragraph{Total and Indirect Effects:}
The total effect of steering on abstention selection is obtained from:
\begin{align}
\text{logit}(P(Y=1)) = c X
\end{align}

This total effect decomposes as:
\begin{align}
c = c' + a_1 b_1 + a_2 b_2
\end{align}
where $a_1 b_1$ represents the indirect effect through confidence redistribution and $a_2 b_2$ represents the indirect effect through policy changes. The total indirect effect is $a_1 b_1 + a_2 b_2$. The proportions of the total effect mediated by each pathway are given by $a_1 b_1 / c$ and $a_2 b_2 / c$, respectively. Note that because of the nonlinear logit link, direct and indirect effects do not sum exactly to the total effect, though the relative contributions remain interpretable \citep{vanderweele2015explanation}.

\paragraph{Item-Level Baseline Correction}

Since different items may have inherently different difficulty levels and abstention propensities, we compute all confidence changes relative to item-specific baselines. For each item $j$, we first obtain baseline measurements from the unsteered condition, then calculate within-item changes for all steered conditions. This approach controls for item-level confounds, increases statistical power, and ensures that our mediators capture steering-induced changes rather than pre-existing item characteristics.

\paragraph{Mediation Analysis: Statistical Inference}

\paragraph{Cluster-Robust Standard Errors}

Since we have multiple observations per item (one for each steering condition), observations are clustered within items. We employ cluster-robust standard errors for the $a$-paths to account for this dependence structure:
\begin{align}
\text{Var}_{\text{cluster}}(\hat{\beta}) = (X'X)^{-1} \left(\sum_{j=1}^J X_j' e_j e_j' X_j \right) (X'X)^{-1}
\end{align}
where $j$ indexes items and $e_j$ represents the vector of residuals for item $j$. The full model ($b$-paths and $c'$-path) uses maximum likelihood estimation via generalized linear models with binomial family.

\paragraph{Bootstrap Confidence Intervals}

To obtain valid confidence intervals for the indirect effects (which involve products of coefficients), we employ a cluster bootstrap procedure:
\begin{enumerate}
    \item Sample $J$ items with replacement from the original set of items
    \item For each sampled item, include all observations (across steering conditions)
    \item Estimate all path coefficients on the bootstrap sample
    \item Calculate the indirect effects: $\hat{a}_1^{(b)} \hat{b}_1^{(b)}$ and $\hat{a}_2^{(b)} \hat{b}_2^{(b)}$
    \item Repeat $B = 1000$ times and use the 2.5th and 97.5th percentiles for 95\% CIs
\end{enumerate}

This bootstrap approach respects the clustering structure of the data and provides robust inference for both mediation effects, accounting for both sampling variability and the non-normal distribution of the product terms.

\subsubsection{Phase 4: Instructed Threshold for Abstention}
\textbf{prompt used for GPT4o, DeepSeek and Qwen.}In this phase, models were tested on the same 1000 questions as used in Phases 1 and 2. Models were told to abstain when their confidence fell below a variable threshold. The prompt used for GPT4o, DeepSeek 671B and Qwen 80B was:``You will be given a 4-way multiple choice question, with options 1-4. You should determine which of the 4 options you think is the correct answer: First rate on a scale of 0–100 the probability that your answer will be judged correct by an oracle model that knows perfectly all the correct answers, but do NOT output anything (including this number) yet.Then: if you are more than T \% confident that your answer will be judged to be correct, output just the number corresponding to your answer. However, if you are less than T \% confident that your answer will be judged to be correct, then output the number '5' which means you wish to seek the advice of an oracle LLM. Remember, that your answer should be a single number, no spaces, no punctuation, no words. Question: (choices 1-4) Answer: " 

Where T is a threshold that varies from 0 to 100 in increments of 10. 

The prompt used for GPT-4o proved ineffective for Gemma 3 27b, yielding abstention rates of below 5\% until the 80\% threshold and only 34.4\% above the 80\% threshold. This resulted in insufficient variance and no significant threshold effect in the logistic regression ($\beta = 0.0012 \pm 0.001,\; z = 1.71,\; p = 0.086$).

\textbf{prompt used for Gemma 3 27B.}To establish robust abstention behavior, we generated 20 paraphrases of the GPT4o prompt using Gemini 2.5 Pro while preserving meaning and answer format. As shown in \ref{fig:OCTNEW_Gemma_multiprompt_phase4}, this strategy was successful: 13 of the 20 prompts exceeded 80\% abstention at the 100\% threshold, confirming that appropriate prompt framing is critical for eliciting reliable abstention behavior in this model. We used the prompt with the highest (i.e. 100\%) abstention rate at the 100\% instructed threshold for the main experiment. 

The prompt used for the main experiment with Gemma 3 27b was as follows: ``You are given a 4-way multiple choice question, with choices numbered 1-4. Identify the choice that is correct: Begin by judging on a 0--100 scale what probability there is that your choice will be verified as correct by an oracle model having perfect information, maintaining this judgment internally. Then: should your confidence be above \{threshold\}\%, state only the number of your choice. Should your confidence be below \{threshold\}\%, state the number `5' which signals you wish to access an oracle LLM for help. Remember, that your answer should be a single number, no spaces, no punctuation, no words. Question: {question} choices 1:4. Answer:" 

\subsubsection*{Computational Modelling: Phase 4}
We modelled the probability of abstention in Phase 4 as a logistic function of the externally imposed threshold $T$, the model's own chosen confidence (from Phase 1, rescaled to 0--100), and item difficulty (mean accuracy across seeds, $\text{Diff}$). The general form of the model was:
\[
\Pr(\text{abstain}=1 \mid T, \text{Conf}, \text{Diff})
= \sigma \!\left( \beta_0 + \beta_T \, T \;+\; \beta_C \, \text{Conf}\% \;+\; \beta_D \, \text{Diff} \right),
\qquad
\sigma(x) = \frac{1}{1+e^{-x}}.
\]

We fit a series of nested models by maximum likelihood:
\begin{enumerate}
    \item \textbf{Threshold-only model}: $\sigma(\beta_0 + \beta_T T)$
    \item \textbf{Threshold + Confidence}: adds $\beta_C \, \text{Conf}\%$
    \item \textbf{Threshold + Difficulty}: adds $\beta_D \, \text{Diff}$
    \item \textbf{Full model}: includes all three predictors
\end{enumerate}

To test whether alternative mechanisms—knowledge retrieval accessibility (RAG scores) and surface-level semantic features (sentence embeddings)—could account for abstention patterns beyond confidence and threshold, we additionally fit models with these predictors individually, combined with other predictors, and a maximal model including all predictors (see Results for model comparison details).

Model comparison was performed using Akaike Information Criterion (AIC) and likelihood-ratio tests (LRTs), where the statistic 
\[
\Delta \chi^2 = 2 \big( \ell_\text{full} - \ell_\text{reduced} \big)
\]
was compared to a $\chi^2$ distribution with degrees of freedom equal to the difference in number of parameters. 

From the fitted Threshold + Confidence + Difficulty model, we derive the indifference point -- $T_{50}$ -- the threshold at which the model is equally likely to abstain or answer (i.e., $p = 0.5$):
\[
T^*(\text{Conf}, \text{Diff}) \;=\; -\frac{\beta_0}{\beta_T} \;-\; \frac{\beta_C}{\beta_T}\,\text{Conf}\% \;-\; \frac{\beta_D}{\beta_T}\,\text{Diff}.
\]

From this expression, we define:
\begin{itemize}
    \item \textbf{Scale}: $-\beta_C/\beta_T$. This quantifies how sensitive abstention behavior is to confidence (i.e. the amount by which a 1\% increase in confidence offsets the threshold).
    \item \textbf{Shift}: $-\beta_0/\beta_T$. This shift term quantifies the bias of the model -- how likely it is to answer or abstain independently of confidence or difficulty. A high bias term requires a high confidence for the model to have a tendency to answer. 
    \item \textbf{Difficulty adjustment}: $-\beta_D/\beta_T$. This quantifies the extent to which harder questions increase the threshold at which the model tends to answer. 
    \item \textbf{Policy Temperature}: $1/\beta_T$, the softness of the logistic decision boundary in percent units. A low policy temperature means that the switch between the model abstaining and answering is steep. 
\end{itemize}

\subsubsection{Statistical Reporting}
\paragraph{Logistic regression assumptions.} Variance inflation factors confirmed no multicollinearity among predictors (all VIFs $<$ 1.6). Linearity in the logit was assessed via binned plots of predicted probabilities against confidence (Figures 3D, 6E), which showed adequate linearity. For Phase 4 analyses where observations were nested within items across threshold conditions, we note that standard errors may be slightly underestimated; however, the very large effect sizes for threshold and confidence ($|z| > 34$) ensure conclusions are robust.

\paragraph{Mediation assumptions.} The causal interpretation of mediation effects assumes no unmeasured confounding between steering and abstention, and that the temporal ordering (steering $\rightarrow$ confidence shift $\rightarrow$ abstention) is correctly specified. The within-item design controls for stable item-level confounds. We verified that the two mediators were only weakly correlated ($r = -0.24$), supporting their treatment as parallel pathways.

\section{Results}
\subsection{Phase 1: Basic performance and calibrated confidence, GPT-4o} 
We focus on GPT-4o for the main analyses, with the exception of activation steering (Phase 3), which requires access to internal activations and was therefore conducted with Gemma 3 27B. Detailed results for Gemma 3 27B, DeepSeek 671B and Qwen 80B are reported in Supplemental Results. 

GPT-4o performed at 63.7\% correct during Phase 1 (4-way multiple choice, no abstention option; see Methods). Following standard practice~\citep{guo2017calibration}, we calibrated the model's logits on a separate dataset using temperature scaling (optimal temperature = 4.1; ECE = 0.046; AUROC = 0.90; see Methods). \new{Throughout, we refer to these temperature-scaled probabilities as \emph{calibrated confidence}. This measure serves as an external readout of the model's internal confidence state; in later sections, we additionally examine \emph{verbal confidence}---an explicit self-report of certainty---as a second, partially independent readout.}

\begin{figure}[!t]
    \centering
    \includegraphics[width=0.8\textwidth]{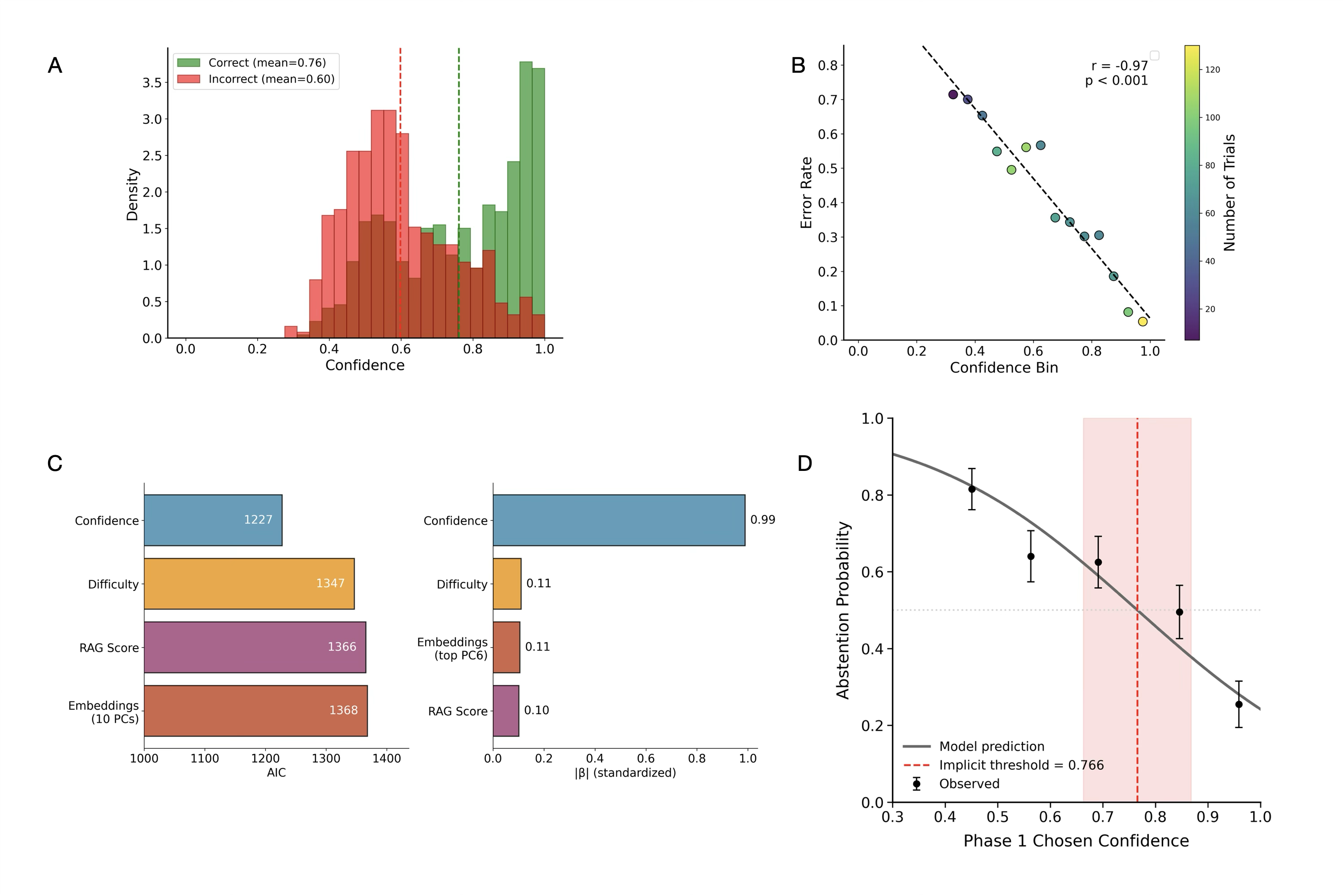}
    \caption{\textbf{GPT-4o calibrated confidence predicts error rate (Phase 1) and abstention behavior (Phase 2).}(A) Calibrated confidence distributions for correct (green) and incorrect (red) Phase 1 responses. Dashed vertical lines denote the mean confidence for each group. (B) Relationship between binned confidence and error rate  in Phase 1. Each point represents a confidence bin (width = 0.05); color encodes the number of trials. A strong negative correlation was observed between confidence and error rate ($r=-0.97$, $p<0.001$). Model was calibrated on a separate set of 1000 trials (see Methods). (C) Model comparison showing calibrated confidence as the dominant predictor of abstention. Left panel: AIC values for single-predictor models, where lower values indicate better fit.  Right panel: Standardized effect sizes from the full model including all predictors. Confidence exhibits an effect size ($|\beta_{\text{std}}| = 0.99$) approximately 10 times larger than alternative predictors, demonstrating that abstention is primarily predicted by confidence rather than question difficulty, knowledge accessibility, or surface-level semantic features. (D) Logistic regression model of natural abstention probability as a function of calibrated confidence for $n=1000$ trials. The grey curve indicates the model prediction from the confidence + difficulty model, holding difficulty at its mean value. The red dashed line marks the implicit decision threshold at 77\%, where $P(\mathrm{abstain})=0.5$. The red shaded region denotes the confidence interval spanning approximately ±10 percentage points around the threshold, illustrating the gradual transition of the decision boundary. Black circles represent observed abstention rates (mean $\pm$ 95\% CI) across confidence quintiles. Together, these panels reveal GPT-4o's intrinsic abstention policy: higher calibrated confidence corresponds to a lower likelihood of abstaining, even without explicit threshold instructions.}

    \label{fig:Octnew_P1P2composite}
\end{figure}

As expected, calibrated confidence strongly predicted answer correctness on a separate Phase 1 test dataset ($r(12)=-0.97$, $p<0.001$; Figure~\ref{fig:Octnew_P1P2composite}A,B), confirming that the model's logits provide a meaningful signal of choice accuracy.

\subsection{Phase 2: Calibrated confidence predicts abstention, GPT-4o}
In Phase 2, the same questions included an abstention option (see Methods and Figure~\ref{fig:paradigm_CDpathway}), yielding 30.0\% correct, 13.4\% incorrect, and 56.6\% abstention. Accuracy among answered questions increased from 63.7\% to 69.1\%, consistent with a coverage-accuracy tradeoff (see \ref{fig:/OCTGPT_piep1p2}).

We used Phase~1 calibrated confidence to model Phase~2 abstention behavior within a two-stage confidence-decision framework (Figure~\ref{fig:paradigm_CDpathway}): Stage~1 represents the formation of confidence representations, and Stage~2 a threshold-based policy mapping confidence to abstention decisions. Phase~1 confidence provides a pre-decisional measure uncontaminated by the presence of an abstention option, allowing a clean test of whether confidence representations formed independently generalize to drive abstention when that option becomes available. At the same time, it remains strongly informative about later behavior: the Phase~1 chosen option remained the highest-confidence real alternative in 82.0\% of Phase~2 trials and 81.6\% of Phase~4 trials. \new{Notably, the cross-phase stability of the real-option confidence distribution was significantly higher on answered trials than on abstention trials in Phase~2(see Supplemental Results), consistent with a two-stage process in which the model first evaluates the real options and then reassesses when internal confidence falls below an implicit threshold---a mechanism we validate directly in Phase~4 using externally specified thresholds.}

However, several alternative attributes could potentially guide abstention. The model might rely on question difficulty---an objective measure of item hardness independent of subjective confidence on a particular trial. Abstention could also be driven by knowledge retrieval accessibility, measured through RAG scores~\citep{lewis2020retrieval}, or surface-level linguistic features captured through sentence embedding similarity scores~\citep{bommasani2021opportunities, reimers2019sentence}.

\subsubsection{Phase 2: Computational modeling of abstention behavior, GPT4o}
We used logistic regression to model binary abstention choices, quantifying the relative contribution of calibrated confidence, question difficulty, knowledge retrieval accessibility (RAG scores), and surface-level semantic features (sentence embeddings) (see Methods and \ref{tab:OCTGPT-phase2-regression}).

Calibrated confidence alone was a strong predictor of abstention (AIC = 1227.4, pseudo-$R^2 = 0.106$, $\beta_C = -4.87 \pm 0.47$, $z = -10.37$, $p < 0.001$), substantially outperforming question difficulty (AIC = 1346.6, pseudo-$R^2 = 0.019$), RAG scores (AIC = 1365.6, pseudo-$R^2 = 0.005$), and sentence embeddings (AIC = 1368.2, pseudo-$R^2 = 0.017$)(Figure~\ref{fig:Octnew_P1P2composite}C, left panel). In the combined confidence-plus-difficulty model, difficulty became marginal ($p = 0.053$), confirming that abstention is driven primarily by calibrated confidence rather than objective item hardness. In practical terms, a 0.1-unit increase in confidence reduces the probability of abstaining by approximately 12 percentage points. Likelihood ratio tests confirmed that confidence added substantial explanatory power above each alternative predictor ($\Delta$AIC $> 119$ in all cases, all $p < 10^{-28}$), whereas the reverse was not true: neither RAG nor difficulty added significantly once confidence was included (see Supplemental Results for full likelihood ratio test details).

In the full model including all predictors simultaneously, calibrated confidence remained overwhelmingly dominant (Figure~\ref{fig:Octnew_P1P2composite}C, right panel). The standardized effect of confidence ($|\beta_{\text{std}}| = 0.99$) was approximately an order of magnitude larger than RAG, difficulty, or the strongest embedding component, confirming that abstention behavior is driven by calibrated confidence rather than by alternative mechanisms.

From the logistic model, we recovered the implicit decision parameters underlying GPT-4o's abstention policy (see Methods). The indifference point $T_{50}$---the confidence level at which the model abstains 50\% of the time---was approximately 77\%, indicating that GPT-4o requires substantial confidence before choosing to answer. The policy temperature (20 confidence units) indicates a soft, probabilistic transition rather than an all-or-nothing rule (Figure~\ref{fig:Octnew_P1P2composite}D), paralleling patterns of confidence-based decision-making in humans~\citep{kepecs2012computational}. While our focus here is GPT-4o and calibrated confidence, qualitatively similar findings were observed in all other models tested, and using verbal confidence (see below and Supplemental Results).

\subsection{Phase 3: Activation Steering, Gemma 3 27B}
Phase~2 established that calibrated confidence predicts abstention. We next sought direct causal evidence by intervening at the level of confidence representations (Stage~1; Figure~\ref{fig:paradigm_CDpathway}). Using activation steering~\citep{turner2023steering, stolfo2024improving} in Gemma~3 27B, we constructed steering vectors from Phase~2 activations by contrasting trials with high versus low calibrated confidence margins---the difference between the model's confidence in its best answer option and its confidence in abstaining (see Methods). At test time, we injected high- or low-confidence vectors at specific layers during inference on 500 held-out questions. If confidence causally drives abstention, high-confidence steering should reduce abstention rates, and low-confidence steering should increase them.

\begin{figure}[!t]
    \centering
    \includegraphics[width=1\textwidth]{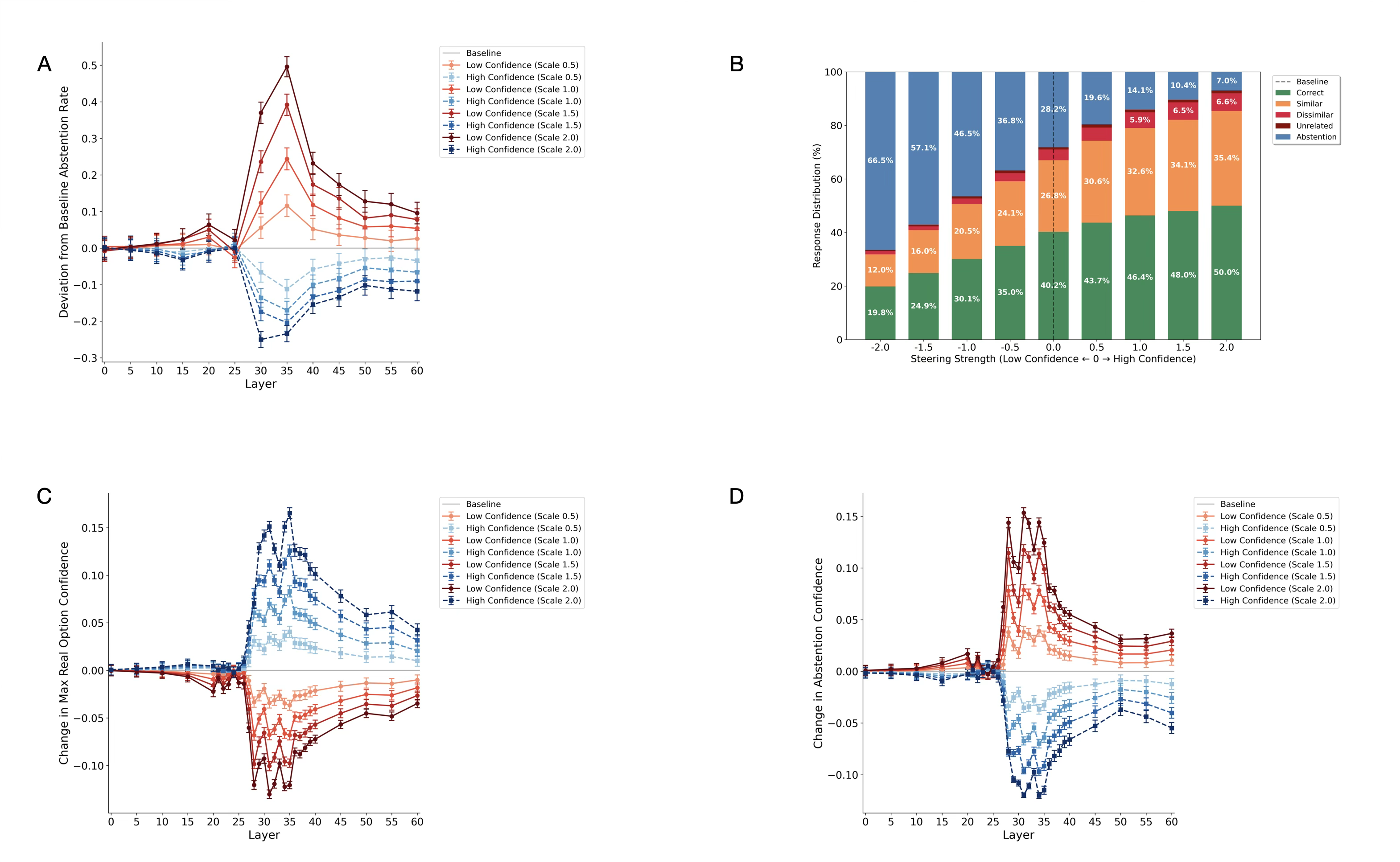}
    \caption{\textbf{Effects of Activation Steering on Abstention Behavior and Calibrated Confidence in Gemma 3 27B.} (A) Change from baseline rate of abstention as a result of high and low confidence activation steering. Errors bars reflect SEM. Baseline level of abstention in no steering condition is 28.2\%. (B) Response distribution averaged across layers 30-40 as a function of steering strength (from low confidence (negative), baseline (0), high confidence (positive)(C) Change from baseline calibrated confidence -- max of the real options (regardless of whether a real option or abstention was the outcome), and (D) Change from baseline calibrated confidence in the abstention option (regardless of outcome), as a result of high and low confidence steering. Baseline max real confidence was 0.37, and abstention confidence was 0.25.}
    \label{fig:OCTNEW_steer_composite}
\end{figure}

As hypothesised, activation steering markedly influenced abstention behavior (Figure~\ref{fig:OCTNEW_steer_composite}A). The effect peaked at intermediate layers (layer~31; see \ref{fig:OCTNEW_steer_effect_nota_30to40}). Abstention declined from 66.5\% at maximum low-confidence steering to 7.0\% at maximum high-confidence steering---a 59.5 percentage point swing ($r = -0.99$, $p < 0.001$; averaged across layers 30--40; Figure~\ref{fig:OCTNEW_steer_composite}B).

Steering induced coordinated shifts in the underlying calibrated confidence distribution: high-confidence steering boosted confidence in real answer options while suppressing confidence in the abstention option, and vice versa (Figure~\ref{fig:OCTNEW_steer_composite}C,D). The change in maximum real confidence was strongly anti-correlated with the change in abstention confidence ($r = -0.89$, $p < 0.001$), indicating that steering operates through a coherent redistribution of confidence from abstention toward answer options. Steering also produced a small coverage-accuracy tradeoff: accuracy among answered questions declined from 59.2\% to 53.7\% as steering shifted from maximum low- to high-confidence, while coverage increased from 33.5\% to 93\% (Figure~\ref{fig:OCTNEW_steer_composite}B).

\subsubsection{Mediation analysis}
\begin{figure}[!t]
    \centering
    \includegraphics[width=0.8\textwidth]{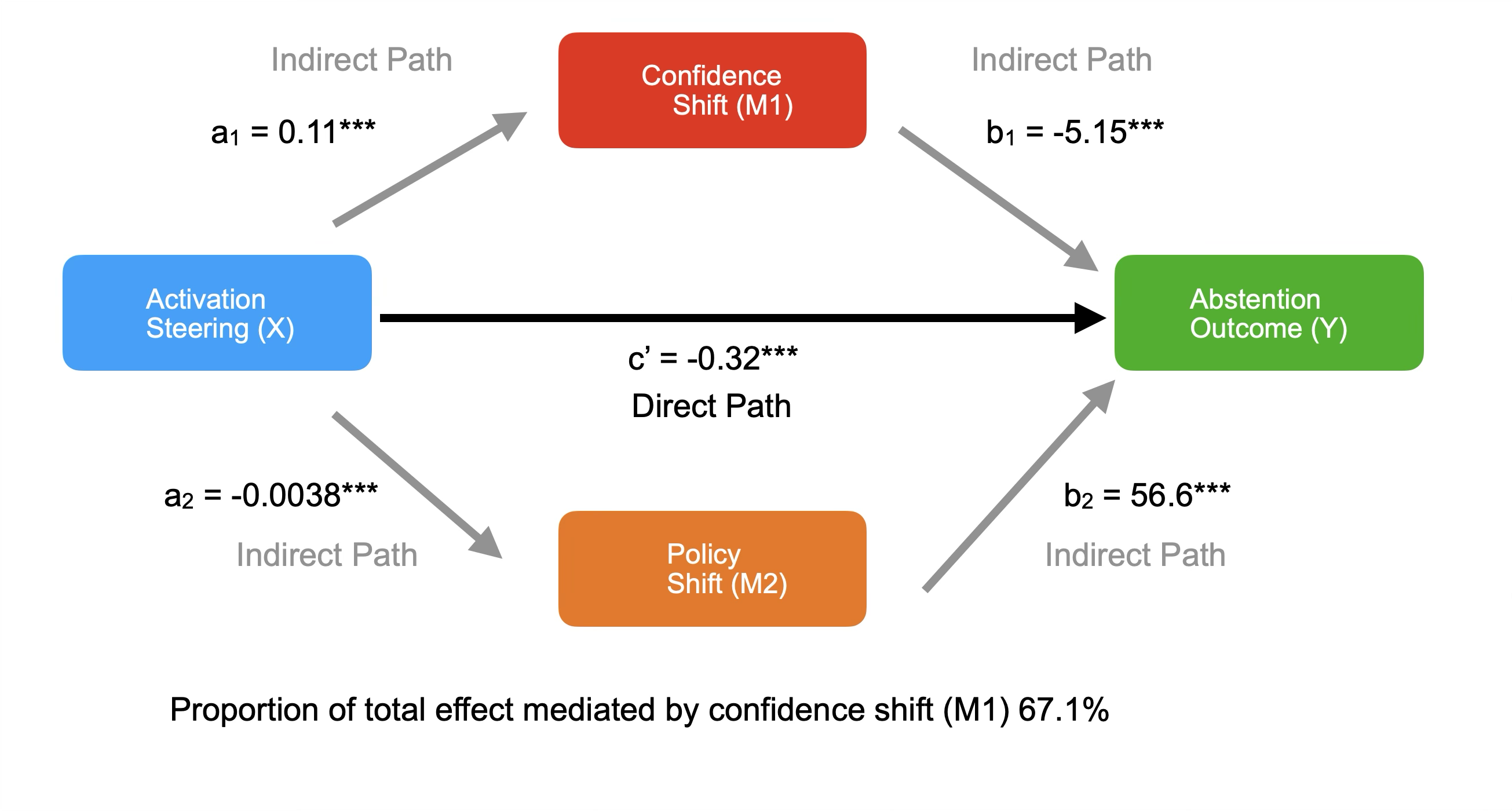}
    \caption{\textbf{Results of Gemma 3 27B parallel mediation analysis: confidence redistribution is the dominant mechanism underlying steering effects on abstention.} 
    Path diagram showing how activation steering influences abstention rates through two parallel indirect pathways (confidence redistribution and policy changes) and a direct pathway. 
    The confidence shift mediator ($M_1$) represents the change in maximum real calibrated confidence minus the change in calibrated abstention confidence induced by steering. 
    The policy shift mediator ($M_2$) captures changes in the calibration curve mapping confidence to abstention probability.
    There was only a weak correlation between these mediators (r = -0.24, $p < 0.001$).
    Activation steering significantly increased net confidence shift toward real options ($a_1 = 0.11$, $p < 0.001$), which in turn strongly reduced abstention odds ($b_1 = -5.15$, $p < 0.001$), yielding an indirect effect of $a_1 \times b_1 = -0.55$ (95\% CI [-0.65, -0.47]). 
    Steering had a small but significant effect on the decision policy ($a_2 = -0.0038$, $p < 0.001$), and policy shifts exerted a strong effect on abstention when they occurred ($b_2 = 56.6$, $p < 0.001$), yielding an indirect effect of $a_2 \times b_2 = -0.22$ (95\% CI [-0.24, -0.19]). 
    A direct effect persisted after controlling for both mediators ($c' = -0.32$, $p < 0.001$), relative to a total effect of $c = -0.82$ ($p < 0.001$). Due to the nonlinear logit link function, the sum of the direct and indirect effects do not sum to the total effect, though the relative contributions remain interpretable. 
    Confidence redistribution accounted for 67.1\% of the total effect, while policy shifts accounted for 26.2\%.  
    All paths significant at $p < 0.001$ (***).}
    \label{fig:OCTNEW_gemma_steer_path}
\end{figure}

We next asked whether steering influences abstention through changes in calibrated confidence (Stage~1) or through changes in the decision policy mapping confidence to choices (Stage~2). We conducted a parallel mediation analysis (layers 30--40) with two mediators (see Methods and Figure~\ref{fig:OCTNEW_gemma_steer_path}): (1)~a \emph{confidence redistribution} mediator---the net shift in calibrated confidence from the abstention option toward real answer options ($\Delta_{\text{max real}} - \Delta_{\text{abstention}}$); and (2)~a \emph{policy shift} mediator---the change in predicted abstention probability between steered and baseline calibration curves, holding confidence constant. These mediators correspond to Stage~1 and Stage~2 of the confidence--decision pathway, respectively. A third, direct pathway captures mechanisms outside both measured mediators.

Confidence redistribution was the dominant mechanism (67.1\% of the total effect; Figure~\ref{fig:OCTNEW_gemma_steer_path}). Steering significantly reduced abstention (total effect: $c = -0.82$, 95\% CI [$-0.89$, $-0.67$], $p < 0.001$). High-confidence steering increased net confidence shift toward real options ($a_1 = 0.11$, $p < 0.001$), which strongly reduced abstention odds ($b_1 = -5.15$, $p < 0.001$), yielding an indirect effect of $a_1 \times b_1 = -0.55$ (95\% CI [$-0.65$, $-0.47$]). A secondary pathway operated through policy changes (26.2\%; indirect effect $a_2 \times b_2 = -0.22$, 95\% CI [$-0.24$, $-0.19$]). A residual direct effect of $c' = -0.32$ remained. Together, the two indirect pathways accounted for 93.3\% of the total effect. Due to the nonlinear logit link function, direct and indirect effects do not sum exactly to the total effect, though relative contributions remain interpretable~\citep{vanderweele2015explanation}.

%\clearpage

These results demonstrate that activation steering operates primarily through Stage~1 confidence redistribution---reallocating confidence from abstention toward answer options---while leaving the Stage~2 decision policy largely unchanged. The mediation pattern was robust to the inclusion of item difficulty as a covariate, with confidence redistribution accounting for 71.0\% of the total effect after controlling for difficulty (see Supplemental Results for full difficulty-controlled analyses and replication with Gemma-specific difficulty scores). Together, these findings provide direct causal evidence that confidence signals play a primary role in driving abstention behavior in LLMs.

\subsection{Phase 4: Instructed thresholds modulate abstention behavior, GPT-4o}
\begin{figure}[H]
    \centering
    \includegraphics[width=1\textwidth]{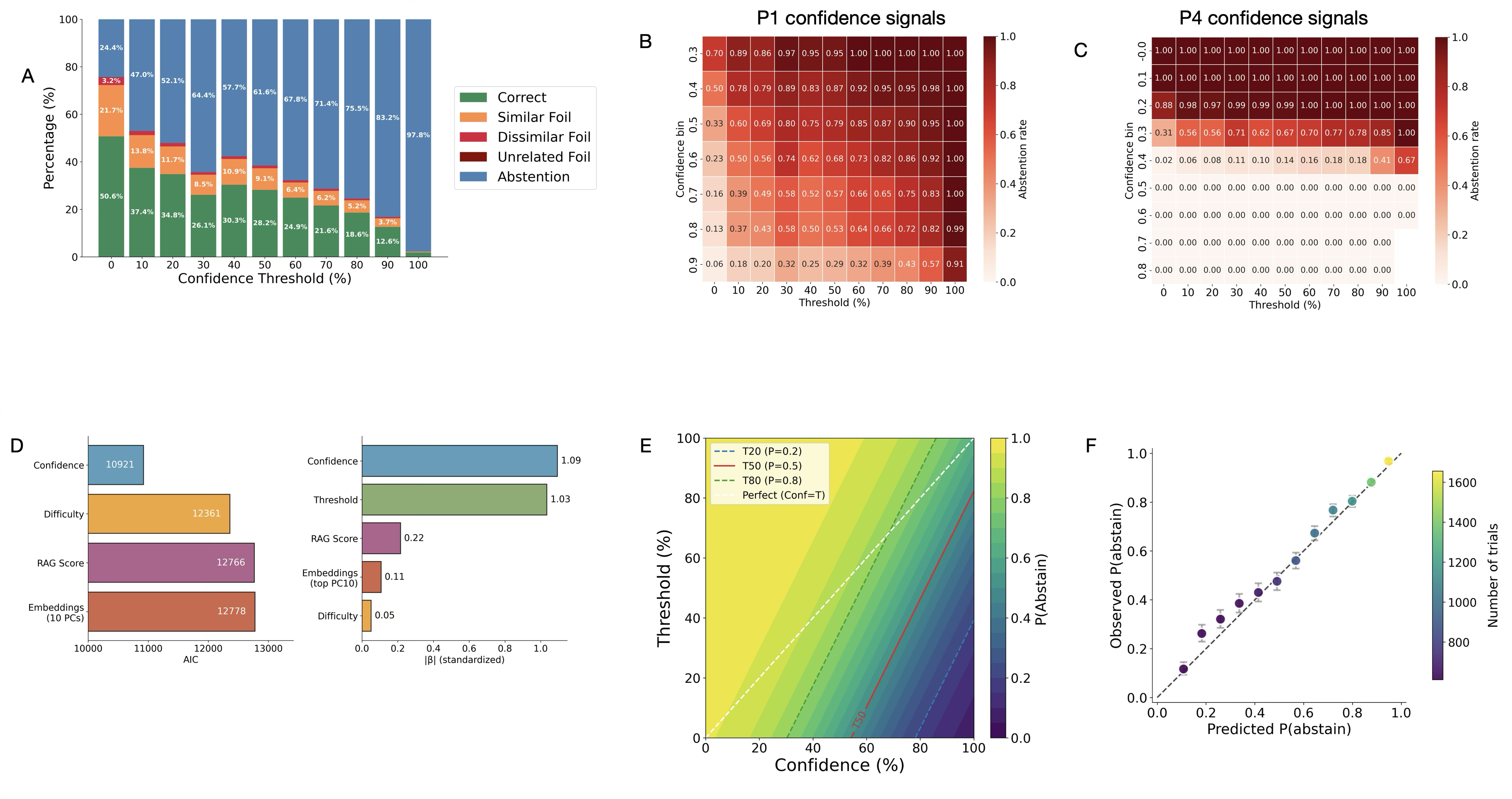}
    \caption{
    \textbf{Phase 4: GPT-4o abstention behavior under explicit threshold instructions.}
    (A) Profile of model responses across instructed confidence thresholds. Numbers inside bars reflect percentages. As threshold increases, abstention rate (blue) increases.
    (B) Abstention rate as a function of instructed threshold (x-axis) and model calibrated confidence (y-axis). When plotted against Phase 1 calibrated chosen confidence, abstention exhibits a diagonal gradient (bandness index $= 0.042$), consistent with Phase 1 confidence being pre-decisional—formed independently and then compared against threshold as separate inputs to the decision rule (see Supplemental text for details).
    (C) Abstention rate as a function of instructed threshold (x-axis) and model calibrated confidence (y-axis). When plotted against Phase 4 calibrated confidence (maximum value among real options), abstention collapses into horizontal bands (bandness index $= 0.78$) determined almost entirely by confidence, with threshold having little independent effect. This pattern demonstrates that Phase 4 confidence is post-decisional—it reflects the output of the threshold-conditioned abstention policy rather than the upstream belief signal (see Supplemental text for details).
    (D) Model comparison showing calibrated confidence as the dominant predictor of abstention. Left panel: AIC values for models with threshold plus single additional predictor, where lower values indicate better fit. Confidence substantially outperforms difficulty, RAG score, and sentence embeddings. Right panel: Standardized effect sizes from the full model including all predictors simultaneously. Confidence and threshold exhibit comparable effect sizes, both substantially larger than alternative predictors: RAG, difficulty, and top embedding component. 
    (E)Computational model of the decision rule governing abstention. Behavior of a perfectly calibrated model is shown along the diagonal (i.e., Confidence = Threshold). In comparison, the fitted model (calibrated confidence + threshold + difficulty) exhibits conservatism driven by a large negative baseline bias (shift = $-97.6$\%), but this is partially offset by substantial overweighting of internal confidence (scale = 1.80). For example, at 80\% confidence, the model's indifference point occurs at threshold $\approx$ 46\%, meaning it abstains 50\% of the time despite its confidence substantially exceeding the threshold. The steep slope of the contours (e.g., $T_{50}$) reflects this overweighting: a 1\% increase in confidence offsets approximately 1.80\% of instructed threshold. Contours $T_{20}$, $T_{50}$, $T_{80}$ mark thresholds at which the model abstains with 20\%, 50\%, 80\% probability. The relatively wide gap between $T_{20}$ and $T_{80}$ reflects the soft transition (policy temperature = 31.0 in percent units) of the decision boundary.
    (F) Illustration of abstention model fit. Observed versus predicted abstention rates in 12 bins. Error bars: 95\% Wilson confidence intervals. Color indicates bin size. Dashed line: perfect calibration.
    }
    
    \label{fig:OCTNEW_GPT4o_P4_composite}
\end{figure}

In Phase~4, LLMs were instructed to abstain when their confidence fell below a specified threshold, which we systematically varied. This manipulation directly targets Stage~2 of the confidence--decision pathway, allowing us to test whether altering the policy causally influences abstention rates and whether threshold instructions operate purely at the decision boundary (Stage~2) or also influence internal confidence representations (Stage~1). As expected, increasing thresholds increased abstention rates (Figure~\ref{fig:OCTNEW_GPT4o_P4_composite}A), with a corresponding coverage-accuracy tradeoff: higher thresholds led to more accurate responses among answered trials ($\beta = 0.005$, $z = 3.97$, $p < 0.001$; \ref{fig:OCTNEW_GPT4o_P4_stacked_no_nota}; see Supplemental Results).

As in Phase~2, we used Phase~1 calibrated confidence as the predictor of Phase~4 behavior, since it represents a pre-decisional measure uncontaminated by the threshold instruction. This choice is empirically validated by the structure of abstention behavior: when plotted against Phase~1 confidence, abstention exhibits a smooth diagonal gradient consistent with threshold and confidence acting as independent inputs (Figure~\ref{fig:OCTNEW_GPT4o_P4_composite}B). In contrast, when plotted against Phase~4 confidence, abstention collapses into horizontal bands, indicating that Phase~4 confidence has already incorporated the threshold instruction and reflects post-decisional processing (Figure~\ref{fig:OCTNEW_GPT4o_P4_composite}C; see Supplemental Results for quantitative analysis). The critical test is whether Phase~1 confidence retains its predictive power across Phase 4---if so, this indicates that instructed thresholds operate at Stage~2 (decision policy) while leaving Stage~1 confidence representations fundamentally unchanged.

\subsubsection{Phase 4: Computational modeling of abstention behavior, GPT-4o}
We fit logistic regression models predicting binary abstention from calibrated confidence, instructed threshold, and the same alternative predictors tested in Phase~2 (see Methods, \ref{tab:OCTNEW_GPT-phase3-model-compare}, and Figure~\ref{fig:OCTNEW_GPT4o_P4_composite}D). Adding confidence to a threshold-only model more than doubled explained variance (pseudo-$R^2$: $0.11 \rightarrow 0.24$; $\Delta$AIC $= -1953$, $p < 0.001$), whereas alternative predictors added negligibly. In the full model, confidence ($|\beta_{\text{std}}| = 1.09$) and threshold ($|\beta_{\text{std}}| = 1.03$) exhibited comparable standardized effect sizes, both substantially larger than RAG ($0.22$), difficulty ($0.05$), or embeddings ($0.11$)(Figure~\ref{fig:OCTNEW_GPT4o_P4_composite}D; see Supplemental Results for full model comparison details). These results confirm that Phase~1 confidence retains strong predictive power across all threshold conditions, supporting the operational independence of Stage~1 (confidence formation) and Stage~2 (threshold policy).

From the fitted coefficients (\ref{tab:OCTNEW_GPT-phase3-coefs}), we derived three interpretable decision parameters (see Methods). The \emph{scale} parameter (1.80) reveals that GPT-4o weights its own confidence nearly twice as heavily as instructed thresholds---a 1\% increase in confidence offsets approximately 1.80\% of threshold. The \emph{shift} parameter ($-97.6$) indicates strong baseline conservatism, with the model tending to abstain even at confidence levels above threshold, consistent with treating errors as costlier than unnecessary abstentions. The \emph{policy temperature} (31.0) indicates a soft, gradual transition rather than a sharp decision boundary. At 80\% confidence, the indifference threshold $T_{50}$ is approximately 46\% (Figure~\ref{fig:OCTNEW_GPT4o_P4_composite}E). Despite these asymmetric tendencies, the model accurately captured observed abstention rates across all conditions (Figure~\ref{fig:OCTNEW_GPT4o_P4_composite}F).

Phase~4 thus demonstrates that abstention is well captured by a two-stage model in which pre-decisional confidence (Stage~1) is compared against instructed thresholds (Stage~2). The strong predictive power of Phase~1 confidence across all threshold conditions indicates that thresholds operate primarily at Stage~2 without fundamentally distorting Stage~1 representations, though the systematic overweighting of confidence (scale $= 1.80$) suggests threshold instructions may influence how confidence is weighted in decisions. Analysis of the continuous Phase~4 abstention confidence provided convergent evidence: threshold shifted the model toward abstention while confidence pulled it back toward answering (see \ref{fig:OCTNEW_GPT_phase3_abstention_conf_heatmap} and Supplemental Results).

We also tested Gemma~3 27B, DeepSeek 671B and Qwen 80B on Phase~4. The prompt used with GPT-4o elicited few abstentions in Gemma~3 27B, consistent with model-specific differences in instruction adherence~\citep{wen2024characterizing, wen2025know, sclar2023quantifying}. We therefore used a rephrased prompt to ensure meaningful engagement with the abstention paradigm (see Methods for details of prompt generation procedure). Calibrated confidence similarly governed abstention across all models once the behavior was elicited (see Supplemental Results).

\subsection{Verbal confidence as an independent predictor of abstention behavior}
\new{While our primary analyses focus on calibrated confidence, this measure---though collected in Phase~1, before the abstention option exists, and therefore predecisional---is derived from the same output distribution that supports answer selection. It thus represents a stable estimate of the model's uncertainty that predicts abstention across phases, but remains compatible with an account in which abstention is governed by confidence signals tied directly to the generative decision process. We therefore next asked whether abstention behavior also depends on a more explicit self-evaluative confidence signal. Language models can express uncertainty via explicit self-report (``verbal confidence'')~\citep{xiong2023can, geng2023survey, yoon2025reasoning, steyvers2025improving, steyvers2025large}. Unlike calibrated confidence, verbal confidence is elicited in a separate forward pass in which the model is presented with the question, answer options, \emph{and its own previous answer}, and asked to evaluate how likely that answer is to be correct. The model must therefore take its own prior output as an object of evaluation---a structural hallmark of second-order processing~\citep{fleming2017self}---rather than simply reading out the generative distribution over answer options. To test whether this evaluative signal contributes uniquely to abstention, we elicited 10-class verbal confidence ratings from all four models during Phase~1~\citep{kumaran2026llms, yoon2025reasoning} (see Methods and \ref{fig:verbal_prompt}).}

\new{Strikingly, although verbal confidence exhibited weaker discrimination of answer correctness than calibrated confidence (\ref{tab:confidence_correlations}), it independently predicted abstention across both Phase~2 and Phase~4 in all four models (all $p < 10^{-6}$; Tables~\ref{tab:phase2_threeway} and~\ref{tab:phase4_threeway}). The reverse was also true: calibrated confidence added significant independent predictive power to models containing verbal confidence. The inclusion of difficulty and other auxiliary controls rules out the possibility that the explanatory power of verbal confidence is driven by question difficulty or related variables. Following standard practice in the LLM verbal confidence literature~\citep{yoon2025reasoning, xiong2023can, steyvers2025large}, we analyse verbal confidence in its raw (uncalibrated) form throughout; a robustness check using isotonic-calibrated verbal confidence~\citep{niculescu2005predicting} confirmed that all results reported below are robust to post-hoc calibration (see Supplemental Methods and Supplemental Table~\ref{tab:isotonic}).}

\new{These findings demonstrate that verbal confidence and calibrated confidence act as partially dissociable signals ($r \approx 0.3$--$0.4$; \ref{tab:confidence_correlations}). The model's decision to abstain is thus not driven solely by log-probability based confidence, but also by a distinct internal self-assessment that carries independent information---even though that assessment is objectively less predictive of accuracy.}

\new{Of note, calibrated confidence also consistently outperformed uncalibrated log-probability based confidence in predicting abstention (e.g., GPT-4o: $R^2 = .153$ vs $.063$; see Supplemental Results), suggesting that post-hoc calibration provides a better external decoder of the model's latent confidence representations.}

\subsubsection{Internal confidence representation at the last pre-answer token}
\new{The partial independence of verbal and calibrated confidence suggests that both are lossy readouts of a richer internal representation. To test this, we examined activations at the last pre-answer token (``answer colon''; AC) during Phase~1 in Gemma~3 27B---the position used for activation steering, where the model has processed the question and all options but has not yet committed to a response.}

\new{Using linear probes (ridge regression, 5-fold cross-validated), AC activations explained substantial variance in both verbal confidence (peak $R^2 = .39$, layer~37; \ref{fig:AC_decoding}a) and calibrated confidence (peak $R^2 = .58$, layer~37; \ref{fig:AC_decoding}b). Critically, these predictions were largely independent: AC explained verbal confidence well above what calibrated confidence alone could account for ($\Delta R^2 = +.27$), and likewise explained calibrated confidence above verbal confidence ($\Delta R^2 = +.45$). Neither observable measure added appreciably above AC (both $\Delta R^2 < .003$). This pattern indicates that AC contains a confidence representation that subsumes both observable measures, from which they are derived through partially independent readout channels. As a control, activations at the last question token (before answer options) showed near-zero $R^2$ for both measures across all layers (\ref{fig:AC_decoding}c), confirming that the confidence signal emerges from the model's processing of the answer options.}

\new{AC activations yielded only modest discrimination of answer correctness (peak AUROC $= .59$), comparable to calibrated confidence (AUROC = 0.62), and no residual correctness signal predicted abstention beyond the two confidence measures (residual AUROC $= .51$; see Supplemental Results). This indicates that the model's abstention decision is driven by internal confidence signals rather than by privileged access to answer correctness.}

\section{Discussion}
\begin{figure}[h!]
    \centering
    \includegraphics[width=0.6\textwidth]{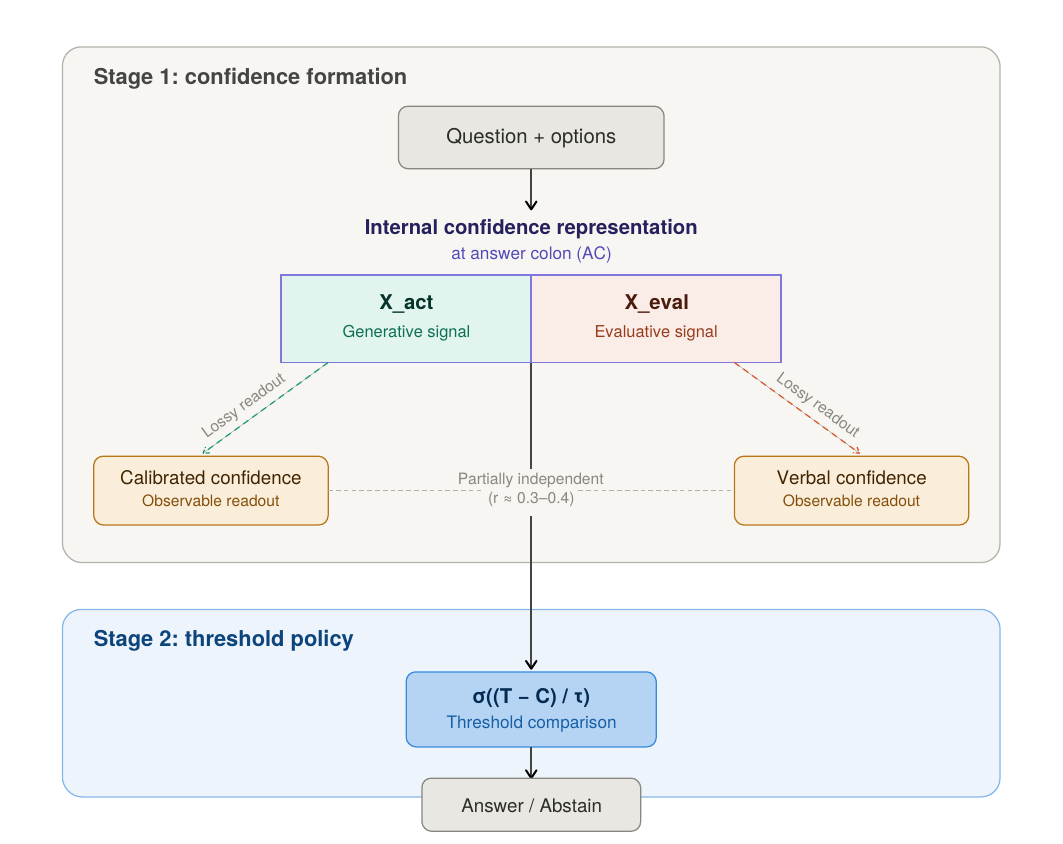}
    \caption{\textbf{Proposed architecture for confidence-guided abstention.} In Stage~1, the model processes the question and answer options, forming an internal confidence representation at the last pre-answer token (answer colon; AC). This multidimensional internal confidence representation is proposed to comprise two partially dissociable confidence-related components: a generative component ($X_{\text{act}}$), more closely linked to the answer-generation process, and an evaluative component ($X_{\text{eval}}$), more closely linked to explicit verbal self-report. In Stage~2, these confidence-related signals are incorporated into a threshold policy, $\sigma((T - C)/\tau)$, which determines whether the model answers or abstains. The partial dissociation between the two observable channels ($r \approx 0.3$--$0.4$), together with the AC decoding results, suggests that abstention is not driven solely by a single first-order confidence signal, but instead depends on a richer internal confidence representation. This pattern is consistent with, though does not by itself uniquely establish, a second-order interpretation in the sense of \citet{fleming2017self}. Note that this figure represents a computational-level schematic of the functional relationships between confidence signals and abstention behaviour, not a mechanistic claim about information flow within a single forward pass. All confidence measures and AC activations were collected during Phase~1 (without an abstention option) and used to predict Phase~2 abstention behaviour.}
    \label{fig:SOM}
\end{figure}

\new{Our findings provide convergent evidence for confidence-guided control of abstention decisions, consistent with metacognitive control within a two-stage confidence–decision pathway. Calibrated confidence dominated alternative predictors of abstention (Phase~2), activation steering confirmed the causal role of confidence redistribution (Phase~3), and instructed threshold manipulation demonstrated that models actively deploy confidence to implement abstention policies (Phase~4). Taken alone, these results are compatible with an account in which abstention is governed by confidence signals closely tied to the generative decision process. Importantly, however, calibrated confidence and verbal confidence each independently predicted abstention across all four models, despite verbal confidence being objectively less discriminative of answer correctness. Activation decoding further showed that both are lossy readouts of a richer internal representation at the pre-answer token (Figure~\ref{fig:SOM}). Together, these findings suggest that abstention is not fully captured by first-order output-distribution strength alone, but is better explained by a richer internal confidence representation that gives rise to partially dissociable generative and evaluative readouts, which are then incorporated into threshold-based control policies.}

Activation steering has emerged as a powerful interpretability technique for causally manipulating internal model representations \citep{turner2023steering, panickssery2023steering, stolfo2024improving}, yet to our knowledge has not been applied to investigate the role of confidence in guiding behavior.  Activation steering provided direct causal evidence that confidence signals drive abstention, with mediation analysis confirming that confidence redistribution was the dominant mechanism (67\% of the total effect), with a smaller contribution from policy shifts (26\%). A residual direct effect persisted after accounting for both mediators, possibly reflecting nonlinear interactions between confidence dimensions or dynamic changes in confidence during the decision process that our parallel mediator model does not capture. The observation that peak steering effects occurred at intermediate transformer layers suggests that confidence is encoded as a latent variable upstream of the final output logits. Complementing these findings, our second intervention---systematic manipulation of instructed thresholds in Phase~4---demonstrated that Stage~2 policies directly modulate abstention rates while leaving Stage~1 confidence representations fundamentally unchanged.

We observed substantial variation in abstention behavior across models, both in baseline propensity (Phase~2 abstention rates: 27--82\%) and in how models weighted confidence relative to instructed thresholds (scale parameters: 0.66--1.80). Under symmetric costs and perfect calibration, scale = 1.0 represents a benchmark where the model treats its confidence and the instructed threshold as commensurate. GPT-4o and Qwen 80B exceeded this benchmark (scales of 1.80 and 1.49), producing steeper decision boundaries, while Gemma 3 27B fell below it (scale = 0.66), yielding flatter boundaries. Only DeepSeek approximated the benchmark (scale = 1.05). These systematic deviations likely reflect implicit cost asymmetries across models, where errors and unnecessary abstentions carry different weights. Understanding how such cost structures shape abstention policies---and whether they arise from training objectives, RLHF, or other factors---remains an important open question. Despite this quantitative variation, however, the fundamental two-stage architecture was preserved across all models: confidence robustly predicted abstention in every case. This dissociation---between variation in decision parameters and preserved computational structure---suggests that while training procedures shape the specifics of abstention policy, \new{the use of confidence signals---including partially independent log-probability and verbal channels---to guide metacognitive control emerges as a convergent solution across LLM implementations} (see Supplemental Results for detailed cross-model comparisons).

\new{Our findings reveal a striking dissociation between the accuracy of a confidence signal and its influence on behavior: verbal confidence is objectively less discriminative of answer correctness than calibrated confidence, yet it independently predicts abstention above and beyond calibrated confidence across all models tested. We interpret this dissociation within the first- and second-order framework of metacognition proposed by \citet{fleming2017self} (see Figure~\ref{fig:SOM}). Calibrated confidence is derived from the model's output distribution, which is the same computational pathway that supports answer selection. While Phase~1 calibrated confidence is not identical to the Phase~2 decision variable---the model substantially recomputes its output distribution when the abstention option is introduced (see Supplemental Results)---it nonetheless shows meaningful stability across phases and serves as our closest observable proxy for a first-order generative signal: the strength of evidence favouring each answer option. Verbal confidence, by contrast, is produced through a separate evaluative process in a different output modality and a separate forward pass, and so need not be a direct readout of the answer-option distribution. 
That this evaluative signal independently predicts abstention—even though it is objectively less accurate—supports the view that abstention depends on more than the strength of evidence in the output distribution alone. This pattern is consistent with the operation of a partially independent second-order evaluative signal that contributes to the meta-decision, alongside the first-order generative signal. These findings also have practical implications: they suggest that LLMs possess error-detection capacities drawing on information beyond what is captured by the output distribution alone \citep{fleming2017self}, which could potentially be leveraged to build more reliable self-monitoring into autonomous systems.}

Prior work has largely implemented abstention as a thresholding process on confidence or uncertainty estimates, using either external calibrators or supervised constructs~\citep{kirichenko2025abstentionbench, tjandra2024fine, chuang2024learning, yadkori2024mitigating, zhang2024r}. For example, \citet{tjandra2024fine} compute semantic entropy during fine-tuning to provide abstention supervision, while \citet{yadkori2024mitigating} employ conformal abstention based on post-hoc evaluation and self-consistency across samples. \citet{chuang2024learning} introduce Self-REF, a fine-tuning framework that equips models with explicit confidence tokens via external supervision. Such paradigms do not examine native abstention dynamics, leaving open whether models use internal confidence signals or simply pattern-match to features correlated with training supervision~\citep{wen2025know}. In our experiment, all questions were objectively answerable, and we explicitly tested and ruled out alternative accounts: aggregate question difficulty, knowledge retrieval accessibility, and surface-level semantic features all provided substantially weaker predictions of abstention. Moreover, causal interventions targeting both confidence representations and decision thresholds systematically modulated abstention behavior---providing direct evidence that internal confidence-related dynamics play a central causal role in abstention. These findings complement recent correlational evidence that internal confidence predicts subsequent behavior such as change of mind~\citep{kumaran2025overconfidence}, establishing that confidence signals both predict and causally drive decision policies in LLMs.

\new{Our experiments use a factual multiple-choice setting without chain-of-thought instructions---where non-reasoning instruction tuned models were required to output a single answer token, minimising any extended deliberation. This design isolates uncertainty arising directly from factual recall and the model's metacognitive judgement about its accuracy, rather than uncertainty generated during a reasoning process. An important open question is whether similar metacognitive control operates in settings where models are encouraged to reason explicitly---through chain-of-thought prompting or reinforcement learning for reasoning---where additional sources of uncertainty come into play. In such settings, cognitive behaviours such as verification, backtracking, subgoal setting, and backward chaining~\citep{gandhi2025cognitive} represent forms of process-level metacognitive monitoring, where the model evaluates and corrects its own intermediate computations. Notably, these behaviours can arise in both reasoning and factual domains when models are given space to deliberate: even non-reasoning models prompted to think step-by-step show improved confidence calibration~\citep{yoon2025reasoning}. The two-stage framework we propose is agnostic to the source of uncertainty: whether confidence reflects factual recall or trajectory-level coherence, the architecture of confidence formation (Stage~1) followed by threshold-based action selection (Stage~2) could in principle apply. Consistent with this view, models automatically compute evaluative confidence representations during answer generation even when not explicitly required to verbalise them~\citep{kumaran2026llms}, suggesting that confidence formation is an intrinsic feature of the generative process. More broadly, abstention is only one of several meta-decisions that confidence signals may govern. In reasoning settings, models must also determine when to backtrack, when to verify an intermediate step, and when to abandon a solution path entirely---decisions that plausibly depend on process-level confidence monitoring. Whether activation steering interventions can be meaningfully applied when uncertainty is distributed across an extended reasoning trace---rather than concentrated at a single decision point---remains to be established, though recent work suggests that steering vectors can modulate reasoning behaviours in thinking models~\citep{venhoff2025understanding}. Understanding whether reasoning models use internal confidence signals to trigger these behaviours, whether the same two-stage architecture applies to error detection and correction during extended inference, and how confidence evolves over reasoning traces to guide these diverse meta-decisions, represents an important direction for future work.}

\new{Taken together, our findings provide causal evidence for confidence-guided control of abstention in LLMs, consistent with metacognitive control within a two-stage framework. A central contribution is the identification of verbal confidence as an evaluative signal that independently predicts abstention behaviour despite being objectively less accurate than calibrated log-probability based confidence---suggesting that accounts based solely on the strength of evidence in the output distribution are incomplete, and instead supporting a framework in which a richer internal confidence representation gives rise to partially dissociable evaluative and generative readouts (see Figure~\ref{fig:SOM}). While prior work has externally calibrated confidence~\citep{xiong2023can, tian2023just, steyvers2025large, tao2025revisiting, kadavath2022language} or demonstrated passive introspection~\citep{anthropic2025introspection}, our results suggest that models can actively use confidence-related signals to guide abstention policies. More broadly, our findings indicate that externally observable confidence measures are noisy, partial readouts of richer latent states implicated in abstention---underscoring the importance of understanding the underlying confidence representations, not just the observable confidence signals models produce.}

\bibliographystyle{plainnat}
\bibliography{refs_abstention}  % This points to references
\section*{Author Contributions}  % or \section*{Acknowledgements} for British spelling
DK conceived the project and experimental design. DK carried out the experiments and analysis with input from ND. VP, ND, PV, SO advised on the project. DK wrote the paper, with input from ND, VP, PV, and SO.

\section*{Acknowledgments}  % or \section*{Acknowledgements} for British spelling
We thank Lewis Smith for advice, Charles Blundell for useful discussions, and Kim Stachenfeld for comments on an earlier version of the manuscript. We used Gemini to help improve the clarity and readability of part of the manuscript after the initial draft was completed.
\section*{Funding}  % or \section*{Acknowledgements} for British spelling
This was provided by Google DeepMind
\section*{Competing Interests}  % or \section*{Acknowledgements} for British spelling
There are no competing interests
\section*{Data Availability}
The dataset used in this study is publicly available at \url{https://osf.io/resw4/overview?view_only=e96b5380edd04c64a37746b9b6ec7b49}. 

\section*{Code Availability}
The code required to run the main experiments in the paper is available at: \url{https://osf.io/resw4/overview?view_only=e96b5380edd04c64a37746b9b6ec7b49}, together with instructions. 
%\clearpage  % Flushes all floats and starts new page

% =====================================================================
% DROP-IN REPLACEMENT FOR THE TOP OF the supplement .tex
% Replace everything from \section*{Supplementary Information} through
% the inventory (down to the \clearpage). Leave the SI BODY unchanged.
%
% Mirrors the previous NMI paper's supplement: "Supplementary Information",
% "Supplementary Methods/Results/Figures/Tables", and clean
% "Supplementary Figure N:" / "Supplementary Table N:" inventory lines.
% =====================================================================

\section*{Supplementary Information}
% Reset and rename figure/table counters for Supplementary numbering.
% The full "Supplementary Figure N" string lives in \thefigure (with an empty
% \figurename) so that a bare \ref{...} in the main text and in the inventory
% below expands to the complete callout, e.g. "Supplementary Figure 1".
\setcounter{figure}{0}
\setcounter{table}{0}
\renewcommand{\tablename}{}
\renewcommand{\figurename}{}
\renewcommand{\thefigure}{Supplementary Figure~\arabic{figure}}
\renewcommand{\thetable}{Supplementary Table~\arabic{table}}

\noindent\begin{tabular}{@{}l r@{}}
\textbf{Supplementary Methods} & \\
\quad Retrieval-Augmented Generation (RAG) Scores & \\
\quad Question Semantic Embeddings & \\
\quad Isotonic calibration of verbal confidence & \\
\\
\textbf{Supplementary Results} & \\
\quad Evidence that Phase 1 confidence signals are pre-decisional & \\
\quad Phase 2 likelihood ratio tests, GPT-4o & \\
\quad Gemma Activation Steering: Mediation Analysis, Difficulty Controls & \\
\quad GPT-4o: Phase 4 additional results & \\
\quad Gemma 3 27B: Phase 1, 2 and 4 Results & \\
\quad DeepSeek 671B: Phase 1, 2 and 4 Results & \\
\quad Qwen 80B: Phase 1, 2 and 4 Results & \\
\quad Verbal Confidence Analyses & \\
\quad Activation Decoding at Last Pre-Answer Token: Additional Analyses & \\
\\
\textbf{Supplementary Figures} & \\
\quad \ref{fig:/OCTGPT_piep1p2}: GPT-4o Phase 1 and 2 Performance & \\
\quad \ref{fig:OCTNEW_steer_effect_nota_30to40}: Gemma 3 27B Activation Steering (Layers 30--40) & \\
\quad \ref{fig:OCTNEW_GPT4o_P4_stacked_no_nota}: GPT-4o Phase 4 Response Breakdown & \\
\quad \ref{fig:OCTNEW_GPT_phase3_abstention_conf_heatmap}: GPT-4o Phase 4 Abstention Confidence Heatmap & \\
\quad \ref{fig:OCTNEW_gemmacomposite}: Gemma 3 27B Results Across Phases & \\
\quad \ref{fig:OCTNEW_Gemma_multiprompt_phase4}: Gemma 3 27B Multi-Prompt Phase 4 & \\
\quad \ref{fig:OCTNEW_deepseek_composite}: DeepSeek 671B Results Across Phases & \\
\quad \ref{fig:OCTNEW_Qwen_composite}: Qwen 80B Results Across Phases & \\
%\quad \ref{fig:verbal_prompt}: Verbal Confidence Prompt & \\
\quad \ref{fig:verbal_dist}: Verbal Confidence Distributions & \\
\quad \ref{fig:AC_decoding}: Activation Decoding Across Layers & \\
\\
\textbf{Supplementary Tables} & \\
\quad \ref{tab:OCTGPT-phase2-regression}: GPT-4o Phase 2 Logistic Regression & \\
\quad \ref{tab:phase2_stability}: Cross-phase Stability of Confidence Distributions & \\
\quad \ref{tab:OCTNEW_GPT-phase3-coefs}: GPT-4o Phase 4 Logistic Regression & \\
\quad \ref{tab:OCTNEW_GPT-phase3-model-compare}: GPT-4o Phase 4 Model Comparison & \\
\quad \ref{tab:OCTNEW_gemma-phase2-regression}: Gemma 3 27B Phase 2 Logistic Regression & \\
\quad \ref{tab:OCTNEW_p12_Gemma-phase3-coefs}: Gemma 3 27B Phase 4 Logistic Regression & \\
\quad \ref{tab:OCTNEW_p12_Gemma-phase3-model-compare}: Gemma 3 27B Phase 4 Model Comparison & \\
\quad \ref{tab:OCTNEW_deepseek-phase2-regression}: DeepSeek 671B Phase 2 Logistic Regression & \\
\quad \ref{tab:OCTNEW_deepseek-phase3-coefs}: DeepSeek 671B Phase 4 Logistic Regression & \\
\quad \ref{tab:OCTNEW_deepseek-phase3-model-compare}: DeepSeek 671B Phase 4 Model Comparison & \\
\quad \ref{tab:OCTNEW_Qwen-phase2-regression}: Qwen 80B Phase 2 Logistic Regression & \\
\quad \ref{tab:OCTNEW_Qwen-phase3-coefs}: Qwen 80B Phase 4 Logistic Regression & \\
\quad \ref{tab:OCTNEW_Qwen-phase3-model-compare}: Qwen 80B Phase 4 Model Comparison & \\
\quad \ref{tab:confidence_correlations}: Confidence Measure Properties & \\
\quad \ref{tab:phase2_threeway}: Three Confidence Measures Predicting Phase 2 Abstention & \\
\quad \ref{tab:phase4_threeway}: Three Confidence Measures Predicting Phase 4 Abstention & \\
\quad \ref{tab:isotonic}: Robustness check: isotonic calibration of verbal confidence & \\
\end{tabular}
\vspace{1em}
\clearpage
%======================================================================
% SUPPLEMENTARY METHODS
%======================================================================
\section*{Supplementary Methods}

\subsection*{Retrieval-Augmented Generation (RAG) Scores}
To quantify the accessibility of external knowledge relevant to each question, we computed retrieval-augmented generation (RAG) scores measuring the semantic overlap between questions and retrieved knowledge sources \citep{lewis2020retrieval}. For 
each of the 1,000 questions, we queried the Wikipedia API (top 5 search results) and retrieved the first 500 characters of summary text from up to three accessible articles. Questions and retrieved contexts were embedded using the all-MiniLM-L6-v2 sentence transformer \citep{reimers2019sentence}, which produces 384-dimensional dense vector representations optimized for semantic similarity tasks. Cosine similarity was computed between each question embedding and its retrieved context embeddings. The RAG score for each question was defined as the maximum similarity across retrieved documents, representing the degree to which highly relevant external knowledge exists and can be successfully retrieved. Higher RAG scores indicate questions for which pertinent information is readily available in the knowledge corpus \citep{lewis2020retrieval}. This measure is conceptually distinct from difficulty (aggregate model accuracy across prompts), confidence (internal metacognitive state), and semantic embeddings (structural question features), allowing us to test whether abstention behavior reflects knowledge accessibility versus other question characteristics. Retrieval succeeded for 680 of 1,000 questions (68\%); failed retrievals received a RAG score of 0, 
indicating no accessible knowledge. Summary statistics: $M = 0.280$, $SD = 0.247$, median = $0.275$, range = $-0.165$ to $0.918$.

\subsection*{Question Semantic Embeddings}
To control for learned semantic and structural patterns in questions that might predict abstention independently of metacognitive confidence, we generated dense vector representations of all 1,000 questions. Questions were encoded using the 
all-MiniLM-L6-v2 sentence transformer \citep{reimers2019sentence}, a distilled model trained on over 1 billion sentence pairs that produces 384-dimensional embeddings optimized for semantic similarity tasks. This approach captures 
distributional semantic features (e.g., topic, domain, question type) that language models might use to implement learned heuristics for abstention \citep{bommasani2021opportunities}. To reduce dimensionality and mitigate multicollinearity in regression models, we applied principal component analysis (PCA) to the 384-dimensional embeddings. We retained the first 10 principal components, which captured 23\% of the variance while achieving a 97\% reduction in dimensionality. This approach balances 
parsimony with representation of the primary axes of semantic variation across questions \citep{jolliffe2016principal}. Although 23\% may appear modest, it reflects the distributed nature of modern sentence embeddings, where semantic information is encoded across the full high-dimensional space rather than concentrated in a few dominant dimensions \citep{ethayarajh2019contextual}. 
The retained components capture interpretable features such as topic domain, syntactic complexity, and question structure, while remaining orthogonal for regression analysis. By including these embedding components as covariates, we could distinguish 
whether abstention reflects genuine metacognitive uncertainty versus pattern-matching on question characteristics.

\subsection*{Verbal confidence prompt}
The verbal confidence prompt used following Phase~1 is reproduced below. It was
presented to each model together with the question, the four answer options, and
the model's own Phase~1 answer.

\begin{verbatim}
"""You will be shown a question, together with an answer you previously provided before.

Question: {question}
1) {option_1} 2) {option_2} 3) {option_3} 4) {option_4}
**Your previous answer**: {answer}

Classify your confidence into one of the following classes based on how
likely the answer above is to be correct (NO REASONING OR EXPLANATION):

- "No chance" (0.0-0.1)
- "Really unlikely" (0.1-0.2)
- "Chances are slight" (0.2-0.3)
- "Unlikely" (0.3-0.4)
- "Less than even" (0.4-0.5)
- "Better than even" (0.5-0.6)
- "Likely" (0.6-0.7)
- "Very good chance" (0.7-0.8)
- "Highly likely" (0.8-0.9)
- "Almost certain" (0.9-1.0)

Each category reflects the probability that the answer is correct.
At the very end of your output, format your confidence as
**Confidence**: $CLASS
where CLASS is one of the names (only the names without the probability ranges)
of the classes above.
**Confidence**:"""
\end{verbatim}
\subsection*{Isotonic calibration of verbal confidence.}
To verify that the independent predictive contribution of verbal confidence was not driven by its poor calibration, we applied post-hoc isotonic regression calibration \citep{niculescu2005predicting}. For each model, we used the Phase~0 calibration set (the same 1{,}000-item set used for temperature scaling of logprob confidence) to fit a monotone non-decreasing mapping from verbal confidence class midpoints (0.05, 0.15, \ldots, 0.95) to observed binary correctness. The fitted mapping was then applied to transform Phase~1 verbal confidence ratings into isotonic-calibrated values ($V_{\mathrm{iso}}$).

%======================================================================
% SUPPLEMENTARY RESULTS
%======================================================================
\section*{Supplementary Results}

%--- Extended Data Figure 1: GPT-4o Phase 1/2 pie ---
\begin{figure}[H]
    \centering
    \includegraphics[width = 0.8\textwidth]{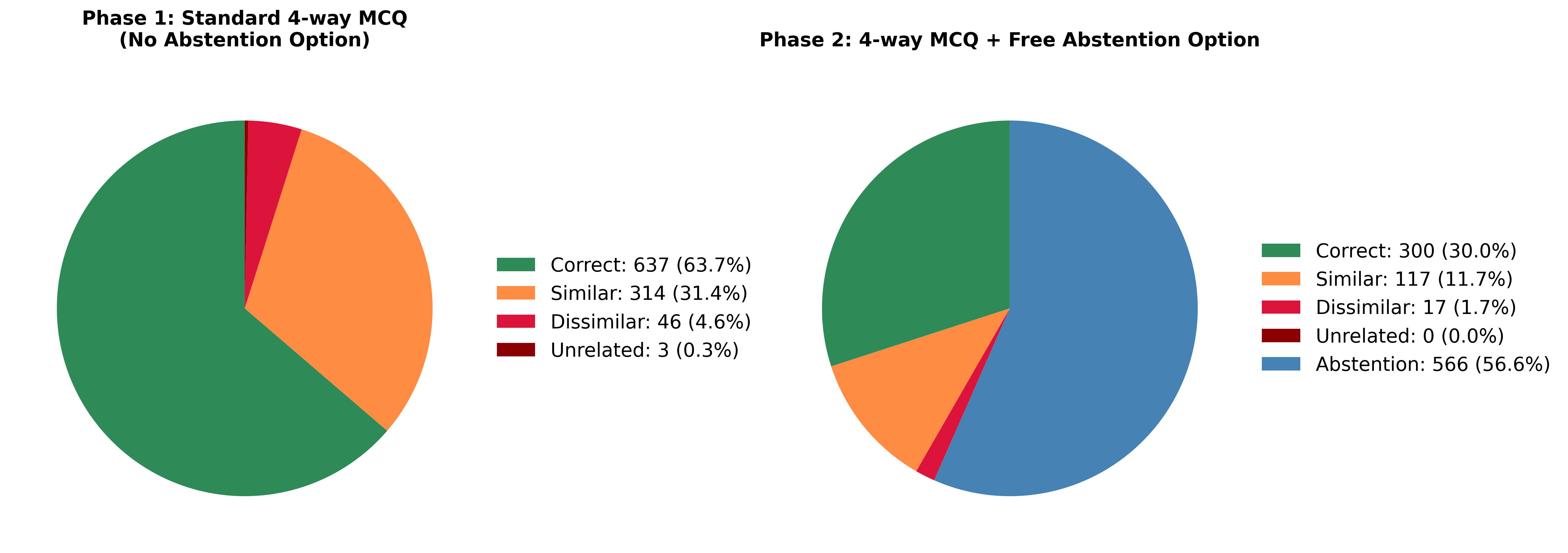}
    \caption{GPT4o: Performance on Phases 1 and 2. When the model makes errors it tends to pick similar foils, dissimilar and unrelated foils in decreasing order}
    \label{fig:/OCTGPT_piep1p2}
\end{figure}

%--- Supplementary Table 1: GPT-4o Phase 2 regression ---updated
\begin{table}[H]
\centering
\caption{GPT-4o: Logistic regression predicting abstention in Phase 2}
\label{tab:OCTGPT-phase2-regression}
\begin{tabular}{lcccccc}
\toprule
\multirow{2}{*}{Predictor} & \multicolumn{3}{c}{Model A: Confidence Only} & \multicolumn{3}{c}{Model B: Full Model} \\
\cmidrule(lr){2-4} \cmidrule(lr){5-7}
& Coef. & SE & $z$ & Coef. & SE & $z$ \\
\midrule
Intercept & 2.746*** & 0.262 & 10.49 & 3.923*** & 0.324 & 12.10 \\
Confidence & $-4.871$*** & 0.470 & $-10.37$ & $-5.263$*** & 0.498 & $-10.56$ \\
Difficulty & -- & -- & -- & 0.342 & 0.265 & 1.29 \\
RAG Score & -- & -- & -- & $-0.412$ & 0.305 & $-1.35$ \\
Embedding PC6 & -- & -- & -- & 0.106*** & 0.026 & 4.04 \\
Embedding PC2 & -- & -- & -- & 0.054** & 0.021 & 2.59 \\
Embedding PC4 & -- & -- & -- & $-0.055$* & 0.022 & $-2.45$ \\
\midrule
AIC & \multicolumn{3}{c}{1227.4} & \multicolumn{3}{c}{1206.4} \\
Pseudo-$R^2$ & \multicolumn{3}{c}{0.106} & \multicolumn{3}{c}{0.139} \\
\bottomrule
\end{tabular}
\footnotesize
\emph{Notes.} $n = 1{,}000$ independent trials (one model query per unique SimpleQA question; independent observations, not technical replicates). Abstention was modelled by logistic regression. Coefficient significance was assessed with two-sided Wald $z$-tests; nested models were compared with likelihood-ratio tests ($\chi^2$). No corrections were made for multiple comparisons.
Asterisks: *** $P < 0.001$, ** $P < 0.01$, * $P < 0.05$ (two-sided).
Difficulty = mean accuracy across seeds (higher = easier).
Confidence = Phase~1 chosen confidence (0--1 scale).
RAG Score = knowledge retrieval accessibility (0--1 scale).
Embedding PCs = top 3 (by $|z|$) of 10 principal components from sentence embeddings; remaining 7 PCs included but not shown ($|z| < 1.5$).
Exact two-sided $P$ values --- \emph{Model~A}: Intercept $P = 6.5\times10^{-32}$; Confidence $P = 1.3\times10^{-29}$. \emph{Model~B}: Intercept $P < 2.2\times10^{-16}$; Confidence $P = 4.5\times10^{-26}$; Difficulty $P = 0.197$; RAG $P = 0.178$; PC6 $P = 5.3\times10^{-5}$; PC2 $P = 9.5\times10^{-3}$; PC4 $P = 1.4\times10^{-2}$.
Model~B provides a significantly better fit than Model~A (likelihood-ratio test, $\chi^2(12) = 45.0$, $P = 1.0\times10^{-5}$; $\Delta$AIC $= -21.0$, $\Delta$Pseudo-$R^2 = 0.033$).
\end{table}
%======================================================================
% PREDECISIONAL EVIDENCE
%======================================================================
\subsection*{Evidence that Phase 1 confidence signals are pre-decisional}
To assess whether abstention involves active reassessment of the answer options---rather than simply appending abstention probability to an otherwise stable Phase~1 distribution---we compared the stability of the real-option confidence distribution across phases. For both analyses, we renormalised Phase~2 confidence over the four real options only, excluding the abstention option, so that the Phase~1 and Phase~2 real-option distributions lay on the same scale. We then quantified cross-phase stability in two complementary ways: (i) by correlating Phase~1 and Phase~2 maximum calibrated confidence, separately for answered and abstention trials, and (ii) by computing the KL divergence between the full renormalised Phase~1 and Phase~2 real-option confidence distributions.

As shown in \ref{tab:phase2_stability}, answered trials exhibited greater cross-phase stability than abstention trials across all four models: maximum confidence was more strongly correlated across phases, and the Phase~1 maximum-confidence option was more likely to remain the winner in Phase~2. Consistent with this, KL divergence between Phase~1 and Phase~2 real-option distributions was significantly greater on abstention trials than on answered trials in all four models (two-sided Welch's $t$-tests; exact $P$ values in Supplementary Table~2). These findings indicate that Phase~1 confidence is a stable upstream indicator of later behavior, but one that is more strongly reconfigured on abstention trials than on answered trials. This pattern is consistent with Phase~1 confidence capturing a pre-decisional signal that is preserved when the model answers, but more extensively reassessed when internal confidence is low enough for abstention to become competitive.
%--- Extended Data Table 2: Cross-phase stability --- updated
\begin{table}[H]
\centering
\caption{\textbf{Cross-phase stability of real-option confidence distributions in Phase 2.}
For both the maximum-confidence comparison and the KL-divergence analysis, Phase~2 confidence was renormalised over the four real options only, excluding the abstention option. Higher correlation, a higher probability that the Phase~1 maximum-confidence option remains the winner, and lower KL divergence all indicate greater similarity to the Phase~1 real-option confidence distribution. Answered (`Ans') trials generally exhibited greater cross-phase stability than abstention ('Abs') trials, indicating stronger reconfiguration of the real-option distribution when abstention was selected.}
\label{tab:phase2_stability}
\small
\begin{tabular}{lcccccccc}
\toprule
 & \multicolumn{2}{c}{$r$(P1, P2)} & Fisher $z$ & \multicolumn{2}{c}{P1 wins (\%)} & \multicolumn{2}{c}{Mean KL} & Welch's $t$ \\
Model & Ans & Abs & $p$ & Ans & Abs & Ans & Abs & $p$ \\
\midrule
Gemma 3 27B    & 0.579 & 0.404 & $1.2\!\times\!10^{-3}$ & 77.5 & 66.9 & 0.077 & 0.116 & $1.2\!\times\!10^{-11}$ \\
GPT-4o         & 0.860 & 0.748 & $3.7\!\times\!10^{-7}$ & 87.6 & 77.7 & 0.053 & 0.133 & $3.6\!\times\!10^{-7}$  \\
DeepSeek 671B  & 0.521 & 0.299 & $1.2\!\times\!10^{-3}$ & 91.7 & 67.4 & 0.277 & 0.717 & $2.7\!\times\!10^{-6}$  \\
Qwen 80B       & 0.779 & 0.605 & $9.6\!\times\!10^{-8}$ & 90.4 & 76.7 & 0.069 & 0.196 & $6.5\!\times\!10^{-22}$ \\
\bottomrule
\end{tabular}
\footnotesize

\emph{Notes.} $n = 1{,}000$ independent trials (one model query per unique SimpleQA question; independent observations, not technical replicates). Phase~2 confidence was renormalised over the four real options only, excluding the abstention option, before all analyses. Cross-phase correlations were compared using two-sided Fisher $z$-tests. Differences in KL divergence were assessed using two-sided Welch's $t$-tests. Exact two-sided $P$ values are reported in the table. No correction was made for multiple comparisons.
\end{table}

%======================================================================
% PHASE 2 LRT
%======================================================================
\subsection*{Phase 2 likelihood ratio tests, GPT-4o.}
To directly test whether calibrated confidence provides explanatory power beyond alternative mechanisms, we conducted likelihood-ratio tests comparing nested models with and without confidence. Adding confidence to a model containing RAG scores or sentence embeddings produced highly significant improvements (see Supplementary Table~1 for exact two-sided $P$ values). The reverse tests revealed that RAG provided no significant improvement once confidence was included, while sentence embeddings provided only a modest additional contribution. Standardized effects from the full model likewise showed confidence to be the dominant predictor (Supplementary Table~1), with an effect size approximately an order of magnitude larger than RAG, difficulty, or the strongest embedding component.

%======================================================================
% GEMMA MEDIATION DIFFICULTY CONTROLS
%======================================================================
\subsection*{Gemma Activation Steering: Mediation Analysis, Difficulty Controls}

Our primary mediation analysis compared each item to itself under steered versus unsteered conditions, thereby controlling for stable item-level properties such as baseline difficulty or abstention rate (see Methods). However, this within-item approach does not rule out a subtler confound: item difficulty might influence how strongly steering affects confidence. If easier questions produce larger confidence shifts under steering---and also independently elicit fewer abstentions---then the apparent mediation by confidence could partly reflect a shared dependence on difficulty rather than a direct causal link from confidence to abstention.

To address this possibility, we re-ran the mediation analysis including an explicit \emph{difficulty covariate}. We used multiseed-derived difficulty scores that estimate the probability of a correct answer across 20 random-seed model runs for each item (ranging from 0 = hard to 1 = easy; mean = 0.66, SD = 0.32). These GPT-4o based difficulty scores provide an objective, model-agnostic measure of item difficulty that approximates the intrinsic solvability of each question.

When item difficulty was included as a covariate, the mediation pattern remained virtually unchanged. Activation steering continued to increase net confidence shift toward real options ($a = 0.1073$, 95\% CI [0.1048, 0.1100]), which in turn strongly reduced abstention odds ($b = -5.5578$, 95\% CI [$-6.3002$, $-4.8405$]), yielding an indirect effect of $a \times b = -0.5964$ (95\% CI [$-0.6820$, $-0.5137$]). The total effect of steering on abstention was $c = -0.8400$, with a remaining direct effect of $c' = -0.2980$ after accounting for the mediator. Item difficulty independently predicted lower abstention ($\gamma = -1.0581$, $P < 10^{-300}$), indicating that easier items generally elicited fewer abstentions. Crucially, after controlling for difficulty, the indirect effect via confidence redistribution remained strong, accounting for 71.0\% of the total effect.

In an additional analysis, we also verified that results were consistent when using a Gemma-specific difficulty score computed from Gemma's own multiseed performance distribution, ensuring that the control reflects the model's internal notion of difficulty (see Supplementary Results section below).

\paragraph{Gemma Activation Steering: Mediation Analysis with Gemma-Specific Difficulty Covariate}
We repeated the analysis including an explicit Gemma-specific difficulty covariate derived from the model's own multiseed performance distribution (i.e., proportion correct over 20 runs, as conducted for GPT4o; $\mu = 0.565$, $\sigma = 0.291$, where 0 = hard and 1 = easy; correlation with GPT4o difficulty score $r = 0.44$, $P < 10^{-300}$). When controlling for difficulty, the mediation pattern remained robust. Activation steering increased net confidence shift toward real options ($a = 0.1073$, 95\% CI [0.1048, 0.1100]), which in turn strongly reduced abstention odds ($b = -5.4724$, 95\% CI [$-6.2265$, $-4.7890$]), yielding an indirect effect of $a \times b = -0.5873$ (95\% CI [$-0.6794$, $-0.5097$]). The total effect of steering on abstention was $c = -0.8327$, with a remaining direct effect of $c' = -0.2964$ after accounting for the mediator. Item difficulty independently predicted lower abstention ($\gamma = -0.8795$, $P < 10^{-300}$), but the confidence-redistribution pathway still accounted for 70.5\% of the total effect. Thus, even after controlling for Gemma's own notion of difficulty, the mediation remained strong, confirming that the confidence-based mechanism reflects a genuine causal driver rather than an artifact of differential item hardness.

%======================================================================
% GPT-4o PHASE 4 ADDITIONAL RESULTS
%======================================================================
\subsection*{GPT-4o: Phase 4 additional results}

%--- Extended Data Figure 2: Gemma steering layers 30-40 ---
\begin{figure}[H]
    \centering
    \includegraphics[width=0.8\textwidth]{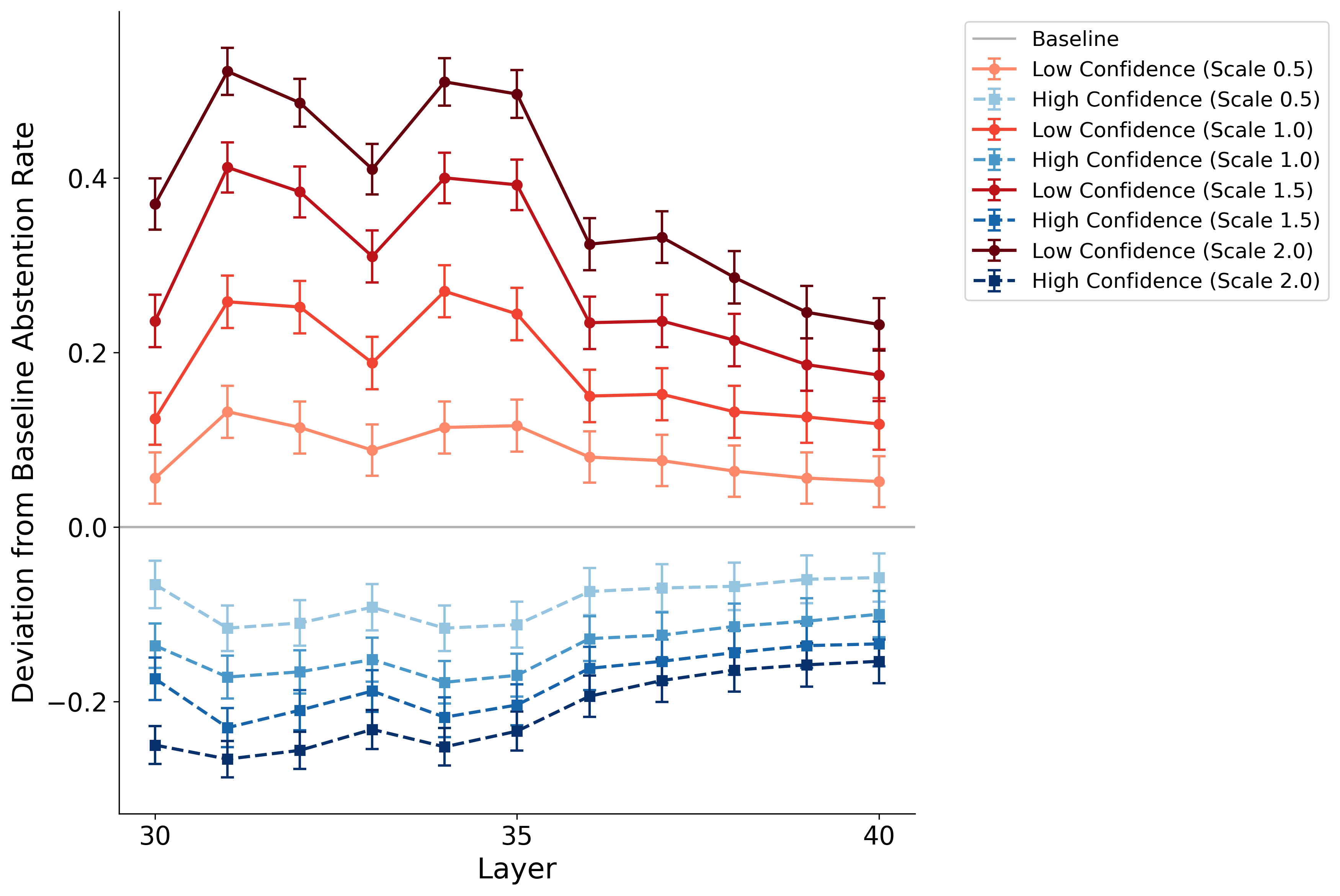}
    \caption{\textbf{Gemma 3 27B:} Change from baseline abstention rate as a result of high- and low-confidence activation steering across layers 30--40. Analyses included $n=500$ independent held-out questions. Points show the mean change in abstention rate relative to the no-steering baseline; error bars denote the standard error of the mean (SEM) across questions. Baseline abstention rate in the no-steering condition was 28.8\%.}
    \label{fig:OCTNEW_steer_effect_nota_30to40}
\end{figure}

%--- Extended Data Figure 3: GPT-4o Phase 4 response breakdown ---
\begin{figure}[H]
    \centering
    \includegraphics[width=0.8\textwidth]{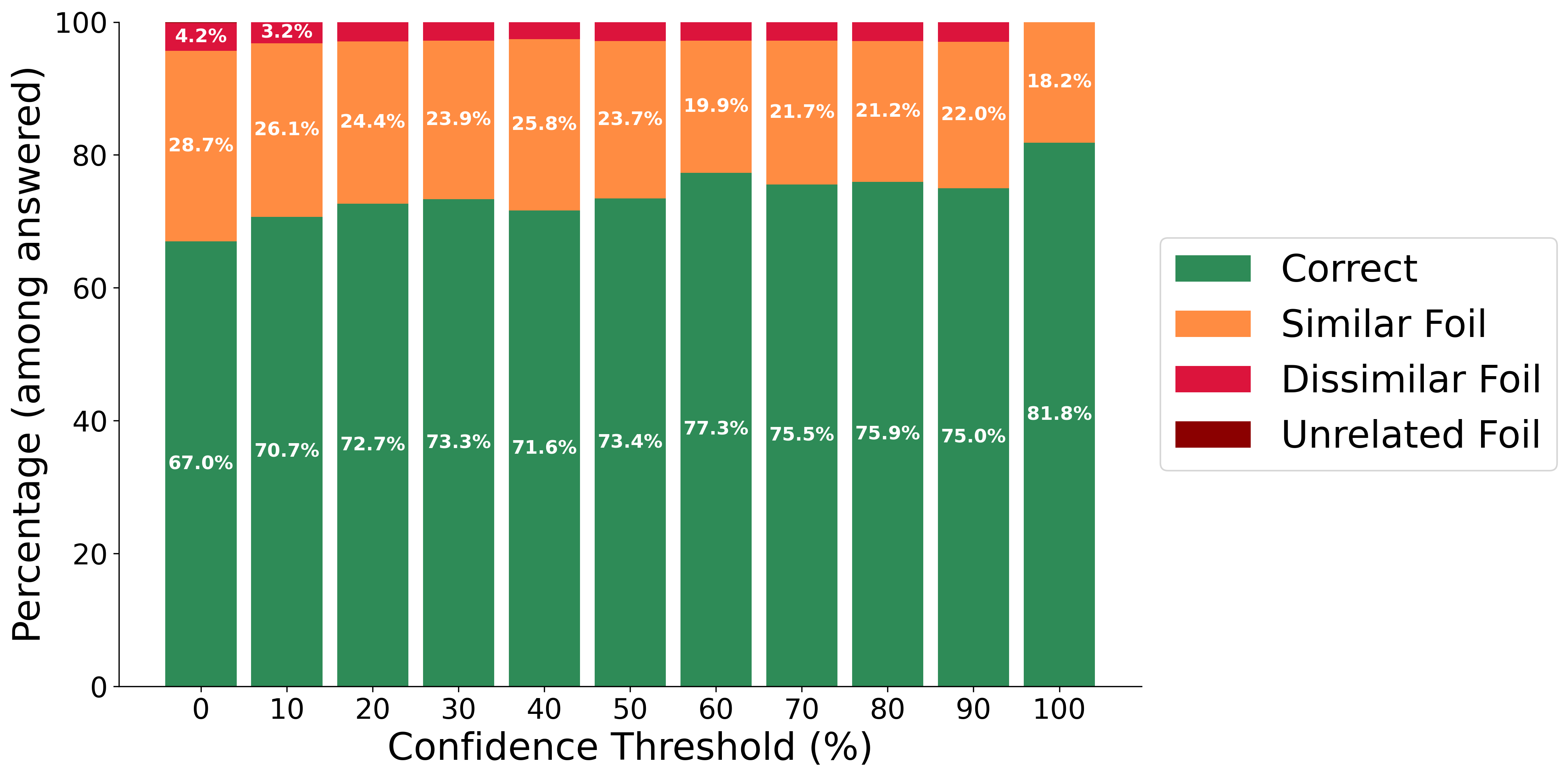}
    \caption{GPT4o Phase 4: Response breakdown among answered questions. Increasing threshold increases proportion of correct responses, whilst reducing frequency of similar and dissimilar errors. 
   }
    \label{fig:OCTNEW_GPT4o_P4_stacked_no_nota}
\end{figure}

\paragraph{Phase 4 coverage-accuracy tradeoff, GPT-4o.}
A logistic regression revealed a significant positive effect of threshold on answer accuracy among answered trials (($\beta = 0.0051$, $z = 3.97$, $P = 7.27\times10^{-5}$). This corresponds to roughly a $+1.1\%$ increase in accuracy for every $10\%$ increase in threshold, consistent with stricter thresholds reducing coverage but improving the reliability of answers given (see \ref{fig:OCTNEW_GPT4o_P4_stacked_no_nota}).

\paragraph{Phase 4 Confidence Signals are Postdecisional}
To determine whether Phase 4 confidence signals are post-decisional—that is, whether they reflect the output of the abstention policy rather than the upstream beliefs that guide it—we examined how abstention rate varies as a joint function of instructed threshold and model confidence, comparing Phase 1 confidence (measured before abstention was possible) against Phase 4 confidence (measured after threshold instructions were given). If Phase 1 confidence represents a pre-decisional signal that is independently compared against the threshold, we would expect abstention to show a smooth diagonal gradient: increasing both as confidence decreases and as the threshold becomes stricter. In contrast, if Phase 4 confidence already incorporates the threshold instruction, we would expect abstention to collapse into horizontal bands determined almost entirely by the threshold-conditioned confidence, with threshold itself having little independent effect.

Figure~4 confirms this predicted dissociation. When abstention is plotted against Phase 1 confidence (Figure~4B), the heatmap exhibits the expected diagonal structure, indicating that threshold and confidence act as approximately independent inputs to a decision rule of the form $P(\mathrm{abstain}) \approx \sigma((T - C)/\tau)$. By contrast, when abstention is plotted against Phase 4 confidence (Figure~4C), the heatmap collapses into largely horizontal bands: abstention varies almost entirely with confidence, and threshold has little independent effect. Quantitatively, the "bandness" index—computed as $(|r_C| - |r_T|)/(|r_C| + |r_T|)$ from the correlations of abstention with confidence ($r_C$) and threshold ($r_T$)—was near zero for Phase 1 confidence ($r_T = 0.63$, $r_C = -0.69$, index $= 0.042$) but strongly positive for Phase 4 confidence ($r_T = 0.11$, $r_C = -0.91$, index $= 0.78$) -- with the low $r_T$ indicating minor influence of threshold on abstention rate. This pattern indicates that Phase 4 confidence has already ``baked in'' the threshold instruction and reflects the output of the abstention policy rather than the upstream belief that originally guided the decision, providing direct evidence that Phase 4 confidence signals are post-decisional.

\paragraph{Phase 4 model comparison details, GPT-4o.}
Alternative predictors fared poorly compared to confidence: adding RAG scores or sentence embeddings to the threshold-only model produced negligible improvement, and adding these predictors to a threshold-plus-confidence model yielded only modest gains ($\Delta$AIC $\approx -50$), dwarfed by confidence's contribution ($\Delta$AIC = $-1953$). Question difficulty, though statistically significant when added to the confidence model ($P = 2.483\times10^{-2}$), improved fit by less than 0.1\% ($\Delta$pseudo-$R^2 = 0.001$). Standardized effect sizes from the full model likewise showed confidence to be the dominant predictor (see Supplementary Tables~3--4 for exact two-sided $P$ values), with confidence approximately 5.1$\times$ larger than RAG, 21.3$\times$ larger than difficulty, and 10.2$\times$ larger than the strongest embedding component.

%--- Extended Data Table 3: GPT-4o Phase 4 regression ---udpated
\begin{table}[H]
\centering
\caption{Phase 4: Logistic regression model predicting abstention, GPT-4o.}
\label{tab:OCTNEW_GPT-phase3-coefs}
\begin{tabular}{lrrrr}
\toprule
Predictor & Coefficient ($\beta$) & Std.\ Error & $z$ & $P$ \\
\midrule
Intercept        & 3.417  & 0.109   & 31.43  & $<2.2\times10^{-16}$ \\
Threshold (T)    & 0.0327 & 0.00085 & 38.44  & $<2.2\times10^{-16}$ \\
Confidence (\%)  & $-0.0582$ & 0.00171 & $-34.11$ & $5.996\times10^{-255}$ \\
Difficulty        & 0.159  & 0.088   & 1.81   & $7.081\times10^{-2}$ \\
RAG Score         & $-0.875$ & 0.103 & $-8.47$ & $2.423\times10^{-17}$ \\
Embedding PC10    & $-0.043$ & 0.010 & $-4.47$ & $7.713\times10^{-6}$ \\
Embedding PC3     & $-0.030$ & 0.007 & $-4.11$ & $3.922\times10^{-5}$ \\
Embedding PC7     & $-0.036$ & 0.009 & $-4.08$ & $4.509\times10^{-5}$ \\
\bottomrule
\end{tabular}

\footnotesize

\emph{Notes.} $n = 11{,}000$ independent trials (one model query per unique SimpleQA question at each instructed threshold; independent observations, not technical replicates). Abstention was modelled by logistic regression. Coefficient significance was assessed using two-sided Wald $z$-tests. No correction was made for multiple comparisons. Full model includes threshold, confidence, difficulty, RAG score, and 10 embedding principal components (top 3 by $|z|$ shown; remaining 7 PCs: $|z|<3.3$). Standardized effect sizes: Confidence ($|\beta_{\mathrm{std}}|=1.09$) and Threshold ($|\beta_{\mathrm{std}}|=1.03$) are comparable and substantially larger than RAG ($|\beta_{\mathrm{std}}|=0.22$), Difficulty ($|\beta_{\mathrm{std}}|=0.05$), and the strongest embedding component ($|\beta_{\mathrm{std}}|=0.11$). Exact two-sided Wald-test $P$ values are reported in the table.
\end{table}

%--- Extended Data Table 4: GPT-4o Phase 4 model comparison ---updated
\begin{table}[H]
\centering
\caption{Phase 4: Model comparison for abstention predictions, GPT-4o.}
\label{tab:OCTNEW_GPT-phase3-model-compare}
\begin{tabular}{lccc}
\toprule
Model & AIC ($\downarrow$) & pseudo-$R^2$ ($\uparrow$) & Key likelihood-ratio tests \\
\midrule
Threshold only (T)            & 12874.3 & 0.105 & -- \\
T + Difficulty                & 12360.7 & 0.141 & -- \\
T + RAG                       & 12765.8 & 0.113 & -- \\
T + Embeddings (10 PCs)       & 12778.1 & 0.114 & -- \\
T + Confidence                & 10921.0 & 0.241 & vs.\ T: $\chi^2(1)=1955.3,\;P<2.2\times10^{-16}$ \\
                              &         &       & vs.\ T+RAG: $\chi^2(1)=1897.9,\;P<2.2\times10^{-16}$ \\
                              &         &       & vs.\ T+Embed: $\chi^2(1)=1907.6,\;P<2.2\times10^{-16}$ \\
T + Confidence + Difficulty   & 10917.9 & 0.242 & vs.\ T+Conf: $\chi^2(1)=5.0,\;P=2.483\times10^{-2}$ \\
Full (all predictors)         & 10800.4 & 0.251 & vs.\ T+Conf: $\chi^2(14)=142.6,\;P<2.2\times10^{-16}$ \\
\bottomrule
\end{tabular}

\footnotesize

\emph{Notes.} $n = 11{,}000$ independent trials (one model query per unique SimpleQA question at each instructed threshold; independent observations, not technical replicates). Models were compared using two-sided likelihood-ratio tests. Exact two-sided $P$ values are reported in the table. No correction was made for multiple comparisons. Full model includes threshold, confidence, difficulty, RAG score, and 10 embedding principal components. Confidence dominates all alternative predictors: adding confidence to models with T+RAG or T+Embeddings yields $\Delta$AIC $>1800$, while adding RAG or embeddings to T+Confidence yields $\Delta$AIC $<50$.

\end{table}
%======================================================================
% PHASE 4 ABSTENTION CONFIDENCE
%======================================================================
\subsection*{Phase 4: Logistic Regression Analysis of Abstention Confidence Data, GPT-4o}
Whilst our primary aim was to characterize the relationship between instructed thresholds, Phase 1 confidence and abstention behavior, for completeness we also analyzed the relationship between these predictors and the underlying driver of abstention behavior: a continuous variable, Phase 4 confidence in the abstention choice (see Figure 4).

To do this, we performed a linear regression with a nested model that was directly analogous to that used to analyze binary abstention data. The analysis of phase 4 abstention confidence similarly revealed systematic effects of both instructed threshold and phase 1 chosen confidence. Abstention confidence increased with higher instructed thresholds ($\beta_T=0.0033$, SE$=0.000050$, $z=65.6$, $p < 0.001$), and decreased sharply as phase 1 chosen confidence increased ($\beta_C=-0.0042$, SE$=0.00011$, $z=-39.5$, $p < 0.001$). In practical terms, this means that raising the instructed threshold by 20 points increased the model's abstention confidence by 0.066 (on the 0--1 scale). A 20-point increase in chosen confidence reduced abstention confidence by 0.084. Therefore, the more confident the model was in its initial choice -- and the lower the instructed threshold -- the less probability mass it assigned to the abstention option. Item difficulty had a statistically significant but small effect ($\beta_D=0.038$, SE$=0.0060$, $z=6.39$, $p < 0.001$), though its magnitude was minor compared to threshold and confidence: from the hardest to easiest items, abstention confidence increased by only 3.8 percentage points. The full model explained 36.1\% of variance ($R^2=0.361$). Confidence was the dominant predictor, improving model fit five times more than difficulty when added to the baseline threshold model (confidence: $\Delta R^2=0.109$, $\Delta$AIC$=-1731$; difficulty: $\Delta R^2=0.021$, $\Delta$AIC$=-312$).

These findings parallel the results of the binary abstention analyses: both demonstrate that abstention behavior is jointly shaped by instructed threshold and the model's own pre-decisional (i.e. phase 1) confidence. Together, our analyses of abstention confidence demonstrate that instructed threshold shifts the model toward abstention even before the decision boundary is crossed, while confidence pulls it back toward answering --- providing convergent evidence, via a richer signal, for the same underlying mechanism identified in the binary analyses.

%--- Extended Data Figure 4: GPT-4o Phase 4 heatmap ---
\begin{figure}[H]
    \centering
    \includegraphics[width=0.55\textwidth]{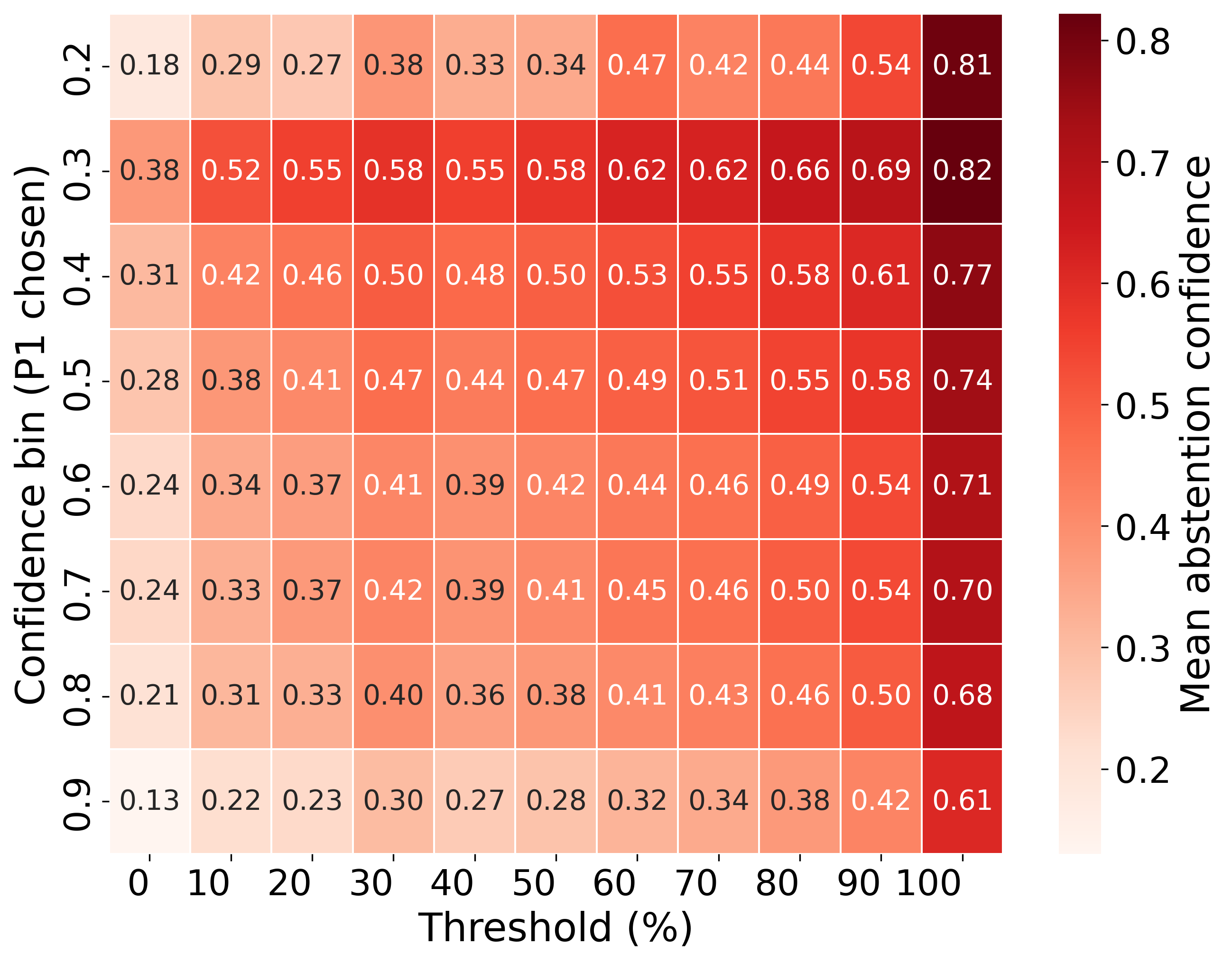}
    \caption{
    Heatmap of GPT-4o's abstention confidence in Phase 4, shown as a function of instructed threshold and Phase 1 chosen confidence. Warmer colors indicate stronger abstention confidence, also specified by numbers inside cells. The continuous abstention confidence reveals a linear additive structure: abstention confidence increases steadily with instructed threshold and decreases steadily with chosen confidence, with difficulty having only a minor influence. See Main text for details}
    \label{fig:OCTNEW_GPT_phase3_abstention_conf_heatmap}
\end{figure}

%======================================================================
% GEMMA 3 27B RESULTS
%======================================================================
\subsection*{Gemma 3 27B Results}
\subsubsection*{Phase 1 and 2 Results for Gemma 3 27B}
\begin{figure}[H]
    \centering
    \includegraphics[width=0.8\textwidth]{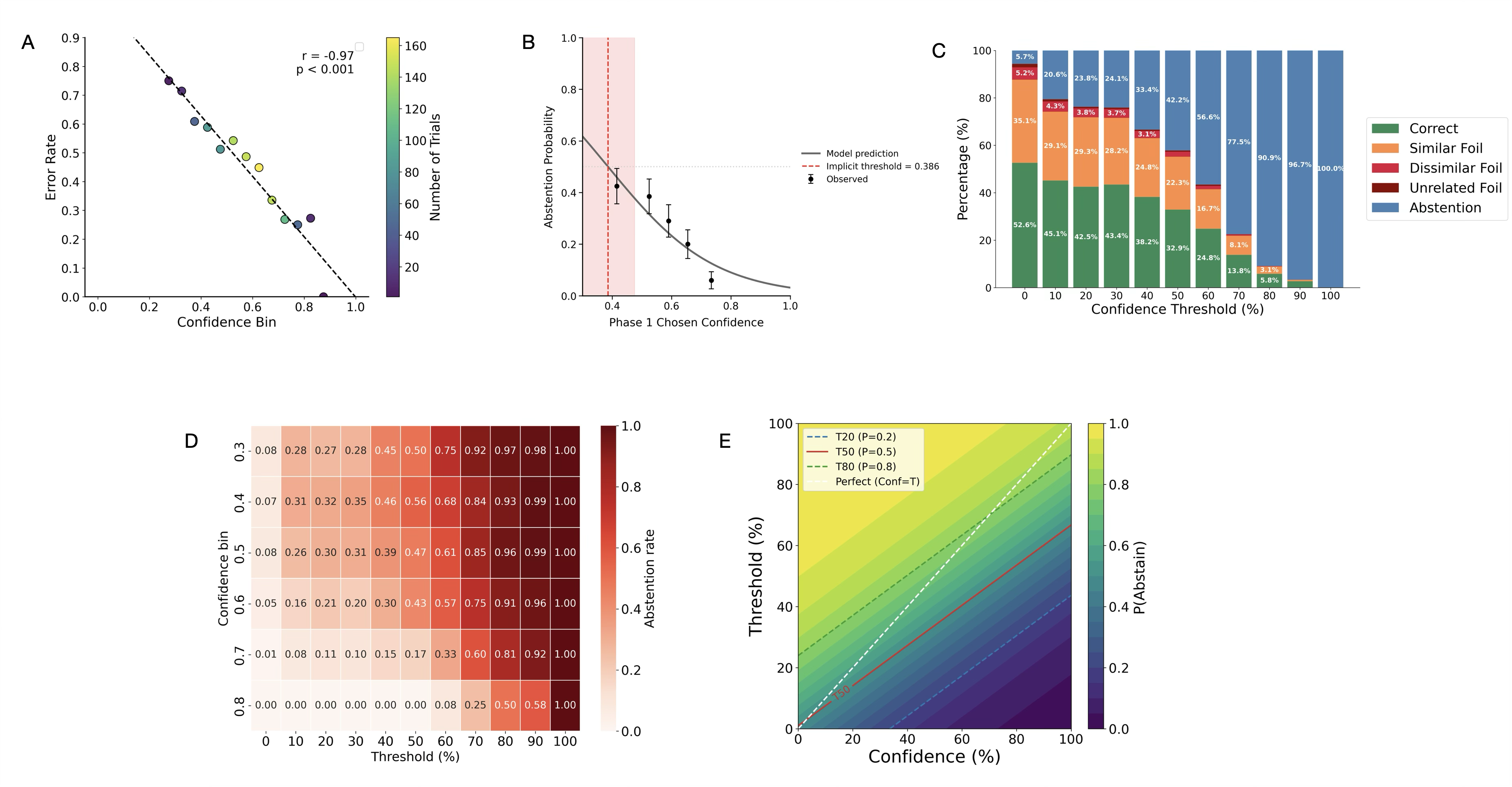}
    \caption{\textbf{Gemma 3 27B results across phases 1,2 and 4.}(A) (A) Gemma 3 27B confidence predicts error rate (Phase 1). Relationship between binned confidence and error rate in Phase 1. Each point represents one confidence bin (bin width = 0.05; $n=14$ confidence bins); colour indicates the number of independent trials contributing to each bin. A strong negative relationship was observed between confidence and error rate (two-sided Pearson correlation, $r(12) = -0.966$, $P = 9.02\times10^{-8}$). The model was calibrated on a separate dataset of 1,000 independent trials (see Methods). (B) Gemma 27B confidence predicts abstention behavior (phase 2). Logistic regression model of natural abstention probability as a function of confidence for $n=1000$ trials. The grey curve indicates the model prediction, fitted to individual binary abstention decisions. The red dashed line marks the implicit decision threshold at 38.6\%, where $P(\mathrm{abstain})=0.5$ (holding difficulty constant at sample mean). The red shaded region denotes the confidence interval spanning approximately $\pm9$ percentage points around the threshold, illustrating the sharpness/softness of the decision boundary (policy temperature = 0.18 units, or 18 percentage points). Black circles represent observed abstention rates (mean $\pm$ 95\% CI) across confidence quintiles. The low threshold (38.6\%) indicates that Gemma 27B continues answering even at relatively low confidence levels, only choosing to abstain when confidence drops below this point, revealing the model's intrinsic abstention boundary even without explicit threshold instructions. (C) Performance of Gemma 27B in Phase 4 as a function of instructed threshold. (D) Profile of abstention behavior shown by Gemma in phase 4, as a function of Phase 1 confidence and instructed threshold. Abstention rate (0-1) increases from left to right (instructed threshold), and from bottom to top (binned confidence). Confidence: calibrated chosen confidence from phase 1, binned into fixed bins of size 0.1 (in the range 0.3-1.0). (E) Decision rule guiding abstention behavior shown by fitted model. Behavior of perfectly calibrated model shown along the diagonal (i.e. Conf = Threshold). In comparison, the fitted model is moderately conservative due to under-weighting its own confidence relative to the instructed threshold (scale = 0.66), such that a 1\% increase in confidence offsets only approximately 0.66\% of threshold. The baseline bias (shift = +1.0\%) is negligible. Contours $T_{20}$, $T_{50}$, $T_{80}$ mark thresholds at which the model abstains with 20\%, 50\%, 80\% probability. Points to the right/below each line correspond to lower abstention rates, while points to the left/above correspond to higher abstention rates. The moderate gap between $T_{20}$ and $T_{80}$ reflects a moderately soft transition (policy temperature = 16.6) of the decision boundary.
}
    \label{fig:OCTNEW_gemmacomposite}
\end{figure}

Gemma 3 performed at 55.1\% correct during phase 1 where there was a 4-way choice between available answers, with no option to abstain.  We calibrated the model using the temperature scaling method of \citep{guo2017calibration}(see Methods), resulting in an optimal scaling temperature of 5.1 (ECE = 0.049; AUROC = 0.81). There was a robust relationship between calibrated confidence and error rate on the Phase 1 test dataset, with higher confidence strongly predicting correct responses (\ref{fig:OCTNEW_gemmacomposite}).

In phase 2, the model was exposed to the same multiple choice questions, but had the additional option to abstain. Here performance was 41.6\% correct, 31.2\% incorrect, and 27.2\% abstention. Accuracy on answered questions was higher in phase 2 compared to phase 1 (60.5\% vs 55.1\%), with coverage reduced from 100\% to 68.8\%.

A logistic regression analysis (see Methods) revealed that question difficulty alone (see difficulty score in Methods) provided a degree of predictive power (AIC = 1143.8, $\text{pseudo-}R^2 = 0.026$, $\beta_D = -1.22$, SE = 0.22, $z = -5.49$, $p < 0.001$). Because Gemma and GPT4o were evaluated on the same multiple-choice items, we used the multi-seed GPT4o difficulty score as an external measure of item hardness. This score reflects the expected probability of a correct answer across random answer-choice allocations, providing a model-agnostic estimate of how objectively hard each question is.

Importantly, adding the model's chosen confidence from Phase 1 improved model fit (AIC = 1077.9, change in AIC = $-65.9$, $\text{pseudo-}R^2 = 0.084$, LR $\chi^2(1) = 67.85$, $p < 0.001$). In the full model, difficulty remained significant ($\beta_D = -0.84$, SE = 0.24, $z = -3.55$, $p < 0.001$). The confidence effect was prominent ($\beta_C = -5.57$, SE = 0.70, $z = -7.94$, $p < 0.001$), indicating that higher confidence strongly reduces abstention probability (see \ref{tab:OCTNEW_gemma-phase2-regression}). In practical terms, a 0.1-unit increase in confidence (e.g., from 0.7 to 0.8) reduces the abstention odds by approximately 43\% (odds ratio = 0.57). 

Whilst an explicit threshold for abstention was not stated in the instructions to the model in Phase 2, the form of our logistic regression model allows us to infer a fitted threshold that underlies the model's behavior (see Methods). We found the confidence level at which the model abstains 50\% of the time, the indifference point $T_{50}$, to be approximately 38.4\% (holding difficulty constant at the sample mean of 0.66) -- with the policy temperature parameter of 0.18 units (equivalent to $1/|\beta_C|$)(see \ref{fig:OCTNEW_gemmacomposite}B).

%--- Extended Data Table 5: Gemma Phase 2 regression ---updated
\begin{table}[H]
\centering
\caption{Gemma 27B: Logistic regression predicting abstention in Phase 2}
\label{tab:OCTNEW_gemma-phase2-regression}
\begin{tabular}{lcccccc}
\toprule
\multirow{2}{*}{Predictor} &
\multicolumn{3}{c}{Model A: Difficulty Only} &
\multicolumn{3}{c}{Model B: Difficulty + Confidence} \\
\cmidrule(lr){2-4} \cmidrule(lr){5-7}
& Coef. & SE & $z$ & Coef. & SE & $z$ \\
\midrule
Intercept  & $-0.234$      & 0.149 & $-1.56$ & $2.692$      & 0.399 & $6.75$ \\
Difficulty & $-1.215$***   & 0.221 & $-5.49$ & $-0.837$***  & 0.236 & $-3.55$ \\
Confidence & --            & --    & --      & $-5.575$***  & 0.702 & $-7.94$ \\
\midrule
AIC & \multicolumn{3}{c}{1143.8} & \multicolumn{3}{c}{1077.9} \\
Pseudo-$R^2$ & \multicolumn{3}{c}{0.026} & \multicolumn{3}{c}{0.084} \\
\bottomrule
\end{tabular}

\footnotesize

\emph{Notes.} $n = 1{,}000$ independent trials (one model query per unique SimpleQA question; independent observations, not technical replicates). Abstention was modelled by logistic regression. Coefficient significance was assessed using two-sided Wald $z$-tests. No correction was made for multiple comparisons. Difficulty = mean accuracy across seeds (higher = easier). Confidence = Phase~1 chosen confidence. Exact two-sided $P$ values --- \emph{Model A}: Intercept $P = 1.19\times10^{-1}$; Difficulty $P = 4.02\times10^{-8}$. \emph{Model B}: Intercept $P = 1.48\times10^{-11}$; Difficulty $P = 3.85\times10^{-4}$; Confidence $P = 2.04\times10^{-15}$.

\end{table}
%--- Extended Data Figure 5: Gemma results across phases ---

\subsubsection*{Gemma 3 27B, Phase 4: Analysis of Performance}
The prompt used for GPT-4o proved ineffective for Gemma 3 27b, yielding abstention rates of below 5\% until the 80\% threshold and only averaging 34.4\% above this threshold. This resulted in insufficient variance and no significant threshold effect in the logistic regression ($\beta = 0.0012 \pm 0.001,\; z = 1.71,\; p = 0.086$).

\begin{figure}[H]
    \centering
    \includegraphics[width=0.8\textwidth]{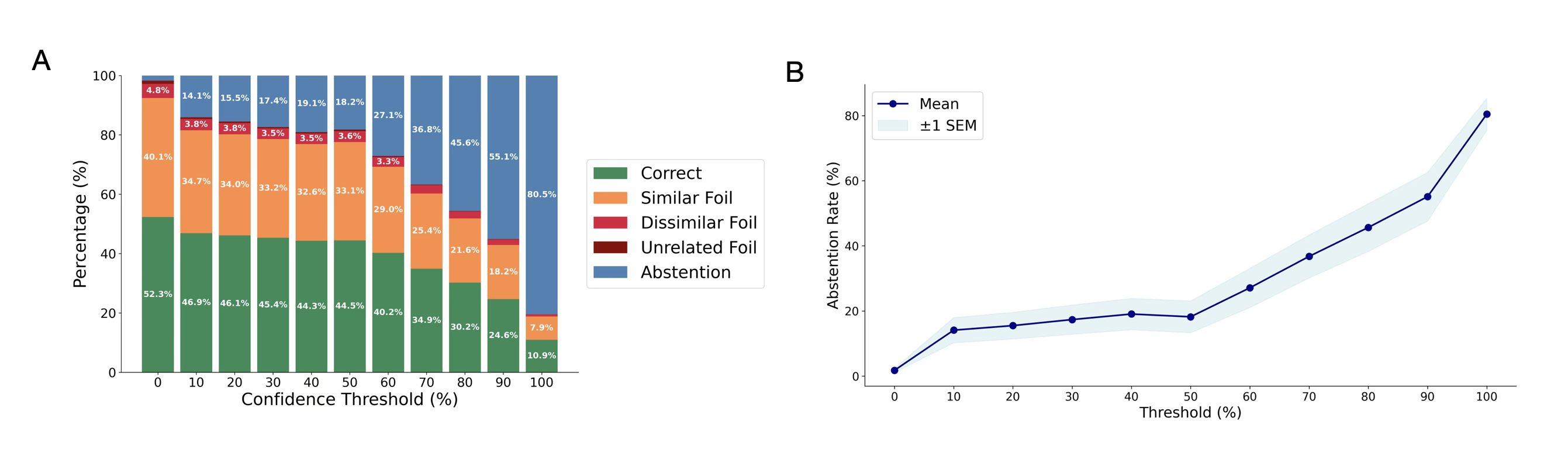}
    \caption{\textbf{Performance of Gemma 27B on Phase 4 averaged across 20 prompts.} (A) Performance of Gemma 3 27B averaged across 20 prompts (see Methods for details). (B) Abstention rate as a function of instructed threshold, averaged across 20 prompts.}
    \label{fig:OCTNEW_Gemma_multiprompt_phase4}
\end{figure}

To establish robust abstention behavior, we generated 20 paraphrases of the basic GPT4o prompt using Gemini 2.5 Pro while preserving meaning and answer format. As shown in \ref{fig:OCTNEW_Gemma_multiprompt_phase4}, this strategy was successful: 13 of the 20 prompts exceeded 80\% abstention at the 100\% threshold, confirming that appropriate prompt framing is critical for eliciting reliable abstention behavior in this model. We used the prompt with the highest (i.e. 100\%) abstention rate at the 100\% instructed threshold for the main experiment (see Methods for prompt used). 

We examined the overall pattern of correct, incorrect and abstention responses showed by the model. As expected, there was a clear effect of threshold on abstention rate -- with increasing abstention rates as the threshold was increased (see \ref{fig:OCTNEW_gemmacomposite}C). 

%--- Extended Data Figure 6: Gemma multi-prompt ---

We examined whether abstention behavior was systematically shaped by the instructed threshold, and whether this influence was modulated by the model's own confidence (\ref{fig:OCTNEW_gemmacomposite}D). As for GPT4o, for this analysis we used chosen confidence estimates from phase 1, since these were elicited in the absence of an abstention option. 

\subsubsection*{Gemma 27B: Computational Modeling of binary abstention data, Phase 4}
We next sought to fit a quantitative model to the profile of abstention behavior observed. Specifically, we asked whether chosen confidence from phase 1 was a robust predictor of abstention behavior, above and beyond any effects of instructed threshold and difficulty. To do this, we set up a nested logistic regression model whose dependent variable was binary abstention decisions, and included predictors for chosen confidence, instructed threshold, and our measure of question difficulty. We found that a model that included chosen confidence, threshold and difficulty predictors (AIC = 9544.4; $\text{pseudo-}R^2 = 0.374$) outperformed baseline models that included either threshold only (AIC = 10021.8; $\text{pseudo-}R^2 = 0.342$) or threshold and difficulty (AIC = 9824.5; $\text{pseudo-}R^2 = 0.355$) or threshold and confidence (AIC = 9632.0; $\text{pseudo-}R^2 = 0.368$) (see \ref{tab:OCTNEW_p12_Gemma-phase3-model-compare}). As such, a likelihood ratio test (LRT) showed a highly significant benefit for the addition of confidence to a threshold and difficulty model ($\chi^2(1) = 282.15,\; p < 0.001$). These results, therefore, provide robust evidence that both instructed threshold and phase 1 chosen confidence markedly influence abstention behavior.

We next examined the fitted coefficients. This indicated that abstention rate increased with threshold ($\beta_T=0.060$, SE$=0.0011$, $z=55.4$, $p < 0.001$) and decreased with confidence ($\beta_C=-0.040$, SE$=0.0024$, $z=-16.4$, $p < 0.001$). Difficulty was also significant ($\beta_D=-0.78$, SE$=0.083$, $z=-9.40$, $p < 0.001$), consistent with a greater tendency to abstain on harder items (see \ref{tab:OCTNEW_p12_Gemma-phase3-coefs}). We derived three interpretable quantities: scale (sensitivity to confidence relative to threshold; 0.66), shift (baseline abstention bias; $+1.0$), and policy temperature (softness of the boundary; 16.6). We also computed the indifference threshold, $T_{50}$, at which the model is equally likely to answer or abstain: for example, at a confidence level of 80\% the model's indifference point is a threshold of approximately 53.5\% (see \ref{fig:OCTNEW_gemmacomposite}E). Given that a perfectly calibrated model abstains when its own confidence is equal to the instructed threshold (assuming symmetric costs), this result indicates that the model is moderately conservative in its tendency to abstain. The scale parameter of 0.66 indicates that the model under-weights its own confidence relative to the instructed threshold: a 1\% increase in confidence offsets only approximately 0.66\% of threshold.
%--- Extended Data Table 6: Gemma Phase 4 regression ---updated
\begin{table}[H]
\centering
\caption{Gemma 3 27B: Logistic regression predicting abstention (Phase 4), full model}
\label{tab:OCTNEW_p12_Gemma-phase3-coefs}
\begin{tabular}{lrrrr}
\toprule
Predictor & Coefficient ($\beta$) & Std.\ Error & $z$ & $P$ \\
\midrule
Intercept        & $-0.058$ & 0.138  & $-0.42$  & $4.991\times10^{-1}$ \\
Threshold (T)    & 0.060    & 0.0011 & 55.35    & $<2.2\times10^{-16}$ \\
Confidence (\%)  & $-0.040$ & 0.0024 & $-16.42$ & $7.508\times10^{-62}$ \\
Difficulty       & $-0.778$ & 0.083  & $-9.40$  & $1.019\times10^{-10}$ \\
\bottomrule
\end{tabular}

\footnotesize

\emph{Notes.} $n = 11{,}000$ independent trials (one model query per unique SimpleQA question at each instructed threshold; independent observations, not technical replicates). Abstention was modelled by logistic regression. Coefficient significance was assessed using two-sided Wald $z$-tests. No correction was made for multiple comparisons. Exact two-sided Wald-test $P$ values are reported in the table.

\end{table}
%--- Extended Data Table 7: Gemma Phase 4 model comparison ---updated
\begin{table}[H]
\centering
\caption{Gemma 3 27B: Model comparison for abstention predictions (Phase 4)}
\label{tab:OCTNEW_p12_Gemma-phase3-model-compare}
\begin{tabular}{lccc}
\toprule
Model & AIC ($\downarrow$) & pseudo-$R^2$ ($\uparrow$) & Key likelihood-ratio tests \\
\midrule
Threshold only (T)                  & 10021.8 & 0.342 & -- \\
Threshold + Confidence              & 9632.0  & 0.368 &
vs.\ T: $\chi^2(1)=391.82,\;P=3.322\times10^{-87}$ \\
Threshold + Difficulty              & 9824.5  & 0.355 & -- \\
Full (T + Confidence + Difficulty)  & 9544.4  & 0.374 &
vs.\ T+Conf: $\chi^2(1)=89.62,\;P=2.890\times10^{-21}$ \\
\bottomrule
\end{tabular}

\footnotesize

\emph{Notes.} $n = 11{,}000$ independent trials (one model query per unique SimpleQA question at each instructed threshold; independent observations, not technical replicates). Models were compared using two-sided likelihood-ratio tests. Exact two-sided $P$ values are reported in the table. No correction was made for multiple comparisons.

\end{table}
%======================================================================
% DEEPSEEK 671B RESULTS
%======================================================================
\subsection*{DeepSeek 671B Results}
\subsubsection*{Phase 1 and 2 Results for DeepSeek 671B}
\begin{figure}[H]
    \centering
    \includegraphics[width=0.8\textwidth]{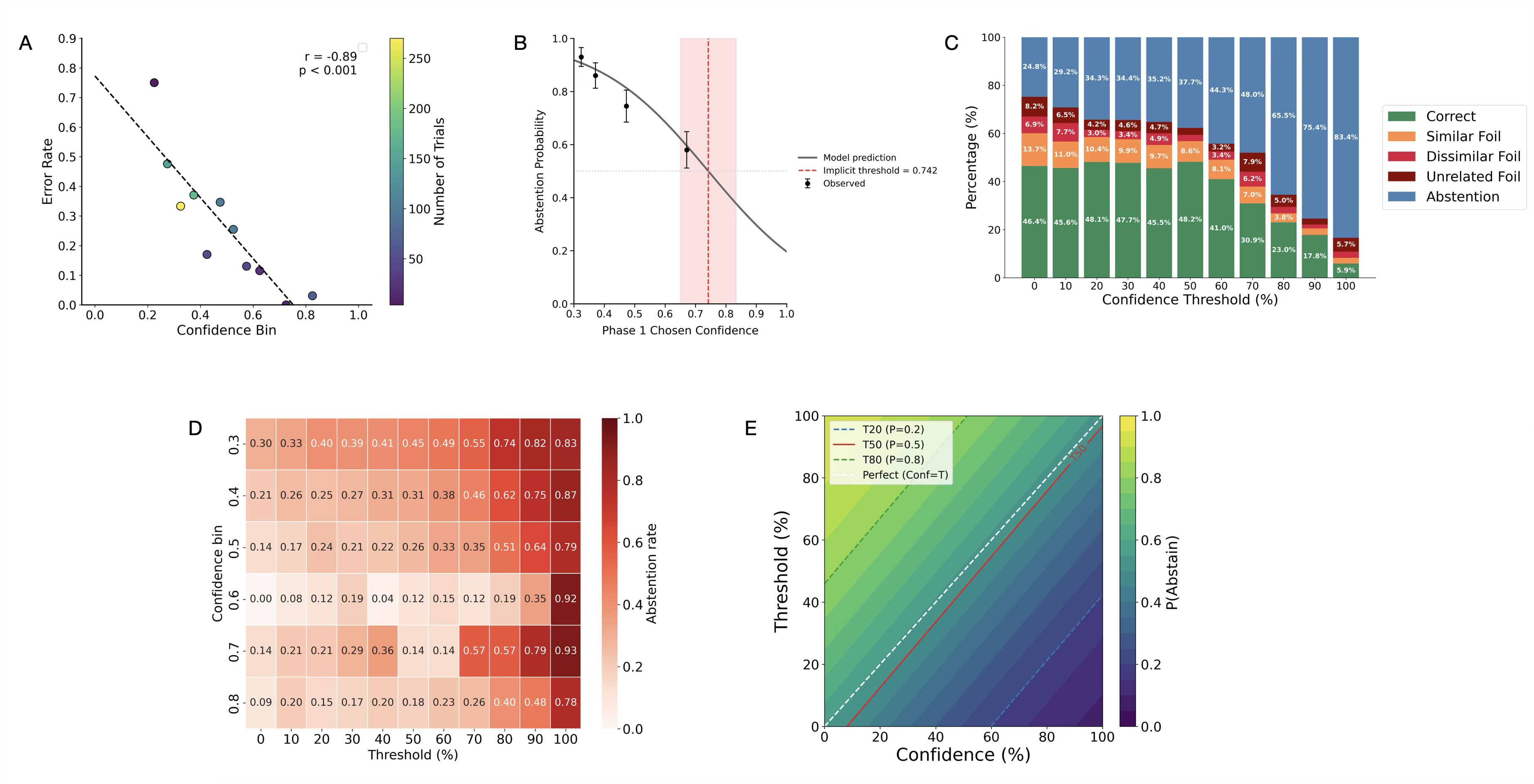}
    \caption{\textbf{DeepSeek 671B results across phases 1,2 and 4.}(A) DeepSeek 671B confidence predicts error rate (Phase 1). Relationship between binned confidence and error rate in Phase 1. Each point represents one confidence bin (bin width = 0.05; $n=14$ confidence bins); colour indicates the number of independent trials contributing to each bin. A strong negative relationship was observed between confidence and error rate (two-sided Pearson correlation, $r(12) = -0.930$, $P = 1.49\times10^{-6}$). The model was calibrated on a separate dataset of 1,000 independent trials (see Methods). (B) DeepSeek 671B confidence predicts abstention behavior (phase 2). Logistic regression model of natural abstention probability as a function of confidence for $n=1000$ trials. The grey curve indicates the model prediction, fitted to individual binary abstention decisions. The red dashed line marks the implicit decision threshold at 74.2\%, where $P(\mathrm{abstain})=0.5$ (holding difficulty constant at sample mean). The red shaded region denotes the confidence interval spanning approximately $\pm9$ percentage points around the threshold, illustrating the sharpness/softness of the decision boundary (policy temperature = 0.18 units, or 18 percentage points). Black circles represent observed abstention rates (mean $\pm$ 95\% CI) across confidence quintiles. The high threshold (74.2\%) indicates that DeepSeek requires relatively high confidence before choosing to answer, readily abstaining when confidence drops below this point, revealing the model's intrinsic abstention boundary even without explicit threshold instructions. (C) Performance of DeepSeek in Phase 4 as a function of instructed threshold. (D) Profile of abstention behavior shown by DeepSeek in phase 4, as a function of confidence and instructed threshold. Confidence: calibrated chosen confidence from phase 1, binned into fixed bins of size 0.1 (in the range 0.3- 0.9). (E) Decision rule guiding abstention behavior shown by fitted model. Behavior of perfectly calibrated model shown along the diagonal (i.e. Conf = Threshold). The fitted model exhibits near-perfect calibration with essentially equal weighting of confidence and threshold (scale = 1.05), such that a 1\% increase in confidence offsets approximately 1.05\% of threshold. The baseline bias is minimal (shift = $-8.5$\%). Contours $T_{20}$, $T_{50}$, $T_{80}$ mark thresholds at which the model abstains with 20\%, 50\%, 80\% probability. Points to the right/below each line correspond to lower abstention rates, while points to the left/above correspond to higher abstention rates. The wide gap between $T_{20}$ and $T_{80}$ reflects a relatively soft transition (policy temperature = 39.3) of the decision boundary, indicating gradual rather than sharp changes in abstention probability across the confidence-threshold space.
}
    \label{fig:OCTNEW_deepseek_composite}
\end{figure}

DeepSeek 671B performed at 67.7\% correct during phase 1 where there was a 4-way choice between available answers, with no option to abstain.  We calibrated the model using the temperature scaling method of \citep{guo2017calibration}(see Methods), resulting in an optimal scaling temperature of 29.3 (ECE = 0.0065; AUROC = 0.83).  A high scaling temperature was required since the model was very overconfident pre-calibration. There was a robust relationship between calibrated confidence and error rate on the Phase 1 test dataset, with higher confidence strongly predicting correct responses (\ref{fig:OCTNEW_deepseek_composite}).

In phase 2, the model was exposed to the same multiple choice questions, but had the additional option to abstain. Here performance was 13.8\% correct, 4.2\% incorrect, and 82.0\% abstention. Accuracy on answered questions was higher in phase 2 compared to phase 1 (76.7\% vs 67.7\%), with coverage reduced from 100\% to 83.0\%.

A logistic regression analysis (see Methods) revealed that question difficulty alone (see difficulty score in Methods) provided a degree of predictive power (AIC = 930.9, $\text{pseudo-}R^2 = 0.017$, $\beta_D = -1.06$, SE = 0.27, $z = -3.88$, $p < 0.001$). Because DeepSeek and GPT4o were evaluated on the same multiple-choice items, we used the multi-seed GPT4o difficulty score as an external measure of item hardness. This score reflects the expected probability of a correct answer across random answer-choice allocations, providing a model-agnostic estimate of how objectively hard each question is.

Adding the model's chosen confidence from Phase 1 substantially improved model fit (AIC = 817.7, change in AIC = $-113.2$, $\text{pseudo-}R^2 = 0.139$, LR $\chi^2(1) = 115.21$, $p < 0.001$). In the full model, difficulty became non-significant ($\beta_D = -0.48$, SE = 0.29, $z = -1.66$, $p = 0.097$). The confidence effect was highly prominent ($\beta_C = -5.46$, SE = 0.53, $z = -10.28$, $p < 0.001$), indicating that higher confidence strongly reduces abstention probability (see \ref{tab:OCTNEW_deepseek-phase2-regression}). In practical terms, a 0.1-unit increase in confidence (e.g., from 0.7 to 0.8) reduces the abstention odds by approximately 42\% (odds ratio = 0.58). 

Whilst an explicit threshold for abstention was not stated in the instructions to the model in Phase 2, the form of our logistic regression model allows us to infer a fitted threshold that underlies the model's behavior (see Methods). We found the confidence level at which the model abstains 50\% of the time, the indifference point $T_{50}$, to be approximately 74.2\% (holding difficulty constant at the sample mean of 0.65) -- with the policy temperature parameter of 0.18 units (equivalent to $1/|\beta_C|$)(see \ref{fig:OCTNEW_deepseek_composite}B).

%--- Extended Data Table 8: DeepSeek Phase 2 regression ---updated
\begin{table}[H]
\centering
\caption{DeepSeek 671B: Logistic regression predicting abstention in Phase 2}
\label{tab:OCTNEW_deepseek-phase2-regression}
\begin{tabular}{lcccccc}
\toprule
\multirow{2}{*}{Predictor} &
\multicolumn{3}{c}{Model A: Difficulty Only} &
\multicolumn{3}{c}{Model B: Difficulty + Confidence} \\
\cmidrule(lr){2-4} \cmidrule(lr){5-7}
& Coef. & SE & $z$ & Coef. & SE & $z$ \\
\midrule
Intercept  & $2.238$***    & 0.212 & $10.55$ & $4.364$***   & 0.313 & $13.96$ \\
Difficulty & $-1.060$***   & 0.273 & $-3.88$ & $-0.481$     & 0.290 & $-1.66$ \\
Confidence & --            & --    & --      & $-5.461$***  & 0.531 & $-10.28$ \\
\midrule
AIC & \multicolumn{3}{c}{930.9} & \multicolumn{3}{c}{817.7} \\
Pseudo-$R^2$ & \multicolumn{3}{c}{0.017} & \multicolumn{3}{c}{0.139} \\
\bottomrule
\end{tabular}

\footnotesize

\emph{Notes.} $n = 1{,}000$ independent trials (one model query per unique SimpleQA question; independent observations, not technical replicates). Abstention was modelled by logistic regression. Coefficient significance was assessed using two-sided Wald $z$-tests. No correction was made for multiple comparisons. Difficulty = mean accuracy across seeds (higher = easier). Confidence = Phase~1 chosen confidence. Exact two-sided $P$ values --- \emph{Model A}: Intercept $P = 4.89\times10^{-26}$; Difficulty $P = 1.05\times10^{-4}$. \emph{Model B}: Intercept $P = 2.23\times10^{-44}$; Difficulty $P = 9.72\times10^{-2}$; Confidence $P = 8.62\times10^{-25}$.

\end{table}
%--- Extended Data Figure 7: DeepSeek results across phases ---

\subsubsection*{DeepSeek 671B, Phase 4: Analysis of Performance}
We examined the overall pattern of correct, incorrect and abstention responses showed by the model. As expected, there was a clear effect of threshold on abstention rate -- with increasing abstention rates as the threshold was increased (\ref{fig:OCTNEW_deepseek_composite}C). 

We examined whether abstention behavior was systematically shaped by the instructed threshold, and whether this influence was modulated by the model's own confidence (\ref{fig:OCTNEW_deepseek_composite}D). As before, we used chosen confidence estimates from phase 1, since these were elicited in the absence of an abstention option. 

\subsubsection*{DeepSeek 671B: Computational Modeling of binary abstention data, Phase 4}
We next sought to fit a quantitative model to the profile of abstention behavior observed. Specifically, we asked whether chosen confidence from phase 1 was a robust predictor of abstention behavior, above and beyond any effects of instructed threshold and difficulty. To do this, we set up a nested logistic regression model whose dependent variable was binary abstention decisions, and included predictors for chosen confidence, instructed threshold, and our measure of question difficulty. 
We found that a model that included chosen confidence, threshold and difficulty predictors (AIC = 13136.4; pseudo-$R^2$ = 0.136) outperformed baseline models that included either threshold only (AIC = 13798.3; pseudo-$R^2$ = 0.092), threshold and difficulty (AIC = 13540.5; pseudo-$R^2$ = 0.109), or threshold and confidence (AIC = 13254.0; pseudo-$R^2$ = 0.128) (see Supplementary Table~10). As such, a likelihood-ratio test showed a highly significant benefit for the addition of confidence to a threshold-plus-difficulty model ($\chi^2(1)=406.12$, $P=2.43\times10^{-90}$). These results, therefore, provide robust evidence that both instructed threshold and phase 1 chosen confidence markedly influence abstention behavior.

We next examined the fitted coefficients. This indicated that abstention rate increased with threshold ($\beta_T=0.025$, SE$=0.00071$, $z=35.9$, $p < 0.001$) and decreased with confidence ($\beta_C=-0.027$, SE$=0.0015$, $z=-17.6$, $p < 0.001$). Difficulty was also significant ($\beta_D=-0.81$, SE$=0.066$, $z=-12.2$, $p < 0.001$), consistent with a greater tendency to abstain on harder items (see \ref{tab:OCTNEW_deepseek-phase3-coefs}). We derived three interpretable quantities: scale (sensitivity to confidence relative to threshold; 1.05), shift (baseline abstention bias; $-8.5$), and policy temperature (softness of the boundary; 39.3). We also computed the indifference threshold, $T_{50}$, at which the model is equally likely to answer or abstain: for example, at a confidence level of 80\% the model's indifference point is a threshold of approximately 75.7\% (see \ref{fig:OCTNEW_deepseek_composite}E). Given that a perfectly calibrated model abstains when its own confidence is equal to the instructed threshold, this result indicates that the model is essentially well-calibrated with near-equal weighting of confidence and threshold. The scale parameter of 1.05 indicates that the model appropriately weights its own confidence relative to the instructed threshold: a 1\% increase in confidence offsets approximately 1.05\% of threshold.
%--- Extended Data Table 9: DeepSeek Phase 4 regression ---updated
\begin{table}[H]
\centering
\caption{DeepSeek 671B: Logistic regression predicting abstention (Phase 4), full model}
\label{tab:OCTNEW_deepseek-phase3-coefs}
\begin{tabular}{lrrrr}
\toprule
Predictor & Coefficient ($\beta$) & Std.\ Error & $z$ & $P$ \\
\midrule
Intercept        & $0.216$  & 0.077   & $2.81$   & $4.95\times10^{-3}$ \\
Threshold (T)    & 0.025    & 0.00071 & 35.93    & $<2.2\times10^{-16}$ \\
Confidence (\%)  & $-0.027$ & 0.0015  & $-17.64$ & $2.71\times10^{-70}$ \\
Difficulty       & $-0.810$ & 0.066   & $-12.19$ & $2.95\times10^{-34}$ \\
\bottomrule
\end{tabular}

\footnotesize

\emph{Notes.} $n = 11{,}000$ independent trials (one model query per unique SimpleQA question at each instructed threshold; independent observations, not technical replicates). Abstention was modelled by logistic regression. Coefficient significance was assessed using two-sided Wald $z$-tests. No correction was made for multiple comparisons. Exact two-sided Wald-test $P$ values are reported in the table.

\end{table}
%--- Extended Data Table 10: DeepSeek Phase 4 model comparison ---updated
\begin{table}[H]
\centering
\caption{DeepSeek 671B: Model comparison for abstention predictions (Phase 4)}
\label{tab:OCTNEW_deepseek-phase3-model-compare}
\begin{tabular}{lccc}
\toprule
Model & AIC ($\downarrow$) & pseudo-$R^2$ ($\uparrow$) & Key likelihood-ratio tests \\
\midrule
Threshold only (T)                  & 13798.3 & 0.092 & -- \\
Threshold + Confidence              & 13254.0 & 0.128 &
vs.\ T: $\chi^2(1)=546.26,\;P=8.206\times10^{-121}$ \\
Threshold + Difficulty              & 13540.5 & 0.109 & -- \\
Full (T + Confidence + Difficulty)  & 13136.4 & 0.136 &
vs.\ T+Diff: $\chi^2(1)=406.12,\;P=2.43\times10^{-90}$ \\
\bottomrule
\end{tabular}

\footnotesize

\emph{Notes.} $n = 11{,}000$ independent trials (one model query per unique SimpleQA question at each instructed threshold; independent observations, not technical replicates). Models were compared using two-sided likelihood-ratio tests. Exact two-sided $P$ values are reported in the table. No correction was made for multiple comparisons.

\end{table}

%======================================================================
% QWEN 80B RESULTS
%======================================================================
\subsection*{Qwen 80B Results}
\subsubsection*{Phase 1 and 2 Results for Qwen 80B}
\begin{figure}[H]
    \centering
    \includegraphics[width=0.8\textwidth]{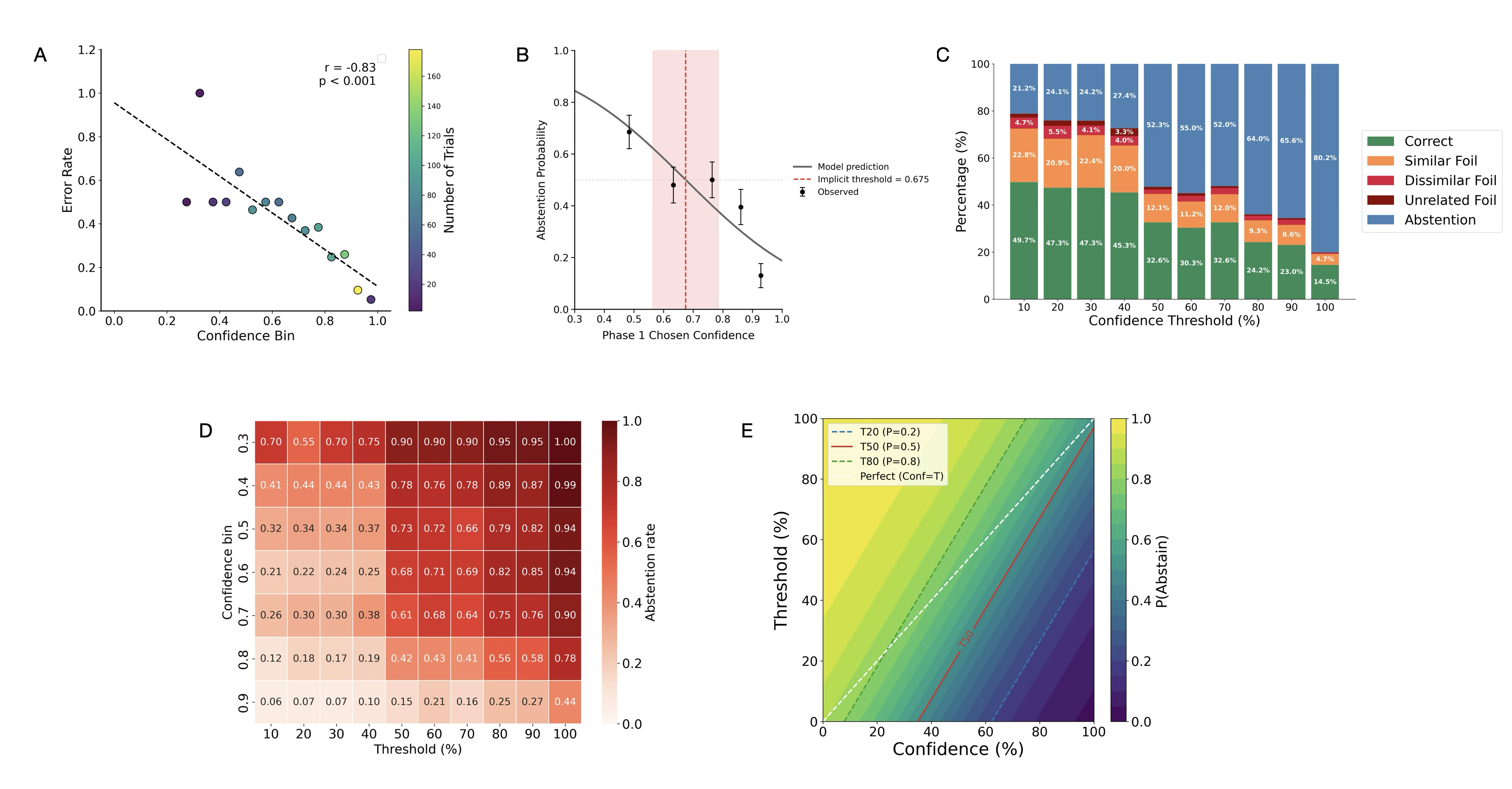}
    \caption{\textbf{Qwen 80B results across phases 1,2 and 4.}(A) Qwen 80B confidence predicts error rate (Phase 1). Relationship between binned confidence and error rate in Phase 1. Each point represents one confidence bin (bin width = 0.05; $n=14$ confidence bins); colour indicates the number of independent trials contributing to each bin. A strong negative relationship was observed between confidence and error rate (two-sided Pearson correlation, $r(12) = -0.833$, $P = 1.15\times10^{-4}$). The model was calibrated on a separate dataset of 1,000 independent trials (see Methods). (B) Qwen 80B confidence predicts abstention behavior (phase 2). Logistic regression model of natural abstention probability as a function of confidence for $n=1000$ trials. The grey curve indicates the model prediction, fitted to individual binary abstention decisions. The red dashed line marks the implicit decision threshold at 67.5\%, where $P(\mathrm{abstain})=0.5$ (holding difficulty constant at sample mean of 0.646). The red shaded region denotes the confidence interval spanning approximately $\pm11$ percentage points around the threshold, illustrating the sharpness/softness of the decision boundary (policy temperature = 0.22 units, or 22 percentage points). Black circles represent observed abstention rates (mean $\pm$ 95\% CI) across confidence quintiles. The moderately high threshold (67.5\%) indicates that Qwen 80B requires relatively high confidence before choosing to answer, readily abstaining when confidence drops below this point, revealing the model's intrinsic abstention boundary even without explicit threshold instructions.(C) Performance of Qwen 80B in Phase 4 as a function of instructed threshold. (D) Profile of abstention behavior shown by Qwen 80B in phase 4, as a function of confidence and instructed threshold. Confidence: calibrated chosen confidence from phase 1, binned into fixed bins of size 0.1 (in the range 0.3- 1.0). (E) Decision rule guiding abstention behavior shown by fitted model. Behavior of perfectly calibrated model shown along the diagonal (i.e. Conf = Threshold). The fitted model substantially over-weights internal confidence relative to instructed threshold (scale = 1.49), such that a 1\% increase in confidence offsets approximately 1.49\% of threshold. The large negative shift ($-52.3$\%) means that at very low confidence levels the model is overly cautious, but the high scale parameter causes the decision boundary to tilt steeply, such that at higher confidence levels (above ~50\%) the model becomes increasingly over-confident and answers more readily than a perfectly calibrated model would. Contours $T_{20}$, $T_{50}$, $T_{80}$ mark thresholds at which the model abstains with 20\%, 50\%, 80\% probability. Points to the right/below each line correspond to lower abstention rates, while points to the left/above correspond to higher abstention rates. The relatively wide gap between $T_{20}$ and $T_{80}$ reflects a moderately soft transition (policy temperature = 29.3 in percent units) of the decision boundary.
}
    \label{fig:OCTNEW_Qwen_composite}
\end{figure}

Qwen 80B performed at 65.8\% correct during phase 1 where there was a 4-way choice between available answers, with no option to abstain.  We calibrated the model using the temperature scaling method of \citep{guo2017calibration}(see Methods), resulting in an optimal scaling temperature of 4.9 (ECE = 0.078; AUROC = 0.89).  There was a robust relationship between calibrated confidence and error rate on the Phase 1 test dataset, with higher confidence strongly predicting correct responses (\ref{fig:OCTNEW_Qwen_composite}).

In phase 2, the model was exposed to the same multiple choice questions, but had the additional option to abstain. Here performance was 40.1\% correct, 16.1\% incorrect, and 43.8\% abstention. Accuracy on answered questions was higher in phase 2 compared to phase 1 (73.1\% vs 65.8\%), with coverage reduced from 100\% to 56.2\%.

A logistic regression analysis (see Methods) revealed that question difficulty alone (see difficulty score in Methods) provided a degree of predictive power (AIC = 1360.0, $\text{pseudo-}R^2 = 0.011$, $\beta_D = -0.76$, SE = 0.20, $z = -3.84$, $p < 0.001$). Because Qwen 80B and GPT4o were evaluated on the same multiple-choice items, we used the multi-seed GPT4o difficulty score as an external measure of item hardness. This score reflects the expected probability of a correct answer across random answer-choice allocations, providing a model-agnostic estimate of how objectively hard each question is.

Adding the model's chosen confidence from Phase 1 substantially improved model fit (AIC = 1256.6, change in AIC = $-103.3$, $\text{pseudo-}R^2 = 0.088$, LR $\chi^2(1) = 105.34$, $p < 0.001$). In the full model, difficulty became non-significant ($\beta_D = 0.04$, SE = 0.23, $z = 0.18$, $p = 0.857$). The confidence effect was highly prominent ($\beta_C = -4.51$, SE = 0.46, $z = -9.73$, $p < 0.001$), indicating that higher confidence strongly reduces abstention probability (see \ref{tab:OCTNEW_Qwen-phase2-regression}). In practical terms, a 0.1-unit increase in confidence (e.g., from 0.7 to 0.8) reduces the abstention odds by approximately 36\% (odds ratio = 0.64). 

Whilst an explicit threshold for abstention was not stated in the instructions to the model in Phase 2, the form of our logistic regression model allows us to infer a fitted threshold that underlies the model's behavior (see Methods). We found the confidence level at which the model abstains 50\% of the time, the indifference point $T_{50}$, to be approximately 67.5\% (holding difficulty constant at the sample mean) -- with the policy temperature parameter of 0.22 units (equivalent to $1/|\beta_C|$)(see \ref{fig:OCTNEW_Qwen_composite}B).

%--- Extended Data Table 11: Qwen Phase 2 regression ---updated
\begin{table}[H]
\centering
\caption{Qwen 80B: Logistic regression predicting abstention in Phase 2}
\label{tab:OCTNEW_Qwen-phase2-regression}
\begin{tabular}{lcccccc}
\toprule
\multirow{2}{*}{Predictor} &
\multicolumn{3}{c}{Model A: Difficulty Only} &
\multicolumn{3}{c}{Model B: Difficulty + Confidence} \\
\cmidrule(lr){2-4} \cmidrule(lr){5-7}
& Coef. & SE & $z$ & Coef. & SE & $z$ \\
\midrule
Intercept  & $0.241$        & 0.142 & $1.70$ & $3.017$***   & 0.323 & $9.34$ \\
Difficulty & $-0.765$***    & 0.199 & $-3.84$ & $0.041$      & 0.227 & $0.18$ \\
Confidence & --             & --    & --      & $-4.510$***  & 0.463 & $-9.73$ \\
\midrule
AIC & \multicolumn{3}{c}{1360.0} & \multicolumn{3}{c}{1256.6} \\
Pseudo-$R^2$ & \multicolumn{3}{c}{0.011} & \multicolumn{3}{c}{0.088} \\
\bottomrule
\end{tabular}

\footnotesize

\emph{Notes.} $n = 1{,}000$ independent trials (one model query per unique SimpleQA question; independent observations, not technical replicates). Abstention was modelled by logistic regression. Coefficient significance was assessed using two-sided Wald $z$-tests. No correction was made for multiple comparisons. Difficulty = mean accuracy across seeds (higher = easier). Confidence = Phase~1 chosen confidence. Exact two-sided $P$ values --- \emph{Model A}: Intercept $P = 8.91\times10^{-2}$; Difficulty $P = 1.24\times10^{-4}$. \emph{Model B}: Intercept $P = 9.03\times10^{-21}$; Difficulty $P = 8.64\times10^{-1}$; Confidence $P = 6.15\times10^{-22}$.

\end{table}
%--- Extended Data Figure 8: Qwen results across phases ---

\subsubsection*{Qwen 80B, Phase 4: Analysis of Performance}
We examined the overall pattern of correct, incorrect and abstention responses showed by the model. As expected, there was a clear effect of threshold on abstention rate -- with increasing abstention rates as the threshold was increased (see \ref{fig:OCTNEW_Qwen_composite}C). 

We examined whether abstention behavior was systematically shaped by the instructed threshold, and whether this influence was modulated by the model's own confidence (\ref{fig:OCTNEW_Qwen_composite}D). As before, we used chosen confidence estimates from phase 1, since these were elicited in the absence of an abstention option.

\subsubsection*{Qwen 80B: Computational Modeling of binary abstention data, Phase 4}
We next sought to fit a quantitative model to the profile of abstention behavior observed. Specifically, we asked whether chosen confidence from phase 1 was a robust predictor of abstention behavior, above and beyond any effects of instructed threshold and difficulty. To do this, we set up a nested logistic regression model whose dependent variable was binary abstention decisions, and included predictors for chosen confidence, instructed threshold, and our measure of question difficulty. We found that a model that included chosen confidence, threshold and difficulty predictors (AIC = 10961.1; $\text{pseudo-}R^2 = 0.207$) outperformed baseline models that included either threshold only (AIC = 12284.5; $\text{pseudo-}R^2 = 0.111$) or threshold and difficulty (AIC = 12062.4; $\text{pseudo-}R^2 = 0.127$). The contribution of difficulty in the full model was near zero: a model including only threshold and confidence had a comparable fit as the full model (AIC = 10962.7; $\text{pseudo-}R^2 = 0.207$) (see \ref{tab:OCTNEW_Qwen-phase3-model-compare}). As such, a likelihood ratio test (LRT) showed a highly significant benefit for the addition of confidence to a threshold and difficulty model ($\chi^2(1) = 1103.35,\; p < 0.001$). These results, therefore, provide robust evidence that both instructed threshold and phase 1 chosen confidence markedly influence abstention behavior.

We next examined the fitted coefficients. This indicated that abstention rate increased with threshold ($\beta_T=0.034$, SE$=0.0009$, $z=37.9$, $p < 0.001$) and decreased with confidence ($\beta_C=-0.051$, SE$=0.0016$, $z=-31.0$, $p < 0.001$). Difficulty was marginally non-significant ($\beta_D=-0.148$, SE$=0.077$, $z=-1.91$, $p = 0.056$) (see \ref{tab:OCTNEW_Qwen-phase3-coefs}). We derived three interpretable quantities: scale (sensitivity to confidence relative to threshold; 1.49), shift (baseline abstention bias; $-52.3$), and policy temperature (softness of the boundary; 29.3 in percent units, or 0.29 in 0--1 units). We also computed the indifference threshold, $T_{50}$, at which the model is equally likely to answer or abstain: for example, at a confidence level of 80\% the model's indifference point is a threshold of approximately 67.1\% (see \ref{fig:OCTNEW_Qwen_composite}E). The scale parameter of 1.49 indicates that the model weights its own confidence considerably more heavily than the instructed threshold: a 1\% increase in confidence offsets approximately 1.49\% of threshold. Given that a perfectly calibrated model abstains when its own confidence is equal to the instructed threshold, this result indicates that the model substantially over-weights its internal confidence relative to the instructed threshold.
%--- Extended Data Table 12: Qwen Phase 4 regression ---updated
\begin{table}[H]
\centering
\caption{Qwen 80B: Logistic regression predicting abstention (Phase 4), full model}
\label{tab:OCTNEW_Qwen-phase3-coefs}
\begin{tabular}{lrrrr}
\toprule
Predictor & Coefficient ($\beta$) & Std.\ Error & $z$ & $P$ \\
\midrule
Intercept        & $1.785$  & 0.113  & $15.83$  & $1.93\times10^{-56}$ \\
Threshold (T)    & 0.034    & 0.0009 & 37.92    & $<2.2\times10^{-16}$ \\
Confidence (\%)  & $-0.051$ & 0.0016 & $-31.02$ & $2.90\times10^{-211}$ \\
Difficulty       & $-0.148$ & 0.077  & $-1.91$  & $5.61\times10^{-2}$ \\
\bottomrule
\end{tabular}

\footnotesize

\emph{Notes.} $n = 10{,}000$ independent trials (one model query per unique SimpleQA question at each instructed threshold; independent observations, not technical replicates). Abstention was modelled by logistic regression. Coefficient significance was assessed using two-sided Wald $z$-tests. No correction was made for multiple comparisons. Exact two-sided Wald-test $P$ values are reported in the table.

\end{table}
%--- Extended Data Table 13: Qwen Phase 4 model comparison ---updated
\begin{table}[H]
\centering
\caption{Qwen 80B: Model comparison for abstention predictions (Phase 4)}
\label{tab:OCTNEW_Qwen-phase3-model-compare}
\begin{tabular}{lccc}
\toprule
Model & AIC ($\downarrow$) & pseudo-$R^2$ ($\uparrow$) & Key likelihood-ratio tests \\
\midrule
Threshold only (T)                  & 12284.5 & 0.111 & -- \\
Threshold + Confidence              & 10962.7 & 0.207 &
vs.\ T: $\chi^2(1)=1323.80,\;P=7.622\times10^{-290}$ \\
Threshold + Difficulty              & 12062.4 & 0.127 & -- \\
Full (T + Confidence + Difficulty)  & 10961.1 & 0.207 &
vs.\ T+Conf: $\chi^2(1)=3.64,\;P=5.639\times10^{-2}$ \\
\bottomrule
\end{tabular}

\footnotesize

\emph{Notes.} $n = 10{,}000$ independent trials (one model query per unique SimpleQA question at each instructed threshold; independent observations, not technical replicates). Models were compared using two-sided likelihood-ratio tests. Exact two-sided $P$ values are reported in the table. No correction was made for multiple comparisons.

\end{table}

%======================================================================
% VERBAL CONFIDENCE ANALYSES
%======================================================================
\subsection*{Verbal Confidence Analyses}

% %--- Extended Data Figure 9: Verbal confidence prompt ---
% \begin{figure}[H]
%   \centering
%   \includegraphics[width=0.8\textwidth, angle = -90]{images/verbal_confidence_prompt.pdf}
%   \caption{{Verbal confidence prompt. This class based confidence prompt was used in \citet{kumaran2026llms} and based on the original version by \citet{yoon2025reasoning}. The midpoint of the confidence class (e.g. 0.95 for almost certain) was used for analysis.}
%   \label{fig:verbal_prompt}
% \end{figure}

%--- Extended Data Figure 11: Verbal confidence distributions ---
\begin{figure}[H]
  \centering
  \includegraphics[width=0.6\textwidth, angle = -90]{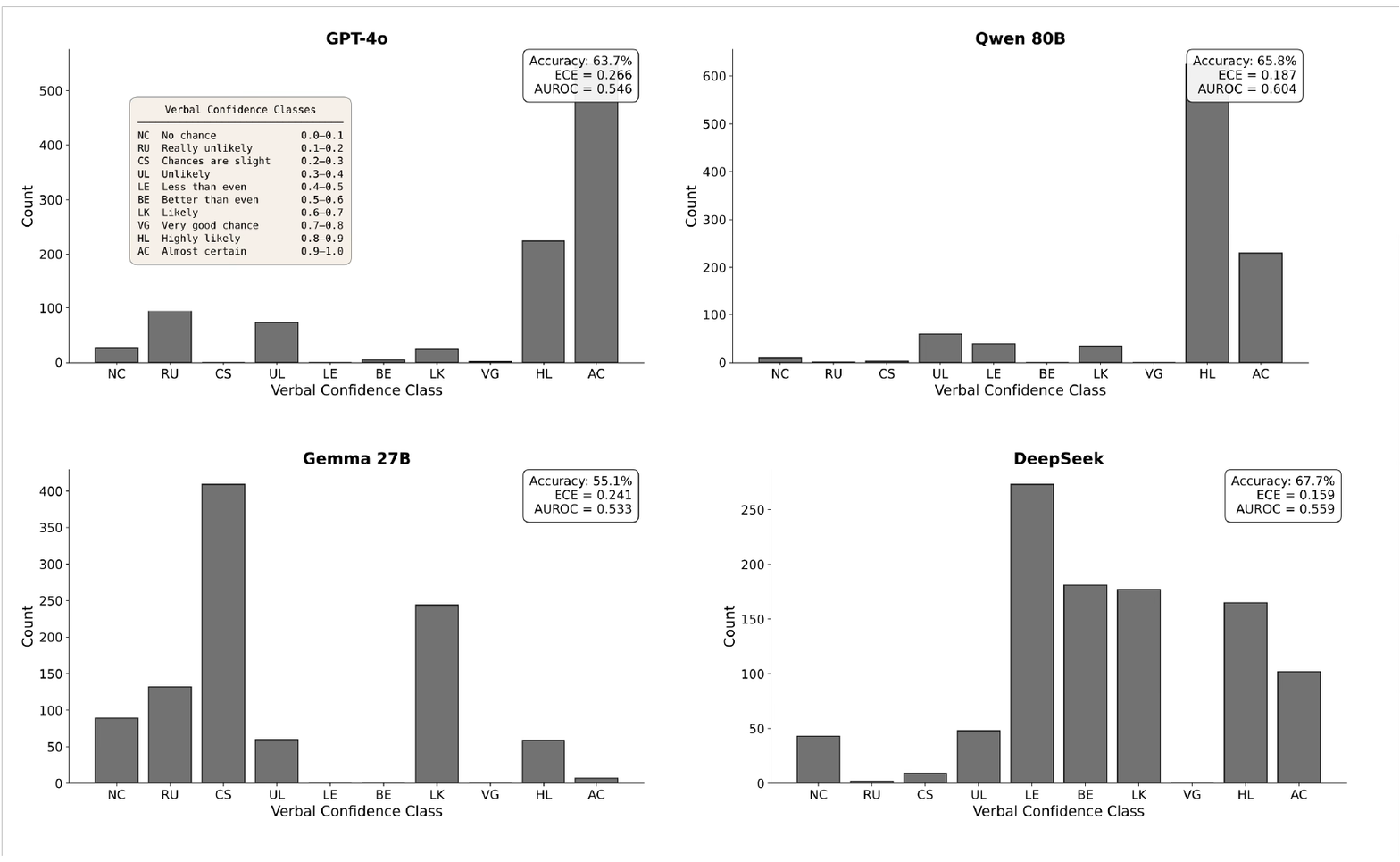}
  \caption{Distribution of verbal confidence class ratings observed in all 4 models.}
  \label{fig:verbal_dist}
\end{figure}

\paragraph{Properties of verbal confidence.}
Across all models, verbal confidence exhibited substantially higher Expected Calibration Error (ECE) than calibrated confidence (verbal: 0.16--0.27; calibrated: 0.03--0.11; see \ref{fig:verbal_dist} and \ref{tab:confidence_correlations}), indicating poorer calibration. However, isotonic calibration~\citep{niculescu2005predicting} fitted on the Phase~0 calibration set reduced verbal ECE to 0.005--0.032 (\ref{tab:confidence_correlations}), demonstrating that the poor calibration of raw verbal confidence reflects the mapping from class midpoints to accuracy rather than a fundamental limitation of the signal. Verbal confidence also showed lower AUROC---that is, weaker discrimination between correct and incorrect answers (verbal: 0.53--0.60; calibrated: 0.63--0.75). Because AUROC is invariant to monotonic transformations, this discrimination gap is the genuine difference between the two measures and is unaffected by any calibration procedure. Verbal and calibrated confidence were moderately correlated (Pearson $r = 0.32$--$0.43$; Spearman $\rho = 0.23$--$0.51$; see \ref{tab:confidence_correlations}), confirming that the two measures are related but not entirely redundant. Critically, isotonic-calibrated verbal confidence continued to independently predict abstention above calibrated logprobs across all model$\times$phase cells (all $p < .002$; see \ref{tab:isotonic}), confirming that the partial independence reflects genuine information content rather than a calibration artefact.

\paragraph{Calibrated confidence predicts abstention better than uncalibrated log-probability based confidence.}
We first asked whether the external calibration step improves the prediction of abstention. In logistic regressions predicting Phase 2 abstention from Phase 1 confidence (controlling for question difficulty, RAG scores, and sentence embeddings), calibrated confidence consistently outperformed uncalibrated log-probability based confidence (e.g., Gemma: $R^2 = .143$ vs .091; GPT-4o: $R^2 = .153$ vs .063; DeepSeek: $R^2 = .191$ vs .043; Qwen: $R^2 = .121$ vs .055 see \ref{tab:phase2_threeway}). The same pattern held in Phase 4 (see \ref{tab:phase4_threeway}). This difference persisted after standardization of both predictors, ruling out a scaling artifact. Thus, the temperature-scaled confidence measure that aligns the model's probability estimates with empirical accuracy also serves as a better predictor of when the model will choose to abstain (see Discussion).

\paragraph{Verbal confidence and calibrated confidence contribute partially independently to abstention.}
The central question is whether verbal confidence carries information about abstention that is not already captured by calibrated confidence. In Phase~2 (free abstention), adding verbal confidence to a model containing calibrated confidence and auxiliary controls significantly improved fit for all four models ($\Delta$AIC = +38 to +82; $\Delta R^2 = +.035$ to $+.064$; all two-sided $P \leq 5.278\times10^{-12}$; see Supplementary Table~15). The reverse was also true: adding calibrated confidence to the verbal+auxiliary model improved fit in every case ($\Delta$AIC = +35 to +90; $\Delta R^2 = +.031$ to $+.087$; all two-sided $P \leq 1.303\times10^{-9}$). These results were robust to post-hoc isotonic calibration of verbal confidence: isotonic-calibrated verbal confidence still independently predicted abstention across all models (all $P \leq 1.19\times10^{-3}$; Supplementary Table~17).

In Phase~4 (threshold-dependent abstention), both directions again showed partially independent contributions across all models (see Supplementary Table~16), with the exception of DeepSeek, where verbal confidence added only very marginally above calibrated confidence ($\Delta R^2 = .002$), though the effect remained statistically significant ($P = 1.117\times10^{-7}$). While the stronger single predictor varied by model, the two confidence channels both carried significant partially independent information about the model's propensity to abstain. As in Phase~2, these findings were robust to isotonic calibration of verbal confidence (all $P \leq 1.19\times10^{-3}$; Supplementary Table~17).

\paragraph{Activation decoding at last pre-answer token: additional analyses.}
To further characterise the internal confidence representation at the answer colon (AC) position, we conducted several additional analyses using Gemma~3 27B Phase~1 activations ($N = 1{,}999$).

\textit{Variance partitioning.} At layer~37, AC activations explained verbal confidence substantially above calibrated confidence ($\Delta R^2 = +.27$) while calibrated confidence added negligibly above AC ($\Delta R^2 = +.002$). The same asymmetry held for predicting calibrated confidence: AC added $\Delta R^2 = +.45$ above verbal confidence, while verbal confidence added only $\Delta R^2 = +.002$ above AC. The AC representation thus subsumes both observable confidence measures, consistent with both being lossy projections of a richer internal state.

\textit{Layer profile for calibrated logprobs.} The $R^2$ for predicting calibrated confidence continued to increase beyond layer~40 (reaching $.74$ at the final layer, 61), reflecting the progressive computation of the output distribution in later transformer layers. By contrast, verbal confidence decoding peaked at mid-to-late layers (37--40) and declined thereafter, consistent with verbal confidence being derived from an intermediate evaluative representation rather than from the final output logits.

\textit{Correctness does not independently predict abstention.} We trained a logistic probe to predict Phase~1 answer correctness from AC activations (peak AUROC~$= .59$ at layer~50) and tested whether the resulting correctness scores predicted Phase~2 abstention above calibrated confidence and verbal confidence. The residual correctness signal---the component of the correctness probe not already captured by the two confidence measures---showed no predictive value for abstention (residual AUROC~$= .51$, i.e., chance level). This indicates that the model's abstention behaviour does not draw on privileged access to answer correctness beyond what is already reflected in its confidence signals.

\textit{Question-content control.} Activations at the last question token (before answer options are presented) yielded negative $R^2$ values for both verbal confidence and calibrated confidence across all layers, confirming that the AC confidence signal is not attributable to question-level features.

\textit{Additional information at AC position.} Interestingly, AC activations predicted Phase~2 abstention substantially beyond calibrated confidence (AUROC~$= .68$), verbal confidence (AUROC~$= .64$), and both combined (AUROC~$= .72$), with a residual AUROC of $.77$ after controlling for both measures. This residual persisted even after additionally controlling for the full four-option calibrated confidence distribution, top-2 margin, and entropy over the 4 possible options (residual AUROC~$= .75$), indicating that the signal is not a richer summary of the output distribution over the 4 options. Question difficulty, semantic embeddings, and retrieval accessibility scores likewise failed to account for the residual (AUROC~$= .75$ after all auxiliary controls), whereas the same controls had a proportionally larger effect on the question-content position (residual AUROC from $.62$ to $.60$) than on AC (from $.77$ to $.75$). This suggests that AC encodes additional abstention-relevant information arising from the model's processing of the answer options---even though the abstention option was not available (i.e. activations were collected during Phase 1) beyond what is captured by any observable measure. The nature of this information remains an open question for future work.

%======================================================================
% VERBAL CONFIDENCE TABLES
%======================================================================
%--- Extended Data Table 14: Confidence measure properties ---
\begin{table}[H]
\centering
\small
\caption{Properties of three confidence measures: calibration (ECE), discrimination (AUROC), and pairwise correlations. All metrics computed on Phase~1 data ($N=1{,}000$ per model).}
\label{tab:confidence_correlations}
\begin{tabular}{l cccc cc cc}
\toprule
 & \multicolumn{4}{c}{\textbf{ECE} $\downarrow$} & \multicolumn{2}{c}{\textbf{AUROC} $\uparrow$} & \multicolumn{2}{c}{\textbf{Pearson $r$}} \\
\cmidrule(lr){2-5} \cmidrule(lr){6-7} \cmidrule(lr){8-9}
\textbf{Model} & Verbal & V\textsubscript{iso} & Cal & Uncal & Verbal & Cal & V$\leftrightarrow$C & V$\leftrightarrow$U \\
\midrule
Gemma 27B & .241 & .018 & \textbf{.033} & .388 & .533 & \textbf{.631} & .373 & .145 \\
GPT-4o & .266 & \textbf{.001} & .037 & .257 & .546 & \textbf{.746} & .316 & .174 \\
DeepSeek & .159 & .032 & \textbf{.107} & .242 & .559 & \textbf{.681} & .410 & .203 \\
Qwen 80B & .187 & \textbf{.005} & .079 & .290 & .604 & \textbf{.703} & .434 & .226 \\
\bottomrule
\end{tabular}
\begin{tablenotes}
\small
\item \textit{Note.} ECE = Expected Calibration Error (lower is better). AUROC = Area Under ROC (higher is better; correct vs.\ incorrect discrimination). Cal = calibrated logprobs (temperature-scaled). Uncal = uncalibrated ($\exp(\text{logprob})$). V\textsubscript{iso} = verbal confidence after isotonic calibration~\citep{niculescu2005predicting} fitted on Phase~0 data (AUROC unchanged as isotonic calibration is a monotonic transformation). Uncalibrated AUROC is omitted as it is identical to calibrated AUROC (AUROC is invariant to monotonic transformations). V$\leftrightarrow$C = Verbal vs.\ Calibrated; V$\leftrightarrow$U = Verbal vs.\ Uncalibrated. \textbf{Bold} = best per metric per model. Spearman correlations: V$\leftrightarrow$C $\rho = .23$--$.51$; V$\leftrightarrow$U $\rho = .17$--$.47$. All correlations $p < 10^{-5}$.
\end{tablenotes}
\end{table}

%--- Extended Data Table 15: Phase 2 three-way ---updated
\begin{table}[H]
\centering
\small
\caption{Three confidence measures predicting free abstention (Phase~2). All models include auxiliary controls (difficulty, RAG score, sentence embeddings).}
\label{tab:phase2_threeway}
\begin{tabular}{l ccccc cc cc}
\toprule
 & \multicolumn{5}{c}{\textbf{Pseudo-$R^2$}} & \multicolumn{2}{c}{\textbf{+Verbal}} & \multicolumn{2}{c}{\textbf{+Calibrated}} \\
\cmidrule(lr){2-6} \cmidrule(lr){7-8} \cmidrule(lr){9-10}
\textbf{Model} & Aux & Cal+A & Uncal+A & Verb+A & C+V+A & $\Delta$AIC & $\Delta R^2$ & $\Delta$AIC & $\Delta R^2$ \\
\midrule
Gemma 27B & .084 & .143 & .091 & \textbf{.175} & .207 & +73 & +.064 & +35 & +.031 \\
  & & & & & & \multicolumn{2}{c}{\scriptsize $P=4.602\times10^{-18}$} & \multicolumn{2}{c}{\scriptsize $P=1.303\times10^{-9}$} \\
\addlinespace[2pt]
GPT-4o & .044 & \textbf{.153} & .063 & .148 & .215 & +82 & +.061 & +90 & +.067 \\
  & & & & & & \multicolumn{2}{c}{\scriptsize $P=5.271\times10^{-20}$} & \multicolumn{2}{c}{\scriptsize $P=1.009\times10^{-21}$} \\
\addlinespace[2pt]
DeepSeek & .034 & \textbf{.191} & .043 & .147 & .233 & +38 & +.043 & +80 & +.087 \\
  & & & & & & \multicolumn{2}{c}{\scriptsize $P=2.121\times10^{-10}$} & \multicolumn{2}{c}{\scriptsize $P=1.696\times10^{-19}$} \\
\addlinespace[2pt]
Qwen 80B & .055 & .121 & .055 & \textbf{.125} & .155 & +46 & +.035 & +40 & +.031 \\
  & & & & & & \multicolumn{2}{c}{\scriptsize $P=5.278\times10^{-12}$} & \multicolumn{2}{c}{\scriptsize $P=8.512\times10^{-11}$} \\
\addlinespace[2pt]
\bottomrule
\end{tabular}

\begin{tablenotes}
\small
\item \textit{Note.} Logistic regression predicting Phase~2 abstention from Phase~1 confidence. $n=1{,}000$ independent trials for Gemma 27B, GPT-4o, and DeepSeek; $n=999$ for Qwen 80B after confidence matching. Each trial is one model query per unique SimpleQA question; observations are independent trials, not technical replicates. Aux = difficulty + RAG score + sentence embeddings. A = Aux. +Verbal: improvement from adding verbal confidence to the Cal+Aux model. +Calibrated: improvement from adding calibrated confidence to the Verbal+Aux model. $\Delta$AIC and $\Delta R^2$ denote improvements. Nested models were compared using two-sided likelihood-ratio tests. Exact two-sided $P$ values are reported in the table. No correction was made for multiple comparisons. \textbf{Bold} indicates the best single confidence measure with auxiliary controls.
\end{tablenotes}
\end{table}

%--- Extended Data Table 16: Phase 4 three-way ---updated
\begin{table}[H]
\centering
\small
\caption{Three confidence measures predicting threshold-dependent abstention (Phase~4). All models include threshold ($T$) and auxiliary controls.}
\label{tab:phase4_threeway}
\begin{tabular}{l ccccc cc cc}
\toprule
 & \multicolumn{5}{c}{\textbf{Pseudo-$R^2$}} & \multicolumn{2}{c}{\textbf{+Verbal}} & \multicolumn{2}{c}{\textbf{+Calibrated}} \\
\cmidrule(lr){2-6} \cmidrule(lr){7-8} \cmidrule(lr){9-10}
\textbf{Model} & Aux & Cal+A & Uncal+A & Verb+A & C+V+A & $\Delta$AIC & $\Delta R^2$ & $\Delta$AIC & $\Delta R^2$ \\
\midrule
Gemma 27B & .376 & .395 & .376 & \textbf{.425} & .430 & +533 & +.035 & +71 & +.005 \\
  & & & & & & \multicolumn{2}{c}{\scriptsize $P=2.447\times10^{-118}$} & \multicolumn{2}{c}{\scriptsize $P=1.087\times10^{-17}$} \\
\addlinespace[2pt]
GPT-4o & .156 & \textbf{.262} & .181 & .224 & .296 & +490 & +.034 & +1037 & +.072 \\
  & & & & & & \multicolumn{2}{c}{\scriptsize $P=6.039\times10^{-109}$} & \multicolumn{2}{c}{\scriptsize $P=5.879\times10^{-228}$} \\
\addlinespace[2pt]
DeepSeek & .125 & \textbf{.151} & .126 & .135 & .152 & +26 & +.002 & +264 & +.018 \\
  & & & & & & \multicolumn{2}{c}{\scriptsize $P=1.117\times10^{-7}$} & \multicolumn{2}{c}{\scriptsize $P=1.100\times10^{-59}$} \\
\addlinespace[2pt]
Qwen 80B & .155 & \textbf{.225} & .158 & .221 & .256 & +428 & +.031 & +474 & +.035 \\
  & & & & & & \multicolumn{2}{c}{\scriptsize $P=1.244\times10^{-95}$} & \multicolumn{2}{c}{\scriptsize $P=1.398\times10^{-105}$} \\
\addlinespace[2pt]
\bottomrule
\end{tabular}

\begin{tablenotes}
\small
\item \textit{Note.} Logistic regression predicting Phase~4 abstention. $n=11{,}000$ independent trials for Gemma 27B, GPT-4o, and DeepSeek; $n=9{,}990$ for Qwen 80B after confidence matching. Each trial is one model query per unique SimpleQA question at each instructed threshold; observations are independent trials, not technical replicates. All models include instructed threshold ($T$) and auxiliary controls (difficulty + RAG score + sentence embeddings). +Verbal: improvement from adding verbal confidence to $T$+Cal+Aux. +Calibrated: improvement from adding calibrated confidence to $T$+Verbal+Aux. $\Delta$AIC and $\Delta R^2$ denote improvements. Nested models were compared using two-sided likelihood-ratio tests. Exact two-sided $P$ values are reported in the table. No correction was made for multiple comparisons. \textbf{Bold} indicates the best single confidence measure with auxiliary controls.
\end{tablenotes}
\end{table}

%--- Extended Data Table 17: Isotonic calibration robustness ---updated
\begin{table}[H]
\centering
\small
\caption{Robustness check: isotonic calibration of verbal confidence.}
\label{tab:isotonic}
\begin{tabular}{l c c rrr rrr}
\toprule
 & & & \multicolumn{3}{c}{\textbf{+Raw Verbal}} & \multicolumn{3}{c}{\textbf{+Isotonic Verbal}} \\
\cmidrule(lr){4-6} \cmidrule(lr){7-9}
\textbf{Model} & \textbf{Phase} & $N$ & $\Delta$AIC & $\Delta R^2$ & $P$ & $\Delta$AIC & $\Delta R^2$ & $P$ \\
\midrule
Gemma 27B  & 2 & 1{,}000  & +73  & +.064 & $4.60\times10^{-18}$  & +21  & +.019 & $1.93\times10^{-6}$   \\
Gemma 27B  & 4 & 11{,}000 & +533 & +.035 & $2.45\times10^{-118}$ & +225 & +.015 & $2.50\times10^{-51}$  \\
GPT-4o     & 2 & 1{,}000  & +82  & +.061 & $5.27\times10^{-20}$  & +94  & +.070 & $1.48\times10^{-22}$  \\
GPT-4o     & 4 & 11{,}000 & +490 & +.034 & $6.04\times10^{-109}$ & +466 & +.033 & $9.61\times10^{-104}$ \\
DeepSeek   & 2 & 1{,}000  & +38  & +.043 & $2.12\times10^{-10}$  & +9   & +.011 & $1.05\times10^{-3}$   \\
DeepSeek   & 4 & 11{,}000 & +26  & +.002 & $1.12\times10^{-7}$   & +9   & +.001 & $1.19\times10^{-3}$   \\
Qwen 80B   & 2 & 999      & +46  & +.035 & $5.28\times10^{-12}$  & +83  & +.062 & $2.98\times10^{-20}$  \\
Qwen 80B   & 4 & 9{,}990  & +429 & +.031 & $1.24\times10^{-95}$  & +698 & +.051 & $3.41\times10^{-154}$ \\
\bottomrule
\end{tabular}

\begin{tablenotes}
\small
\item \textit{Note.} Each row reports a logistic regression predicting abstention. Phase~2 models include calibrated logprob confidence and auxiliary controls (difficulty, RAG score, sentence embeddings); Phase~4 models additionally include instructed threshold. $\Delta$AIC and $\Delta R^2$ reflect improvement from adding verbal confidence, either raw or isotonic-calibrated, to the Cal+Aux baseline model. Isotonic regression was fitted on the Phase~0 calibration set. Nested models were compared using two-sided likelihood-ratio tests. Exact two-sided $P$ values are reported in the table. No correction was made for multiple comparisons. Isotonic-calibrated verbal confidence independently predicted abstention in all model$\times$phase cells (all $P \leq 1.19\times10^{-3}$).
\end{tablenotes}
\end{table}
%======================================================================
% ACTIVATION DECODING FIGURE
%======================================================================

%--- Extended Data Figure 12: Activation decoding ---
\begin{figure}[H]
  \centering
  \includegraphics[width=0.6\textwidth]{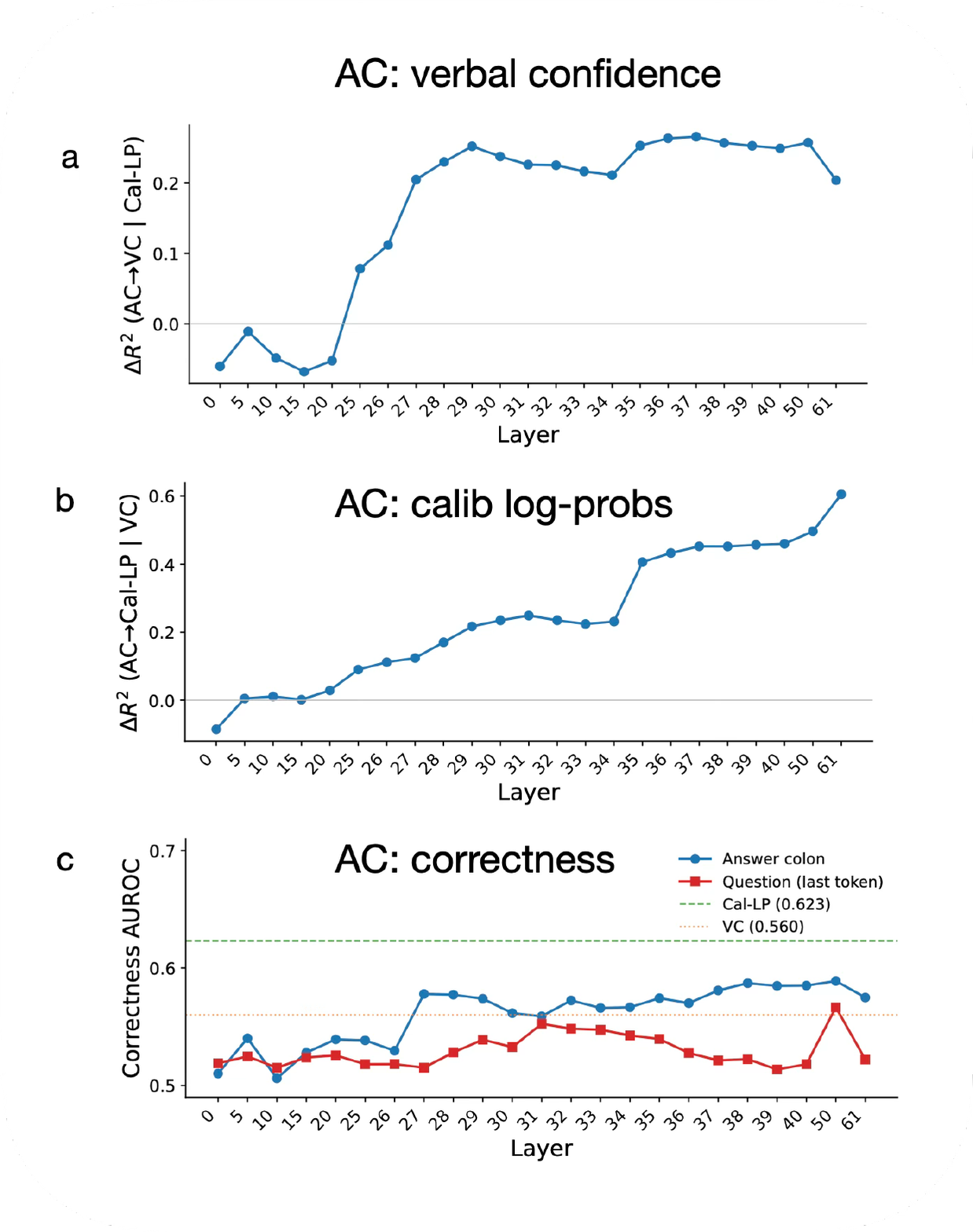}
  \caption{Linear decoding across layers in Gemma 3 27B using activations collected from last pre-answer token (answer-colon; AC) before the model's answer -- using the Phase 1 prompt. a) Variance in verbal confidence obtained by cross validated regularized ridge regression -- above that accounted for by calibrated confidence. Near zero variance explained by activation at the last question token (not shown). b) Variance in calibrated confidence by cross validated regularized ridge regression, over and above that accounted for by verbal confidence. Near zero variance explained by activation at the last question token (not shown). c) Discrimination of binary correctness (AUROC) by AC activations, and control (i.e. last question token) activations. Dashed lines show scalar baselines: calibrated confidence and verbal confidence.}
  \label{fig:AC_decoding}
\end{figure}

\section*{Supplementary References}

\end{document}